\documentclass[10pt,twocolumn,letterpaper]{article}

\usepackage{cvpr}
\usepackage{times}
\usepackage{epsfig}
\usepackage{graphicx}
\usepackage{amsmath}
\usepackage{amssymb}

\usepackage{extarrows}
\usepackage{algorithm}
\usepackage{algpseudocode}

\makeatletter
\@namedef{ver@everyshi.sty}{}
\makeatother

\usepackage{bm}
\usepackage{caption}
\usepackage{subcaption}
\usepackage{todonotes}
\usepackage{tcolorbox}
\usepackage{tikz}
\usepackage{tikz-3dplot}
\usepackage{tikzscale}
\usetikzlibrary{arrows,shapes,chains,matrix,positioning}
\usetikzlibrary{scopes,decorations,shadows,backgrounds,fit}
\usetikzlibrary{decorations.pathreplacing,calc,3d,quotes}
\usepackage{pgfmath}
\usepackage{threeparttable}

\usepackage{array}
\usepackage{mdwmath}
\usepackage{mdwtab}
\usepackage{eqparbox}
\usepackage{fixltx2e}
\usepackage{stfloats}
\usepackage{url}
\usepackage{diagbox}
\usepackage{verbatim}
\usepackage{booktabs}
\usepackage{multirow}
\usepackage{mathtools}
\usepackage{breqn}
\usepackage[pagebackref=true,breaklinks=true,letterpaper=true,colorlinks,bookmarks=false]{hyperref}

\DeclareMathOperator*{\argmax}{arg\,max}

\cvprfinalcopy 

\def\cvprPaperID{1586} 

\ifcvprfinal\fi
\begin{document}

\title{On the Intrinsic Dimensionality of Image Representations}

\author{Sixue Gong\quad\quad Vishnu Naresh Boddeti\quad\quad Anil K. Jain\\
Michigan State University, East Lansing MI 48824\\
{\tt\small \{gongsixu, vishnu, jain\}@msu.edu}
}

\maketitle
\thispagestyle{empty}


\begin{abstract}
This paper addresses the following questions pertaining to the intrinsic dimensionality of any given image representation: (i) estimate its intrinsic dimensionality, (ii) develop a deep neural network based non-linear mapping, dubbed DeepMDS, that transforms the ambient representation to the minimal intrinsic space, and (iii) validate the veracity of the mapping through image matching in the intrinsic space. Experiments on benchmark image datasets (LFW, IJB-C and ImageNet-100) reveal that the intrinsic dimensionality of deep neural network representations is significantly lower than the dimensionality of the ambient features. For instance, SphereFace's \cite{liu2017sphereface} 512-dim face representation and ResNet's \cite{he2016identity} 512-dim image representation have an intrinsic dimensionality of 16 and 19 respectively. Further, the DeepMDS mapping is able to obtain a representation of significantly lower dimensionality while maintaining discriminative ability to a large extent, 59.75\% TAR @ 0.1\% FAR in 16-dim vs 71.26\% TAR in 512-dim on IJB-C \cite{maze2018iarpa} and a Top-1 accuracy of 77.0\% at 19-dim vs 83.4\% at 512-dim on ImageNet-100.
\end{abstract}

\section{Introduction}
\begin{figure*}
    \centering
    \begin{subfigure}[t]{\textwidth}
        \centering
        \includegraphics[width=\textwidth]{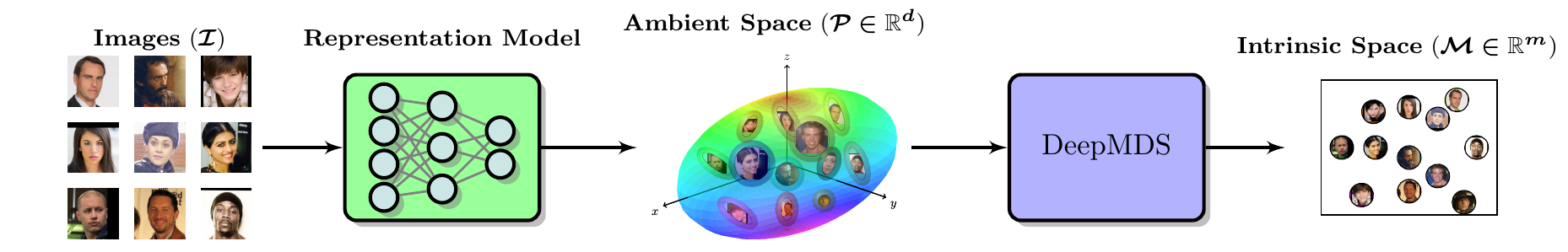}
        \caption{}
    \end{subfigure}
    \begin{subfigure}[t]{0.42\textwidth}
    \centering
	\begin{tikzpicture}[very thick]
        \node[] (a1) {\includegraphics[width=0.6\textwidth]{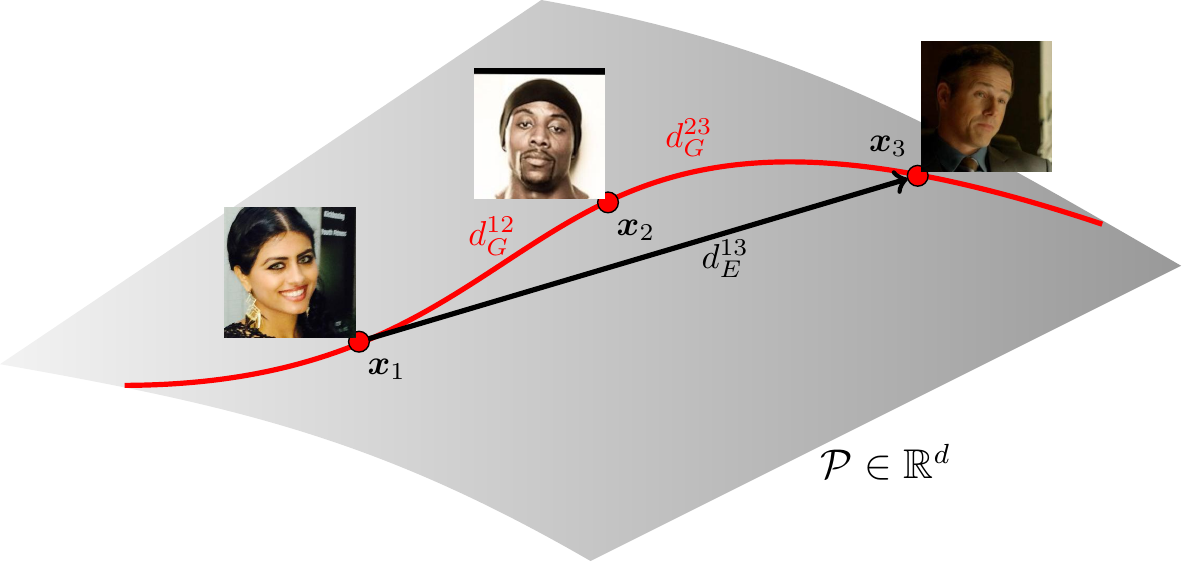}};
        \node[right of=a1, node distance=4.0cm] (a2) {\includegraphics[width=0.3\textwidth]{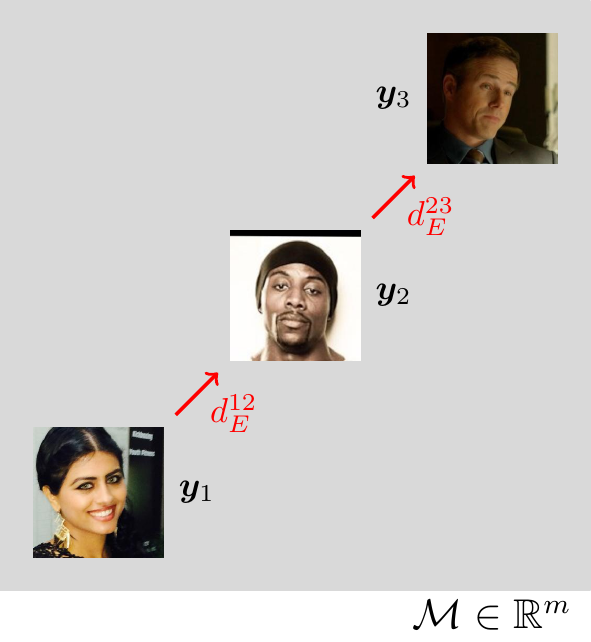}};
        \node[above of=a1, node distance=0.8cm] (a3) {};
        \node[above of=a2, node distance=0.8cm] (a4) {};
        \draw[->] (a3) to [out=60,in=150] (a4);
        \node[above right of=a1, node distance=3cm] (a4) {$\bm{y}=f(\bm{x},\bm{\theta})$};
    \end{tikzpicture}
    \caption{}
    \label{fig:ida}
    \end{subfigure}
    \begin{subfigure}[t]{0.56\textwidth}
    \centering
    \includegraphics[width=\textwidth, trim=1cm 3.2cm 1cm 3.2cm, clip]{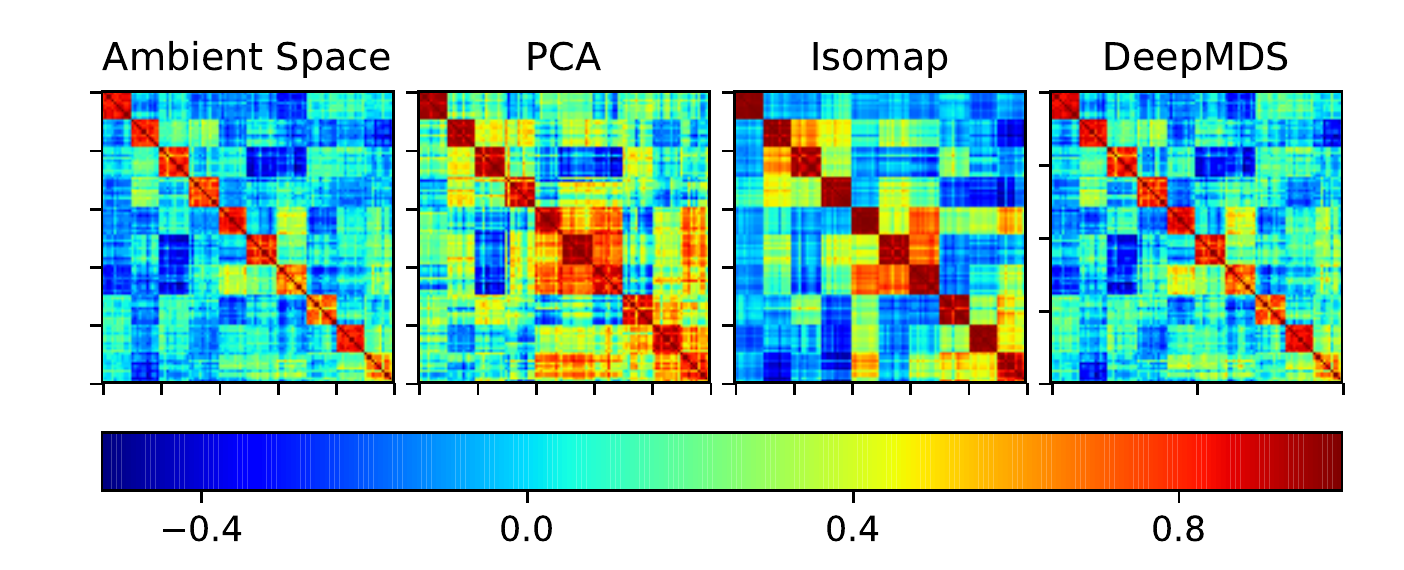}
    \caption{}
    \label{fig:ida}
    \end{subfigure}
    \caption{\footnotesize{\textbf{Overview:} This paper studies the manifold of feature vectors of images $\mathcal{I}$ obtained from a given representation model. (a) We estimate the intrinsic dimensionality (ID) of the ambient space $\mathcal{P}$ and propose DeepMDS, an unsupervised method, to map $\mathcal{P}$ to a low-dimensional intrinsic space $\mathcal{M}$. (b) Illustration of the ambient space $\mathcal{P}$ and intrinsic manifold $\mathcal{M}$ of a face representation. Here, while the ambient and linear dimension of the representation is three, its ID is only two. (b) Heatmaps of similarity scores between face pairs of 10 classes with 10 images per class for a representation with ID of 10-$dim$. The similarity is computed in four different spaces, the 512-$dim$ ambient space $\mathcal{P}$, 10-$dim$ space of linear dimensionality (PCA), 10-$dim$ intrinsic space $\mathcal{M}$ estimated by Isomap \cite{tenenbaum2000global} and by our DeepMDS model. The class separability, as shown by the diagonal blocks, is better maintained by DeepMDS.}}
    \label{fig:overview}
\end{figure*}
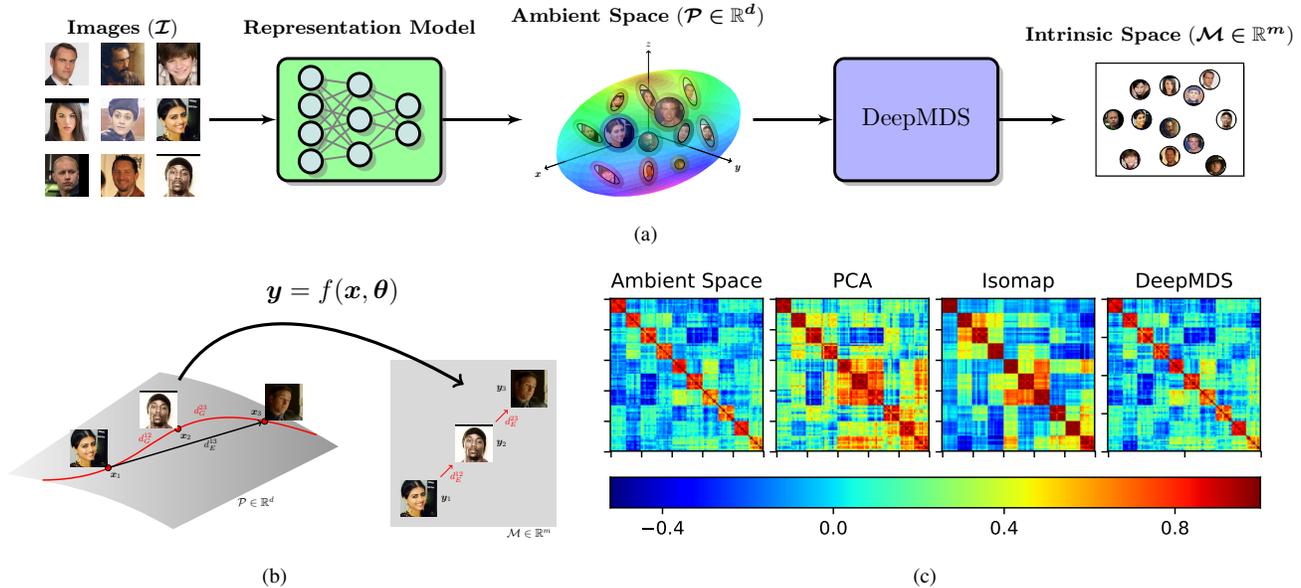

An image representation is an embedding function that transforms the raw pixel representation of the image to a point in a high-dimensional vector space. Learning or estimating such a mapping is motivated by two goals: (a) the compactness of the representation, and (2) the effectiveness of the mapping for the task at hand. While the latter topic has received substantial attention, ranging from PCA based Eigenfaces \cite{turk1991face} to deep neural network (DNN) based feature representations, there has been relatively little focus on the dimensionality of the representation itself. The dimensionality of image representations has ranged from hundreds to thousands of dimensions. For instance, current state-of-the-art image representations have 128, 512, 1024 and 4096 dimensions for FaceNet \cite{schroff2015facenet}, ResNet \cite{he2016identity}, SphereFace \cite{liu2017sphereface} and VGG \cite{simonyan2014very}, respectively. The choice of dimensionality is often determined by practical considerations, such as, ease of learning the embedding function \cite{taigman2015web}, constraints on system memory, etc. instead of the effective dimensionality necessary for image representation. This naturally raises the following fundamental but related questions, \emph{How compact can the representation be without any loss in recognition performance?} In other words, \emph{what is the intrinsic dimensionality of the representation?} And, \emph{how can one obtain such a compact representation?} Addressing these questions is the primary goal of this paper.

The intrinsic dimensionality (ID) of a representation refers to the minimum number of parameters (or degrees of freedom) necessary to capture the entire information present in the representation \cite{bennett1965representation}. Equivalently, it refers to the dimensionality of the $m$-dimensional manifold $\mathcal{M}$ embedded within the $d$-dimensional ambient (representation) space $\mathcal{P}$ where $m \leq d$. This notion of intrinsic dimensionality is notably different from common \emph{linear dimensionality} estimates obtained through e.g., principal component analysis (PCA). This linear dimension corresponds to the best linear subspace necessary to retain a desired fraction of the variations in the data. In principle, linear dimensionality can be as large as the ambient dimension if the variation factors are highly entangled with each other. An illustration of these concepts is provided in Fig. \ref{fig:overview}. 

The ability to estimate the intrinsic dimensionality of a given image representation is useful in a number of ways. At a \textbf{fundamental} level, the ID determines the true capacity and complexity of variations in the data captured by the representation, through the embedding function. In fact, the ID can be used to gauge the information content in the representation, due to its linear relation with Shannon entropy \cite{theiler1990estimating,costa2004geodesic}. Also, it provides an estimate of the amount of redundancy built into the representation which relates to its generalization capability. On a \textbf{practical} level, knowledge of the ID is crucial for devising optimal unsupervised strategies to obtain image features that are minimally redundant, while retaining its full ability to categorize images into different classes. Recognition in the intrinsic space can provide significant savings, both in memory requirements as well as processing time, across downstream tasks like large-scale face matching in the encrypted domain \cite{boddeti2018secure}, image matching and retrieval, etc. Lastly, gap between the ambient and intrinsic dimensionalities of a representation can serve as a useful indicator to drive the development of algorithms that can directly learn highly compact embeddings.

Estimating the ID of given data representation however is a challenging task. Such estimates are crucially dependent on the density variations in the representation, which in itself is difficult to estimate as images often lie on a topologically complex curved manifold \cite{talwalkar2008large}. More importantly, given an estimate of ID, how do we verify that it truly represents the dimensionality of the complex high-dimensional representation space? An indirect validation of the ID is possible through a mapping that transforms the ambient representation space to the intrinsic representation space while preserving its discriminative ability. However, there is no certainty that such a mapping can be found efficiently. In practice, finding such mappings can be considerably harder than estimating the ID itself.

We overcome both of these challenges by (1) adopting a topological dimensionality estimation technique based on the geodesic distance between points on the manifold, and (2) relying on the ability of DNNs to approximate the complex mapping function from the ambient space to the intrinsic space. The latter enables validation of the ID estimates through image matching experiments on the corresponding low-dimensional intrinsic representation of feature vectors.

The key contributions and findings of this paper are:

\noindent\textbf{--} The first attempt to estimate the intrinsic dimensionality of DNN based image representations.

\noindent\textbf{--} An unsupervised DNN based dimensionality reduction method under the framework of multidimensional scaling, called DeepMDS.

\noindent\textbf{--} Numerical experiments yield an ID estimate of, 12 and 16 for FaceNet \cite{schroff2015facenet} and SphereFace \cite{liu2017sphereface} face representations, respectively, and 19 for ResNet-34 \cite{he2016identity} image representation. The estimates are significantly lower than their respective ambient dimensionalities, 128-$dim$ for FaceNet and 512-$dim$ for the others.

\noindent\textbf{--} DeepMDS mapping is significantly better than other dimensionality reduction approaches in terms of its discriminative capability.

\section{Related Work}
\noindent \textbf{Image Representation: } The quest to develop image representations that are simultaneously robust and discriminative have led to extensive research on this topic. Amongst the earliest learning based approaches, Turk and Pentland proposed Eigenfaces \cite{turk1991face} that relied on principal component analysis (PCA) of data. Later on, integrated and high-dimensional spatially local features became prevalent for image recognition, notable examples include local binary patterns (LBP) \cite{ahonen2004face}, scale-invariant feature transform (SIFT) \cite{lowe1999object} and histogram of oriented gradients (HoG) \cite{dalal2005histograms}. In contrast to these hand-designed representations, the past decade has witnessed the development of end-to-end representation learning systems. Convolutional neural network based features now typify the state-of-the-art image representations \cite{he2016identity,szegedy2017inception,liu2017sphereface}. All of these representations are however characterized by features that range from hundreds to thousands of dimensions. While more compact representations are desirable, difficulties with optimizing DNNs with narrow bottlenecks \cite{taigman2015web} have proven to be the primary barrier towards realizing this goal.

\vspace{5pt}
\noindent \textbf{Intrinsic Dimensionality: } Existing approaches for estimating intrinsic dimensionality can be broadly classified into two groups: projection methods and geometric methods. The projection methods \cite{fukunaga1971algorithm,bruske1998intrinsic,verveer1995evaluation} determine the dimensionality by principal component analysis on local subregions of the data and estimating the number of dominant eigenvalues. These approaches have classically been used in the context of modeling facial appearance under different illumination conditions \cite{georghiades2001few} and object recognition with varying pose \cite{murase1995visual}. While they serve as an efficient heuristic, they do not provide reliable estimates of intrinsic dimension. Geometric methods \cite{pettis1979intrinsic,grassberger2004measuring,camastra2002estimating,kegl2003intrinsic,hein2005intrinsic,levina2005maximum} on the other hand model the intrinsic topological geometry of the data and are based on the assumption that the volume of a $m$-dimensional set scales with its size $\epsilon$ as $\epsilon^m$ and hence the number of neighbors less than $\epsilon$ also behaves the same way. Our approach in this paper is based on the topological notion of correlation dimension \cite{grassberger2004measuring,camastra2002estimating}, the most popular type of fractal dimensions. The correlation dimension implicitly uses nearest-neighbor distance, typically based on the Euclidean distance. However, Granata et.al. \cite{granata2016accurate} observe that leveraging the manifold structure of the data, in the form of geodesic distances induced by a neighborhood graph of the data, provides more realistic estimates of the ID. Building upon this observation we base our ID estimates on the geodesic distance between points. We believe that estimating the intrinsic dimensionality would serve as the first step towards understanding the bound on the minimal required dimensionality for representing images and aid in the development of novel algorithms that can achieve this limit.

\vspace{5pt}
\noindent \textbf{Dimensionality Reduction: } There is a tremendous body of work on the topic of estimating low-dimensional approximations of data manifolds lying in high-dimensional space. These include linear approaches such as Principal Component Analysis \cite{jolliffe1986principal}, Multidimensional Scaling (MDS) \cite{kruskal1964multidimensional} and Laplacian Eigenmaps \cite{belkin2003laplacian} and their corresponding non-linear spectral extensions, Locally Linear Embedding \cite{roweis2000nonlinear}, Isomap \cite{tenenbaum2000global} and Diffusion Maps \cite{coifman2006diffusion}. Another class of dimensionality reduction algorithms leverage the ability of deep neural networks to learn complex non-linear mappings of data including deep autoencoders \cite{hinton2006reducing}, denoising autoencoders \cite{vincent2008extracting,vincent2010stacked} and learning invariant mappings either with the contrastive loss \cite{hadsell2006dimensionality} or with the triplet loss \cite{schroff2015facenet}. While the autoencoders can learn a compact representation of data, such a representation is not explicitly designed to retain discriminative ability. Both the contrastive loss and the triplet loss have a number of limitations; (1) require similarity and dissimilarity labels from some source and cannot be trained in a purely unsupervised setting, (2) require an additional hyper-parameter, maximum margin of separation, which is difficult to pre-determine, especially for an arbitrary representation, and (3) do not maintain the manifold structure in the low-dimensional space. In this paper, we too leverage DNNs to approximate the non-linear mapping from the ambient to the intrinsic space. However, we consider an unsupervised setting (i.e., no similarity or dissimilarity labels) and cast the learning problem within the framework of MDS i.e., preserving the ambient graph induced geodesic distance between points in the intrinsic space.

\section{Approach}

Our goal in this paper is to compress a given image representation space. We achieve this in two stages\footnote{Traditional single-stage dimensionality reduction methods use visual aids to arrive at the final ID and intrinsic space, e.g., plotting the projection error against the ID values and looking for a ``knee" in the curve.}: (1) estimate the intrinsic dimensionality of the ambient image representation, and (2) learn the DeepMDS model to map the ambient representation space $\mathcal{P}\in\mathbb{R}^d$ to the intrinsic representation space $\mathcal{M}\in\mathbb{R}^m$ ($m \leq d$). The ID estimates are based on the one presented by \cite{granata2016accurate} which relies on two key ideas, (1) using graph induced geodesic distances to estimate the correlation dimension of the image representation topology, and (2) the similarity of the distribution of geodesic distances across different topological structures with the same intrinsic dimensionality. The DeepMDS model is optimized to preserve the interpoint geodesic distances between the feature vectors in the ambient and intrinsic space, and is trained in a stage-wise manner that progressively reduces the dimensionality of the representation. Basing the projection method on DNNs, instead of spectral approaches like Isomap, addresses the scalability and out-of-sample-extension problems suffered by spectral methods. Specifically, DeepMDS is trained in a stochastic fashion, which allows it to scale. Furthermore, once trained, DeepMDS provides a mapping function in the form of a feed-forward network that maps the ambient feature vector to its corresponding intrinsic feature vector. Such as map can easily be applied to new test data.

\subsection{Estimating Intrinsic Dimension}
We define the notion of intrinsic dimension through the classical concept of \emph{topological dimension} of the support of a distribution. This is a generalization of the concept of dimension of a linear space \footnote{Linear dimension is the minimum number of independent vectors necessary to represent any given point in this space as a linear combination.} to a non-linear manifold. Methods for estimating the topological dimension are all based on the assumption that the behavior of the number of neighbors of a given point on an $m$-dimensional manifold embedded within a $d$-dimensional space scales with its size $\epsilon$ as $\epsilon^m$. In other words, the density of points within an $\epsilon$-ball ($\epsilon \rightarrow 0$) in the ambient space is independent of the ambient dimension $d$ and varies only according to its intrinsic dimensionality $m$. Given a collection of points $\bm{X}=\{\bm{x}_1,\dots,\bm{x}_n\}$, where $\bm{x}_i\in\mathbb{R}^d$, the cumulative distribution of the pairwise distances $C(r)$ between the $n$ points can be estimated as,
\begin{equation}
    \footnotesize{
    C(r) = \frac{2}{n(n-1)}\sum_{i<j=1}^n H(r - \|\bm{x}_i-\bm{x}_j\|) =  \int_{0}^r p(r)dr
    }
\end{equation}
\noindent where $H(\cdot)$ is the Heaviside function and $p(r)$ is the probability distribution of the pairwise distances. In this paper, we choose the correlation dimension \cite{grassberger2004measuring}, a particular type of topological dimension, to represent the intrinsic dimension of the image representation. It is is defined as,
\begin{equation}
	m=\lim_{r\rightarrow 0}\frac{\ln C(r)}{\ln r} \implies \lim_{r\rightarrow 0} C(r) \propto r^m
\end{equation}
Therefore, the intrinsic dimension is crucially dependent on the accuracy with which the probability distribution can be estimated at very small length-scales (distances), i.e., $r\rightarrow 0$. Significant efforts have been devoted to estimating the intrinsic dimension through line fitting in the $\ln C(r)$ vs $\ln r$ space around the region where $r\rightarrow 0$ i.e.,
\begin{gather}
    m = \lim_{(r_2-r_1) \rightarrow0} \frac{\ln C(r_2) - \ln C(r_1)}{\ln r_2 - \ln r_1} \\
    = \lim_{r\rightarrow0} \frac{d \ln C(r)}{d \ln r} = \lim_{r\rightarrow0} \frac{p(r)}{C(r)}r = \lim_{r\rightarrow 0} m(r) \nonumber
\end{gather}
\begin{figure*}[t]
	\centering
	\begin{subfigure}[t]{0.45\textwidth}
	\centering
	\begin{tikzpicture}[very thick]
        \node[] (a1) {\includegraphics[width=0.55\textwidth]{figs/manifold1.pdf}};
        \node[right of=a1, node distance=4cm] (a2) {\includegraphics[trim=16cm 7.3cm 15cm 7.3cm, clip, width=0.4\textwidth]{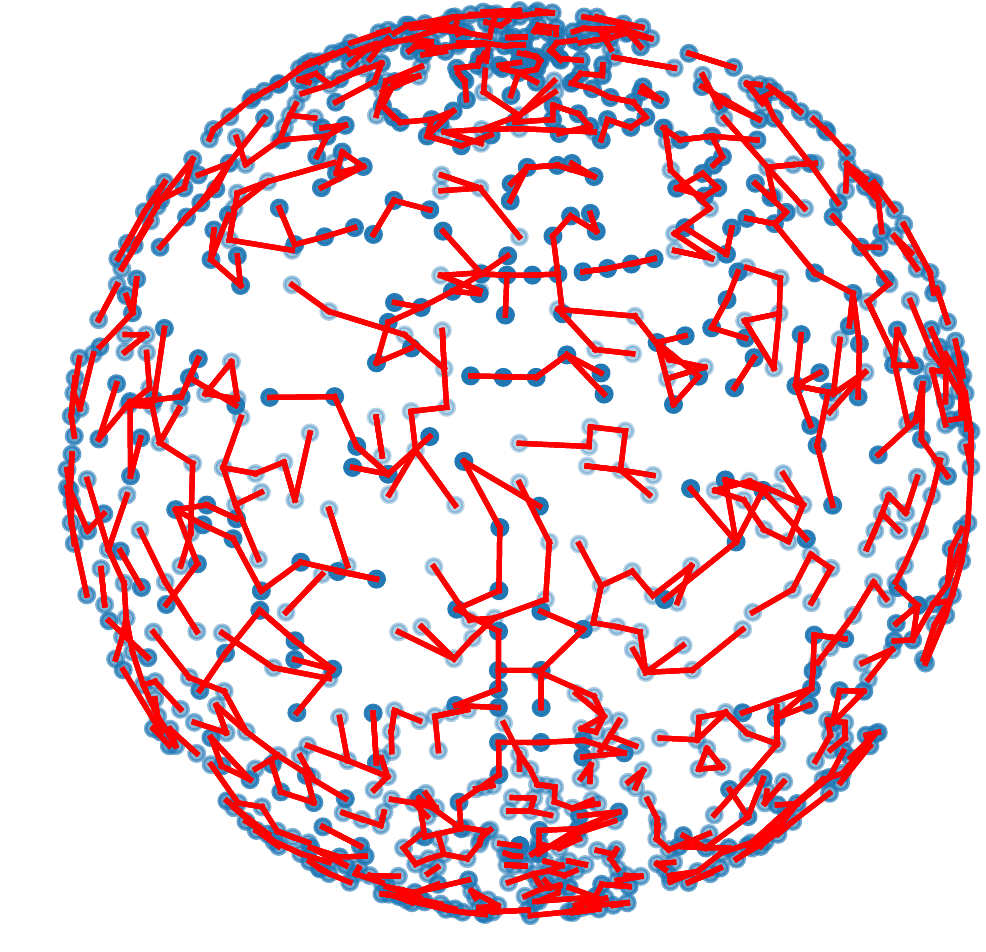}};
    \end{tikzpicture}
    \caption{Graph Induced Geodesic Distance}
    \label{fig:ida}
    \end{subfigure}
    \begin{subfigure}[t]{0.45\textwidth}
    \centering
    \scalebox{0.8}{
	\begin{tikzpicture}[very thick]
        \def\muone{0}
        \def\normalone{\x,{2/exp(((\x-\muone)^2)*2.0)-2}}
        \def\y{0}
        \def\fyone{2/exp(((\y-\muone)^2)*2.0)-2}
        \def\a{-0.7}
        \def\faone{2/exp(((\a-\muone)^2)*2.0)-2}
        \draw[color=blue,domain=-1.8:1.8] plot (\normalone) node[right] {};
        \draw[dashed, help lines, thick] ({\a},{\faone}) -- ({\a},-2);        
        \fill [fill=orange!40] ({\a},-2) -- plot[domain=\a:\y] (\normalone) -- (-\y,-2) -- cycle;
        \draw[<->, help lines, thick] ({\muone}, {\faone}) -- node[midway, below] {$2\sigma$} ({\a}, {\faone});
        \draw (-1.8,0.4) node[left] {$p(r)$};
        \draw (+1.7,-2.1) node[below] {$r$};
        \draw[->] (-1.8,-2) -- (+1.8,-2) node[right] {};
        \draw[->] (-1.8,-2) -- (-1.8,+0) node[above] {};
        \draw[dashed, help lines, thick] ({\muone},{+0}) -- ({\muone},-2) node[below] {$r_{max}$};
        \draw ({\muone},0.2) node[above] {\footnotesize Geodesic Distance};
        \node[] (z) at (4.6,-1) {\includegraphics[width=0.65\textwidth]{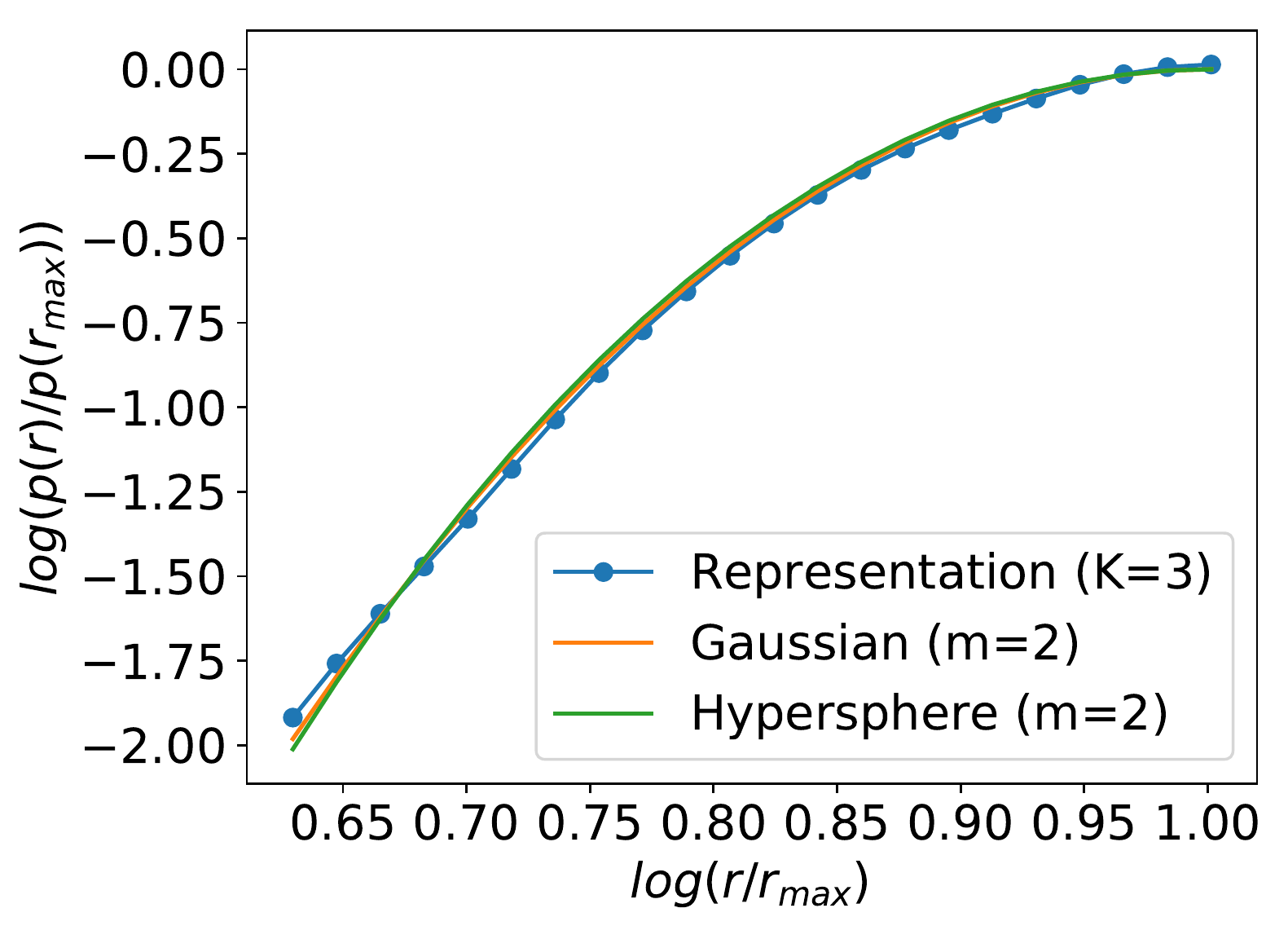}};
    \end{tikzpicture}}
    \caption{Topological Similarity}
    \label{fig:idb}
    \end{subfigure}
	\caption{\footnotesize{\textbf{Intrinsic Dimension:} Our approach is based on two observations: (a) Graph induced geodesic distance between images is able to capture the topology of the image representation manifold more reliably. As an illustration, we show the graph edges for the surface of a unitary hypersphere and a face manifold of ID two, embedded within a 3-$dim$ space. (b) The distribution of the geodesic distances (for distance $r_{max} - 2\sigma \leq r \leq r_{max}$, where $r_{max}$ is the distance at the mode) has been empirically observed \cite{granata2016accurate} to be similar across different topological structures with the same intrinsic dimensionality. The plot shows the distance distribution for a face representation, unitary hypersphere and a Gaussian distribution of ID two embedded within 3-$dim$ space.}}
	\label{fig:id}
\end{figure*}

The main drawback with this approach is the need for reliable estimates of $p(r)$ at very small length scales, which is precisely where the estimates are most unreliable when data is limited, especially in very high-dimensional spaces. Granata et al. \cite{granata2016accurate} present an elegant solution to this problem through three observations, (i) estimates of $m(r)$ can be stable even as $r\rightarrow0$ if the distance between points is computed as the graph induced shortest path between points instead of the euclidean distance, as is commonly the case, (ii) the probability distribution $p(r)$ at intermediate length-scales around the mode of $p(r)$ i.e., $(r_{max} - 2\sigma) \leq r \leq r_{max}$ can be conveniently used to obtain reliable estimates of ID, and (iii) the distributions $p(r)$ of different topological geometries are similar to each other as long as the intrinsic dimensionality is the same, or in other words the distribution $p(r)$ depends only on the intrinsic dimensionality and not on the geometric support of the manifolds.

Figure \ref{fig:id} provides an illustration of these observations. Consider two different manifolds, faces and the surface of a ($m+1$)-dimensional unitary hypersphere (henceforth referred to as $m$-hypersphere $\mathcal{S}^m$), with intrinsic dimensionality of $m=2$ but embedded within $3$-$dim$ Euclidean space. Beyond the nearest neighbor, the distance $r$ between any pair of points in the manifold is computed as the shortest path between the points as induced by the graph connecting all the points in the representation. Figure \ref{fig:idb} shows the distribution of $\log\frac{p(r)}{p(r_{max})}$ vs $\log\frac{r}{r_{max}}$ in the range $r_{max} - 2\sigma \leq r \leq r_{max}$, where $\sigma$ is the standard deviation of $p(r)$ and $r_{max}=\argmax\limits_r p(r)$ corresponds to the radius of the mode of $p(r)$. Interestingly, different topological geometries, namely, a face representation of ID two, a $2$-hypersphere and a $2$-$dim$ Gaussian, all embedded within $3$-$dim$ Euclidean space have almost identical distributions. More generally, the distribution of $\log\frac{p(r)}{p(r_{max})}$ vs $\log\frac{r}{r_{max}}$ in the range $r_{max} - 2\sigma \leq r \leq r_{max}$ is empirically observed to depend only on the intrinsic dimensionality, rather than the geometrical support of the manifold.

The intrinsic dimensionality of the representation manifold can thus be estimated by comparing the empirical distribution of the pairwise distances $\hat{p}_{\mathcal{M}}(r)$ on the manifold to that of a known distribution, such as the $m$-hypersphere in the range $r_{max}-\sigma \leq r \leq r_{max}$ (see appendix for Gaussian example). The distribution of the geodesic distance $p_{\mathcal{S}^m}(r)$ of $m$-hypersphere can be analytically expressed as, $p_{\mathcal{S}^m}(r)=c\sin^{m-1}(r)$, where $c$ is a constant and $m$ is the ID. Given $\hat{p}_{\mathcal{M}}(r)$, we minimize the Root Mean Squared Error (RMSE) between the distributions as,
\begin{equation}
    \footnotesize{
	\begin{aligned}
    \min_{c, m} \mbox{ } \int_{r_{max}-2\sigma}^{r_{max}}\left\|\log \hat{p}_{\mathcal{M}}(r)-\log(c) - (m-1)\log\left(\sin[r]\right)\right\|^2 \nonumber
    \end{aligned}}
\end{equation}
\noindent which upon simplification yields,
\begin{equation}
    \footnotesize{
	\begin{aligned}
    \min_{m} \mbox{ } \int_{r_{max}-2\sigma}^{r_{max}}\left\|\log \frac{\hat{p}_{\mathcal{M}}(r)}{\hat{p}_{\mathcal{M}}(r_{max})} - (m-1)\log\left(\sin\left[\frac{\pi r}{2r_{max}}\right]\right)\right\|^2 \nonumber
    \end{aligned}}
    \label{eq:id}
\end{equation}

The above optimization problem can be solved via a least-squares fit after estimating the standard deviation, $\sigma$, of $p(r)$ (see appendix for details). Such a procedure could, in principle, result in a fractional estimate of dimension. If one only requires integer solutions, the optimal value of $m$ can be estimated by rounding-off the least squares fit solution.

\subsection{Estimating Intrinsic Space}

The intrinsic dimensionality estimates obtained in the previous subsection alludes to the existence of a mapping, that can transform the ambient representation to the intrinsic space, but does not provide any solutions to find said mapping. The mapping itself could potentially be very complex and our goal of estimating it is practically challenging.
\tikzstyle{concat} = [rectangle, draw, fill=yellow!20, text width=1.0em, minimum height=1.0em, text centered, drop shadow]
\tikzstyle{conv} = [rectangle, draw, fill=blue!20, text width=2.0em, text centered, minimum height=6em, drop shadow]
\tikzstyle{batchnorm} = [rectangle, draw, fill=green!20, text width=2.0em, text centered, minimum height=6em, drop shadow]
\tikzstyle{relu} = [rectangle, draw, fill=red!20, text width=2.0em, text centered, minimum height=6em, drop shadow]
\tikzstyle{circ} = [draw,circle,fill=teal!20,node distance=2cm, drop shadow]
\tikzstyle{block} = [rectangle, draw, fill=gray!20, text width=20em, text centered, rounded corners, minimum height=5em]
\begin{figure*}[t]
    \centering
    \begin{tikzpicture}[node distance=1.5cm, auto,>=latex', very thick]
    \node (z) {};
    \node[left of=z, node distance=4cm] (a) {\includegraphics[width=0.3\textwidth]{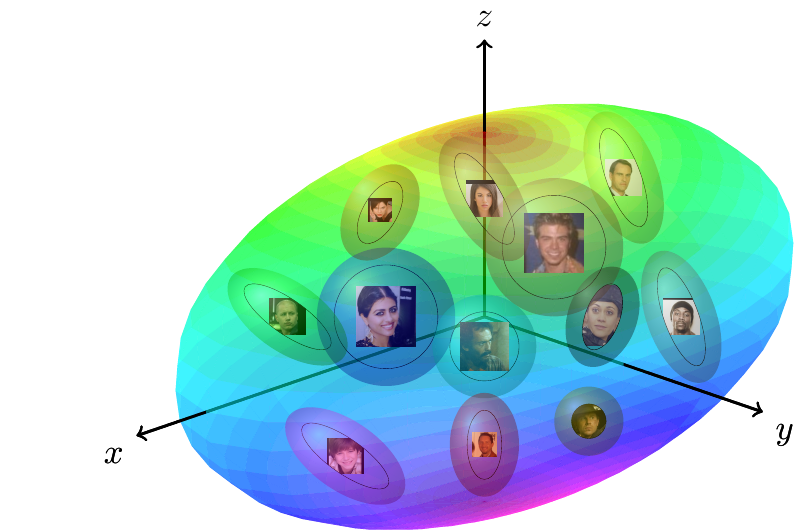}};
    \node[above of=a, node distance=2cm] (a1) {\footnotesize Ambient Space};
    \node[block, right of=a, node distance=6.9cm] (b) {};
    \node[above of=b, node distance=1.5cm] (b1) {\footnotesize Parametric Non-Linear Mapping};
    \node[right of=a, node distance=4.5cm] (c) {\includegraphics[width=0.25\textwidth]{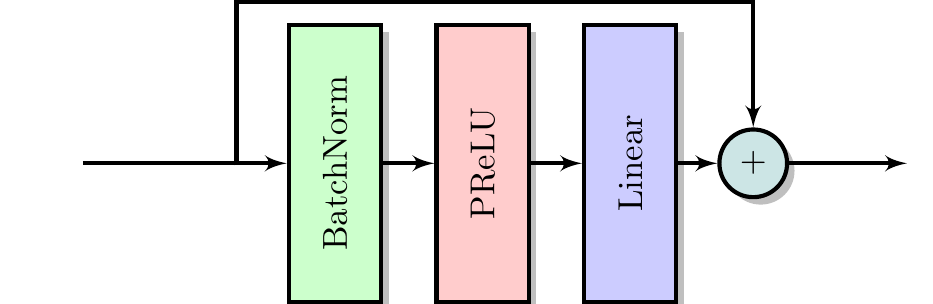}};
    \node[right of=c, node distance=2.4cm] (d) {$\dots$};
    \node[right of=d, node distance=2.0cm] (e) {\includegraphics[width=0.25\textwidth]{figs/mds.pdf}};
    \node[right of=e, node distance=3.7cm] (f) {\includegraphics[width=0.2\textwidth]{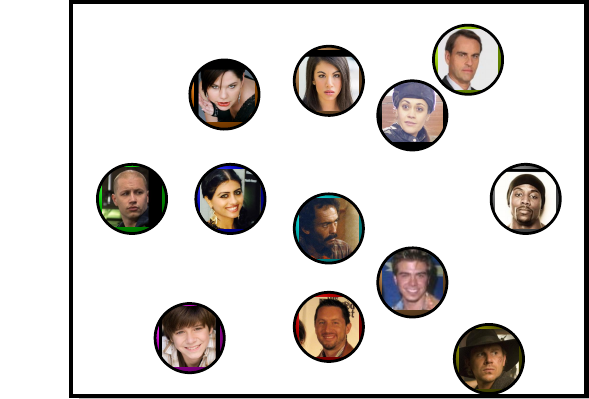}};
    \node[above of=f, node distance=1.5cm] (f1) {\footnotesize Intrinsic Space};
    \end{tikzpicture}
    \caption{\footnotesize{\textbf{DeepMDS Mapping:} A DNN based non-linear mapping is learned to transform the ambient space to a plausible intrinsic space. The network is optimized to preserve distances between pairs of points in the ambient and intrinsic space.}}
    \label{fig:nlle}
\end{figure*}

We base our solution to estimate a mapping from the ambient to the intrinsic space on Multidimensional scaling (MDS) \cite{kruskal1964multidimensional}, a classical mapping technique that attempts to preserve the distances (similarities) between points after embedding them in a low-dimensional space. Given data points $\bm{X}=\{\bm{x}_1,\dots,\bm{x}_n\}$ in the ambient space and $\bm{Y}=\{\bm{y}_1,\dots,\bm{y}_n\}$ the corresponding points in the intrinsic low-dimensional space, the MDS problem is formulated as,
\begin{equation}
	\min \sum_{i < j} \left(d_{H}(\bm{x}_i,\bm{x}_j) - d_{L}(\bm{y}_i,\bm{y}_j)\right)^2
\end{equation}
\noindent where $d_H(\cdot)$ and $d_L(\cdot)$ are distance (similarity) metrics in the ambient and intrinsic space, respectively. Different choices of the metric, leads to different dimensionality reduction algorithms. For instance, classical metric MDS is based on Euclidean distance between the points while using the geodesic distance induced by a neighborhood graph leads to Isomap \cite{tenenbaum2000global}. Similarly, many different distance metrics have been proposed corresponding to non-linear mappings between the ambient space and the intrinsic space. A majority of these approaches are based on spectral decompositions and suffer many drawbacks, (i) computational complexity scales as $\mathcal{O}(n^3)$ for $n$ data points, (ii) ambiguity in the choice of the correct non-linear function, and (iii) collapsed embeddings on more complex data \cite{hadsell2006dimensionality}.

To overcome these limitations, we employ a DNN to approximate the non-linear mapping that transforms the ambient representation, $\bm{x}$, to the intrinsic space, $\bm{y}$ by a parametric function $\bm{y}=f(\bm{x}; \bm{\theta})$ with parameters $\bm{\theta}$. We learn the parameters of the mapping within the MDS framework,

\begin{equation}
	\footnotesize{
	\begin{aligned}
	\min_{\bm{\theta}} \sum_{i=1}^n \sum_{i=1}^n\left[d_H(\bm{x}_i,\bm{x}_j) - d_L(f(\bm{x}_i;\bm{\theta}),f(\bm{x}_j;\bm{\theta}))\right]^2 + \lambda\|\bm{\theta}\|_2^2 \nonumber
	\end{aligned}}
\end{equation}
\noindent where the second term is a regularizer with a hyperparameter $\lambda$. Figure \ref{fig:nlle} shows an illustration of the DNN based mapping. 

In practice, directly learning the mapping from the ambient to the intrinsic space is very challenging, especially for disentangling a complex manifold under high levels of compression. We adopt a curriculum learning \cite{bengio2009curriculum} approach to overcome this challenge and progressively reduce the dimensionality of the mapping in multiple stages. We start with easier sub-tasks and progressively increase the difficulty of the tasks. For example, a direct mapping from $\mathbb{R}^{512} \rightarrow \mathbb{R}^{15}$ is instead decomposed into multiple mapping functions $\mathbb{R}^{512} \rightarrow \mathbb{R}^{256} \rightarrow \mathbb{R}^{128} \rightarrow \mathbb{R}^{64} \rightarrow \mathbb{R}^{32} \rightarrow \mathbb{R}^{15}$. We formulate the learning problem for $L$ mapping functions $\left(\bm{y}^l=f_l(\bm{x};\bm{\theta})\right)$ as:

\begin{equation}
    \footnotesize{
	\begin{aligned}
	\min_{\bm{\theta}_1,\dots,\bm{\theta}_L} \sum_{i=1}^n \sum_{j=1}^n \sum_{l=1}^L & \alpha_l\left[d_H(\bm{x}_i,\bm{x}_j) - d_L(\bm{y}_i^l,\bm{y}^l_j)\right]^2 + \lambda\|\bm{\theta}_l\|_2^2 \nonumber
	\end{aligned}}
\end{equation}
\noindent where $\bm{\theta}_l$ are the parameters of the $l$-th mapping. Appropriately scheduling the $\alpha_l$ weights enables us to set it up as a curriculum learning problem.

\section{Experiments}
In this section, first we will estimate the intrinsic dimensionality of multiple image representations over multiple datasets of varying complexity. Then, we will evaluate the efficacy of the proposed DeepMDS model in finding the mapping from the ambient to the intrinsic space while maintaining its discriminative ability.
\subsection{Datasets}
We choose two different domains of classification problems for our experiments, face verification and image classification. We consider two different face datasets for the former and the ImageNet ILSVRC-2012 for the latter. Recall that DeepMDS is an unsupervised method, so category information associated with the objects or faces is neither used for intrinsic dimensionality estimation nor for learning the mapping from the ambient to intrinsic space.

\vspace{2pt}
\noindent\textbf{LFW \cite{huang2007labeled}:} 13,233 face images of 5,749 subjects, downloaded from the web. These images exhibit limited variations in pose, illumination, and expression, since only faces that could be detected by the Viola-Jones face detector \cite{viola2004robust} were included in the dataset.

\vspace{2pt}
\noindent\textbf{IJB-C \cite{maze2018iarpa}:} IARPA Janus Benchmark-C (IJB-C) dataset consists of 3,531 subjects with a total of 31,334 (21,294 face and 10,040 non-face) still images and 11,779 videos (117,542 frames), an average of 39 images per subject. This dataset emphasizes faces with full pose variations, occlusions and diversity of subject occupation and geographic origin. Images in this dataset are labeled with ground truth bounding boxes and other covariate meta-data such as occlusions, facial hair and skin tone.

\vspace{2pt}
\noindent\textbf{ImageNet \cite{russakovsky2015imagenet}:} The ImageNet ILSVRC-2012 classification dataset consists of 1000 classes, with 1.28 million images for training and 50K images for validation. We use a subset of this dataset by randomly selecting 100 classes with the largest number of images, for a total of 130K training images and 5K testing images.

\subsection{Representation Models}
For the face-verification task, we consider multiple publicly available state-of-the-art face embedding models, namely, 128-$dim$ FaceNet \cite{schroff2015facenet} representation and 512-$dim$ SphereFace \cite{liu2017sphereface} representation. In addition, we also evaluate a 512-$dim$ variant of FaceNet\footnote{https://github.com/davidsandberg/facenet} that outperforms the 128-$dim$ version. All of these representations are learned from the CASIA WebFace \cite{yi2014learning} dataset, consisting of 494,414 images across 10,575 subjects. For image classification on the ImageNet dataset, we choose a pre-trained 34 layer version of the ResNet \cite{he2016identity} architecture.

\subsection{Baseline Methods}
\noindent\textbf{Intrinsic Dimensionality: } We select two different algorithms for estimating the intrinsic dimensionality of a given representation, a classical k-nearest neighbor based estimator \cite{pettis1979intrinsic} and ``Intrinsic Dimensionality Estimation Algorithm" (IDEA) \cite{rozza2012novel}.

\vspace{2pt}
\noindent\textbf{Dimensionality Reduction: } We compare DeepMDS against three dimensionality reduction algorithms, principal component analysis (PCA) for linear dimensionality reduction, Isomap \cite{tenenbaum2000global} and denoising autoencoders \cite{vincent2010stacked} (DAE).

\subsection{Intrinsic Dimensions}

\begin{figure}[t]
	\centering
	\begin{subfigure}[t]{0.22\textwidth}
		\centering
		\includegraphics[width=\textwidth]{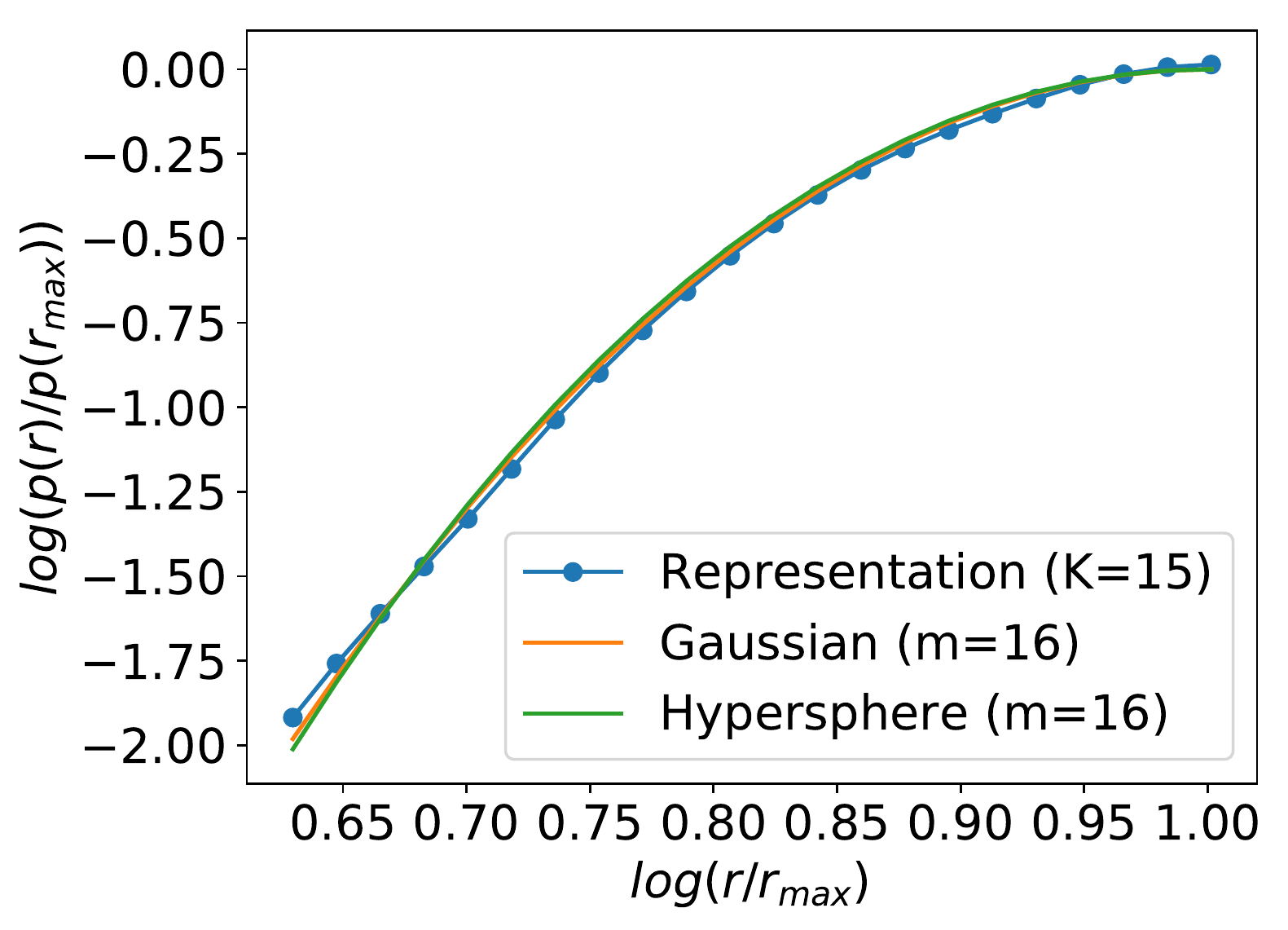}
		\caption{Distance Distribution $p(r)$}
		\label{fig:SphereFace-distance}
	\end{subfigure}
	\begin{subfigure}[t]{0.22\textwidth}
		\centering
		\includegraphics[width=\textwidth]{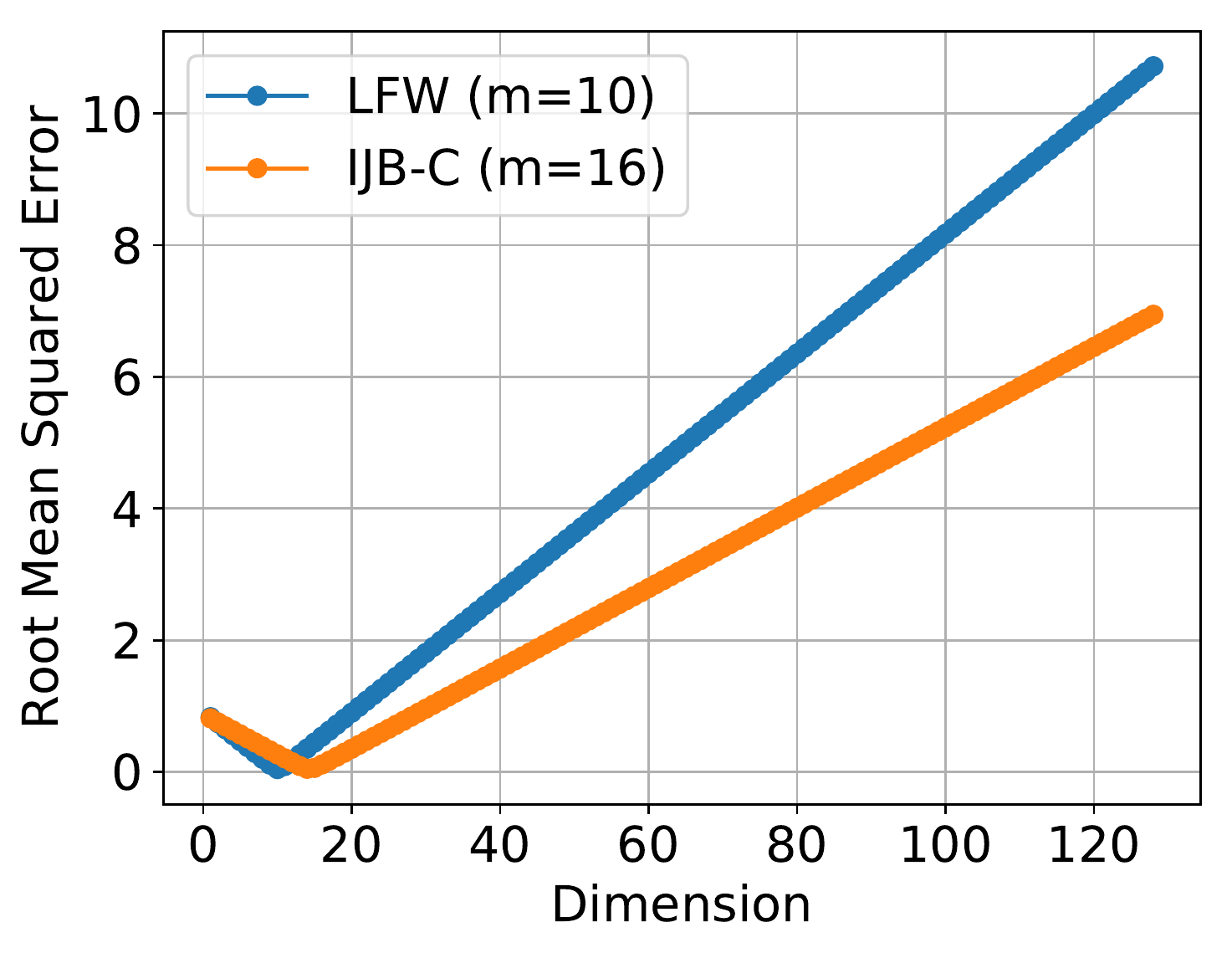}
		\caption{Least Squares Fitting}
		\label{fig:eucdmin}
	\end{subfigure}
	\caption{\textbf{Intrinsic Dimensionality:} (a) Geodesic distance distribution, and (b) global minimum of RMSE.}
	\label{fig:dmin}
\end{figure}
\noindent\textbf{Implementation Details:} The ID estimates for all the methods we evaluate are dependent on the number of neighbors $k$. For the baselines, $k$ is used to compute the parameters of the probability density. For our method, $k$ parameterizes the construction of the neighborhood graph. For the latter, the choice of $k$ is constrained by three factors; (1) $k$ should be small enough to avoid shortcuts between points that are close to each other in the Euclidean space, but are potentially far away in the corresponding intrinsic manifold due to highly complicated local curvatures. (2) On the other hand, $k$ should also be large enough to result in a connected graph i.e., there are no isolated data samples., and (3) $k$ that best matches the geodesic distance distribution of a hypersphere of the same ID i.e., $k$ that minimizes the RMSE. Figure \ref{fig:SphereFace-distance} shows the distance distributions for SphereFace with $k=15$, a 16-hypersphere and a 16-$dim$ Gaussian. The close similarity of the pairwise distance distributions of these manifolds in the graph induced geodesic distance space suggests that the ID of SphereFace (512-dim ambient space) is 16. Figure \ref{fig:eucdmin} shows the optimal RMSE for SphereFace\footnote{Similar curves for other representations and datasets can be found in the appendix.} at different values of $m$. For all the approaches we select the $k$-nearest neighbors using cosine similarity for SphereFace, Euclidean distance for ResNet and arc-length, $d(\bm{x}_1,\bm{x}_2)=\cos^{-1}\left(\frac{\bm{x}_1^T\bm{x}_2}{\|\bm{x}_1\|\|\bm{x}_2\|}\right)$, for FaceNet features, as the latter are normalized to reside on the surface of a unitary hypersphere. Finally, for simplicity, we round the ID estimates to the nearest integer for all the methods.
\begin{table}[t]
	\centering
	\caption{Intrinsic Dimensionality: Graph Distance \cite{granata2016accurate}}
    \label{table:id-graph}
		\centering
        \scalebox{0.77}{
		\begin{tabular}{c c c c c c c c c c c}
		\toprule
		\multirow{2}{*}{\textbf{\textit{Representation}}} && \multirow{2}{*}{dataset} && \multicolumn{7}{c}{\textbf{k}} \\
		\cline{5-11}
		 && && 4 && 7 && 9 && 15 \\
		\midrule
		\multirow{2}{*}{FaceNet-128} && LFW && \textbf{10*} && 13 && 11 && 18 \\
		&& IJB-C && 10 && 10 && 10 && \textbf{11*}\\
		\midrule
		\multirow{2}{*}{FaceNet-512} && LFW && \textbf{10*} && 11 && 11 && 17 \\
		&& IJB-C && 11 && 11 && 12 && \textbf{12*}\\
		\midrule
		\multirow{2}{*}{SphereFace} && LFW && \textbf{10*} && 11 && 13 && 9 \\
		&& IJB-C && 14 && 14 && 16 && \textbf{16*}\\
		\midrule
		ResNet-34 && ImageNet-100 && 16 && 18 && \textbf{19*} && 23 \\
		\bottomrule
		\end{tabular}}
\end{table}

\vspace{2pt}
\noindent\textbf{Experimental Results:} Table \ref{table:id-graph} reports the ID estimates from the graph method for different values of $k$\footnote{* denotes final ID estimate that satisfies all constraints on $k$.} and for different representation models across different datasets. Due to lack of space we report the ID estimates of the baselines in the appendix. We make a number of observations from our results: (1) Surprisingly, the ID estimates across all the datasets, feature representations and ID methods are significantly lower than the dimensionality of the ambient space, between 10 and 20, suggesting that image representations could, in principle, be almost 10$\times$ to 50$\times$ more compact. (2) Both\footnote{Reported in appendix due to space constraints.} the $k$-NN based estimator \cite{pettis1979intrinsic} and the IDEA estimator \cite{rozza2012novel} are less sensitive to the number of nearest neighbors in comparison to the graph distance based method \cite{granata2016accurate}, but are known to underestimate ID for sets with high intrinsic dimensionality \cite{verveer1995evaluation}. 

\subsection{Dimensionality Reduction}
Given the estimates of the dimensionality of the intrinsic space, we learn the mapping from the ambient space to \emph{a plausible} intrinsic space with the goal of retaining the discriminative ability of the representation. The true intrinsic representation (ID and space) is unknown and therefore not feasible to validate directly. However, verifying its discriminate power can serve to indirectly validate both the ID estimate and the learned intrinsic space.

\vspace{2pt}
\noindent\textbf{Implementation Details:} We first extract image features through the representations i.e., FaceNet-128, FaceNet-512 and SphereFace for face images and ResNet-34 for ImageNet-100. The architecture of the proposed DeepMDS model is based on the idea of skip connection laden residual units \cite{he2016identity}. We train the mapping from the ambient to intrinsic space in multiple stages with each stage comprising of two residual units. Once the individual stages are trained, all the $L$ projection models are jointly fine-tuned to maintain the pairwise distances in the intrinsic space. We adopt a similar network structure (residual units) and training strategy (stagewise training and fine-tuning) for the stacked denoising autoencoder baseline. From an optimization perspective, training the autoencoder is more computationally efficient than the DeepMDS model, $\mathcal{O}(n)$ vs $\mathcal{O}(n^2)$.

The parameters of the network are learned using the Adam \cite{kingma2014adam} optimizer with a learning rate of $3\times10^{-4}$ and the regularization parameter $\lambda=3\times10^{-4}$. We observed that using the cosine-annealing scheduler \cite{loshchilov2016sgdr} was critical to learning an effective mapping. To facilitate classification on ImageNet in the intrinsic space, after learning the projection, we separately learn a linear as well as a k-nearest neighbor ($k$-NN) classifier on the projected feature vectors of the training set.

\begin{table}[t]
	\centering
	\caption{LFW Face Verification for SphereFace Embedding}
    \label{table:verification-lfw}
		\centering
        \scalebox{0.8}{
		\begin{tabular}{c c c c c c c c c}
		\toprule
		\multirow{2}{*}{\textbf{\textit{Dimension}}} && \multicolumn{7}{c}{\textbf{Dimension Reduction method}} \\
		\cline{3-9}
		&& PCA && Isomap && DAE && DeepMDS \\
		\midrule
		512 && \multicolumn{7}{c}{96.74\%} \\
		\cline{3-9}
		256 && \textbf{96.75}\% && 92.88\% && 77.80\% && 96.73\% \\
		128 && \textbf{96.80}\% && 93.18\% && 32.95\% && 96.44\% \\
		64 && 91.71\% && 95.00\% && 32.04\% && \textbf{96.50}\% \\
		32 && 66.38\% && 95.31\% && 11.71\% && \textbf{96.31}\% \\
		16 && 32.67\% && 89.47\% && 27.53\% && \textbf{95.95}\% \\
		10 (ID) && 16.04\% && 77.31\% && 6.73\% && \textbf{92.33}\% \\
		\bottomrule
		\end{tabular}}
\end{table}

\begin{figure*}[t]
    \centering
    \begin{subfigure}[t]{0.23\textwidth}
        \centering
        \includegraphics[width=\textwidth]{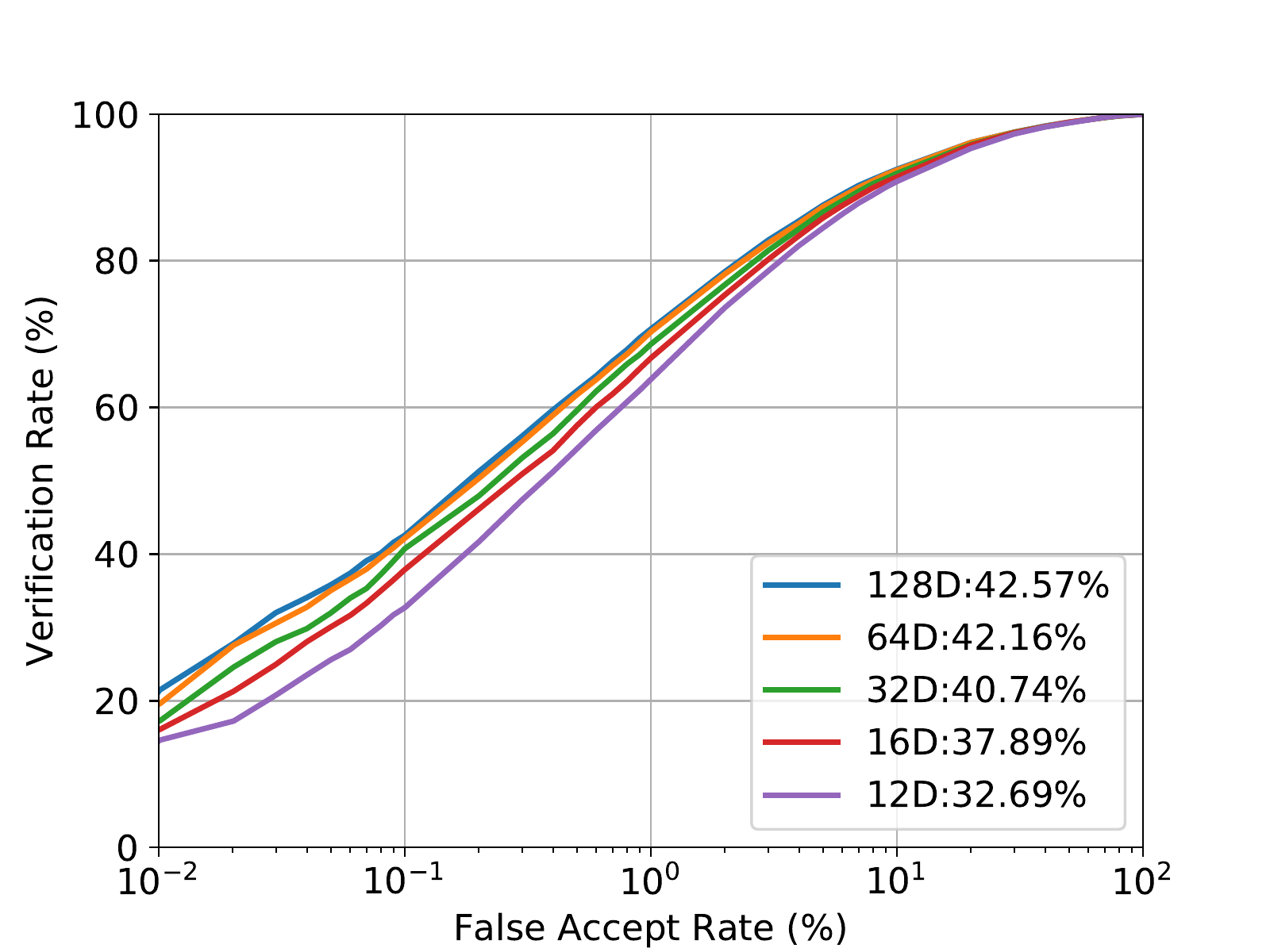}
        \caption{FaceNet-128}
        \label{fig:ijbc-pca}
    \end{subfigure}
    \begin{subfigure}[t]{0.23\textwidth}
        \centering
        \includegraphics[width=\textwidth]{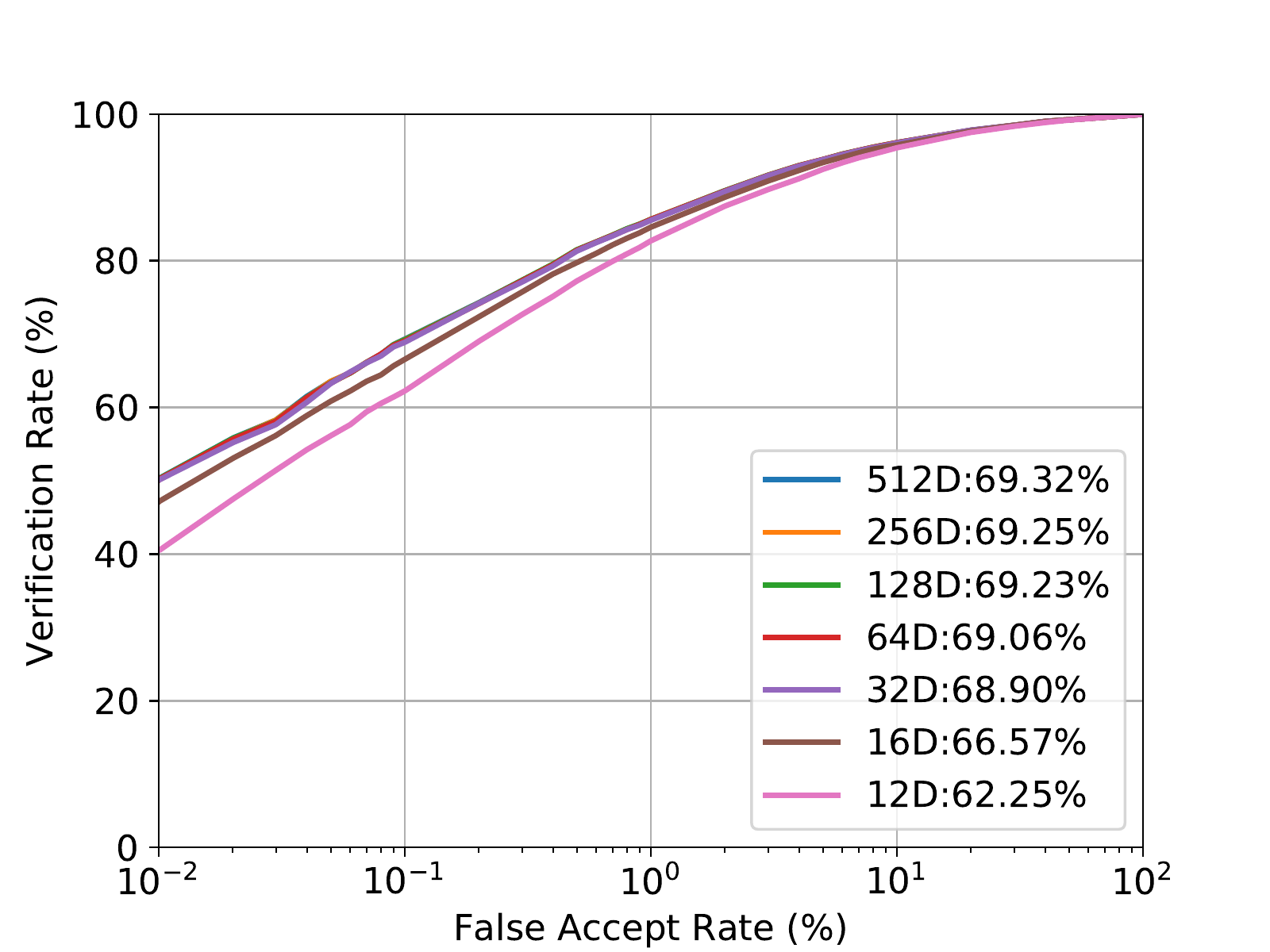}
        \caption{FaceNet-512}
        \label{fig:ijbc-isomap}
    \end{subfigure}
    \begin{subfigure}[t]{0.23\textwidth}
        \centering
        \includegraphics[width=\textwidth]{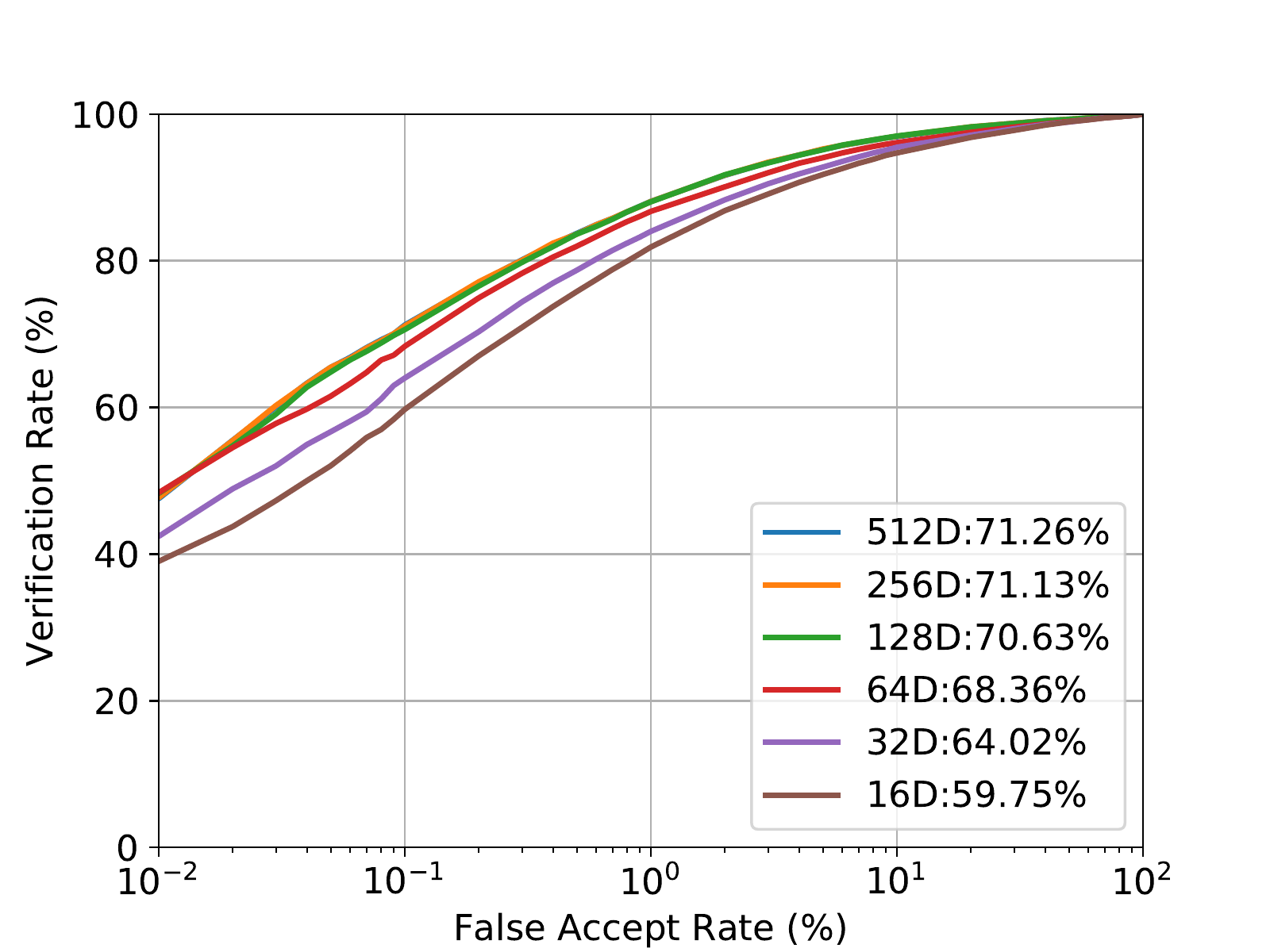}
        \caption{SphereFace}
        \label{fig:ijbc-auto}
    \end{subfigure}
    \begin{subfigure}[t]{0.23\textwidth}
        \centering
        \includegraphics[width=\textwidth]{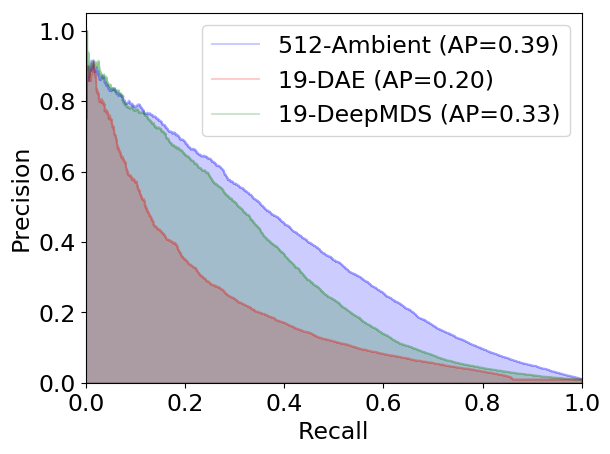}
        \caption{Image Retrieval}
        \label{fig:ijbc-auto}
    \end{subfigure}
    \caption{\footnotesize{Face Verification on IJB-C \cite{maze2018iarpa} (TAR @ 0.1\% FAR in legend) for the (a) FaceNet-128, (b) FaceNet-512 and (c) SphereFace embeddings and (d) Image retrieval on ImageNet-100 for the ambient 512-$dim$ ResNet-34 representation, the intrinsic 19-$dim$ space obtained from DAE and DeepMDS.}}
    \label{fig:ijbc-roc}
\end{figure*}
\noindent\textbf{Experimental Results:} We evaluate the efficacy of the learned projections, namely PCA, Isomap and DeepMDS, in the learned intrinsic space and compare their respective performance in the ambient space. Face representations are evaluated in terms of verification (TAR @ FAR) performance and classification on ImageNet-100 in terms of accuracy (Top-1 and Top-5). Given the ID estimate, designing an appropriate scheme for mapping the intrinsic manifold is much more challenging than the ID estimation itself. To show how dimensionality of the intrinsic space influences the performance of image representations, we evaluate and compare their performance at multiple intermediate spaces.

Face verification is performed on the IJB-C dataset following its verification protocol and on the LFW dataset following the BLUFR \cite{liao2014benchmark} protocol. Due to space constraints we only show results on the DeepMDS model here, corresponding results for the baseline dimensionality reduction methods can be found in the appendix. Figure \ref{fig:ijbc-roc} shows the ROC curves for the IJB-C dataset and the precision-recall curves for a image retrieval task on ImageNet-100. Table \ref{table:verification-lfw} reports the verification rate at FAR of 0.1\% on the LFW dataset. Similarly, Table \ref{table:imagenet} shows the Top-1 and Top-5 accuracy on ImageNet-100 for a pre-trained ResNet-34 representation via a parametric (linear) as well as a non-parametric (k-NN) classifier.
\begin{table}[t]
    \centering
    \caption{ImageNet-100 Classification (\%) for ResNet-34\label{table:imagenet}}
    \scalebox{0.65}{
    \begin{tabular}{c c c c c c c c c c}
        \toprule
        & \multirow{2}{*}{\textbf{Classifier}} & \multirow{2}{*}{\textbf{Method}} & \multicolumn{6}{c}{\textbf{Dimension}} \\
        \cline{4-9}
        & && 512 & 256 & 128 & 64 & 32 & 19 (ID)  \\
        \midrule
        \multirow{4}{*}{Top-1} & \multirow{2}{*}{Linear} & DAE & 80.0 & \textbf{80.9} & 73.2 & 70.0 & 63.1 & 50.2 \\
        & & DeepMDS & 80.0 & 79.4 & \textbf{76.1} & \textbf{71.4} & \textbf{70.2} & \textbf{68.0} \\
        \cline{2-9}
        & \multirow{2}{*}{$k$-NN} & DAE & 83.4 & \textbf{81.3} & \textbf{79.1} & 76.4 & 76.7 & 73.4 \\
        & & DeepMDS & 83.4 & 80.9 & 78.7 & \textbf{77.8} & \textbf{77.1} & \textbf{77.0} \\
        \midrule
        \multirow{2}{*}{Top-5} & \multirow{2}{*}{Linear} & DAE & 96.0 & \textbf{95.5} & 90.2 & \textbf{88.0} & 84.2 & 76.5 \\
        && DeepMDS & 96.0 & 95.3 & \textbf{93.1} & 85.2 & \textbf{85.2} & \textbf{84.8} \\
        \bottomrule
    \end{tabular}}
    \label{tab:my_label}
\end{table}

We make the following observations from these results: (1) for all the tasks the performance of the DeepMDS features up to 32 dimensions (for faces) is comparable to the original 128-$dim$ and 512-$dim$ features. The 10-$dim$ space of DeepMDS on LFW, consisting largely of frontal face images with minimal pose variations and facial occlusions, achieves a TAR of 92.33\% at 0.1\% FAR, a loss of about 4.5\% compared to the ambient space. The 12-$dim$ space of DeepMDS on IJB-C, with full pose variations, occlusions and diveristy of subject, achieves a TAR of 62.25\% at 0.1\% FAR, compared to 69.32\% in the ambient space. (2) the proposed DeepMDS model is able to learn a low-dimensional space up to the ID with a performance penalty of 5\%-10\% for compression factors of 30$\times$ to 40$\times$ for 512-dim representations, underscoring the fact that learning a mapping from ambient to intrinsic space is more challenging than estimating the ID itself. (3) In both tasks, we observe that the DeepMDS model is able to retain significantly more discriminative ability compared to the baseline approaches even at high levels of compression. Although DAE achieves comparative results on ImageNet-100 classification, DeepMDS significantly outperforms DAE for image retrieval tasks. While Isomap is more competitive than the other baselines it suffers from some  drawbacks: (i) Due to its iterative nature, it does not provide an explicit mapping function for new (unseen) data samples, while the autoencoder and DeepMDS models can map such data samples. Therefore, Isomap cannot be utilized to evaluate classification accuracy on the validation/test set of ImageNet-100 dataset, and (ii) Computational complexity of Isomap is $\mathcal{O}(n^3)$ and hence does not scale well to large datasets (IJB-C, ImageNet) and needs approximations, such as Nystr{\"o}m approximation \cite{talwalkar2008large}, for tractability.

\begin{table}[!h]
    \centering
    \caption{DeepMDS Training Methods (TAR @ 0.1\% FAR) \label{table:stagewise-ablation}}
    \scalebox{1.0}{
    \begin{tabular}{c c c c c}
        \toprule
        {\footnotesize \textbf{Method}} & {\footnotesize Direct} & {\footnotesize Direct+IS} & {\footnotesize Stagewise + Finetune} & {\footnotesize Stagewise} \\
        \midrule
        {\footnotesize\textbf{TAR}} & {\footnotesize 80.25} & {\footnotesize 86.15} & {\footnotesize 90.42} & {\footnotesize \textbf{92.33}}\\ 
        \bottomrule
    \end{tabular}}
\end{table}
\vspace{2pt}
\noindent\textbf{Ablation Study:} Here we demonstrate the efficacy of the stagewise learning process for training the DeepMDS model. All models have the same capacity. We consider four variants: (1) \textbf{Direct} mapping from the ambient to intrinsic space, (2) \textbf{Direct+IS:} direct mapping from ambient to intrinsic space with intermediate supervision at each stage i.e., optimize aggregate intermediate losses, (3) \textbf{Stagewise} learning of the mapping, and (4) \textbf{Stagewise+Fine-Tune:} the projection model trained stagewise and then fine-tuned. Table \ref{table:stagewise-ablation} compares the results of these variations on the LFW dataset (BLUFR protocol). Our results suggest that stagewise learning of the non-linear projection models is more effective at progressively disentangling the ambient representation. Similar trend was observed on larger datasets (IJB-C and ImageNet). In fact, stagewise training with fine-tuning was critical in learning an effective projection, both for DeepMDS as well as DAE.

\section{Concluding Remarks}
This paper addressed two questions, given a DNN based image representation, what is the minimum degrees of freedom in the representation i.e., its intrinsic dimension and can we find a mapping between the ambient and intrinsic space while maintaining the discriminative capability of the representation? Contributions of the paper include, (i) a graph induced geodesic distance based approach to estimate the intrinsic dimension, and (ii) DeepMDS, a non-linear projection to transform the ambient space to the intrinsic space. Experiments on multiple DNN based image representations yielded ID estimates of 9 to 20, which are significantly lower than the ambient dimension (10$\times$ to 40 $\times$). The DeepMDS model was able to learn a projection from ambient to the intrinsic space while preserving its discriminative ability, to a large extent, on the LFW, IJB-C and ImageNet-100 datasets. Our findings in this paper suggest that image representations could be significantly more compact and call for the development of algorithms that can directly learn more compact image representations.

\section{Appendix}

In this supplementary material we include; (1) Section \ref{sec:direct}: direct training of low-dimensional representations, (2) Section \ref{sec:intrinsic}: intrinsic dimensionality estimates from the baseline approaches \cite{rozza2012novel,pettis1979intrinsic}, (3) Section \ref{sec:baselines}: evaluation of the baseline dimensionality reduction techniques on the LFW and IJB-C datasets, (4) Section \ref{sec:derivations}: derivations of the intrinsic dimensionality estimation process, (5) Section \ref{sec:rmse}: RMSE and fitting plots for the graph distance based approach \cite{granata2016accurate}, and (6) Section \ref{sec:swiss-roll} intrinsic dimensionality estimation and learning and visualizing the learned projections on the Swiss Roll dataset.

\subsection{Direct Training \label{sec:direct}}
Our findings in this paper, that many current DNN representations can be significantly compressed, naturally begs the question: \emph{can we directly learn embedding functions that yield compact and discriminative embeddings in the first place?} Taigman et al. \cite{taigman2015web} study this problem in the context of learning face embeddings, and noted that a compact feature space creates a bottleneck in the information flow to the classification layer and hence increases the difficulty of optimizing the network when training from scratch. Given the significant developments in network architectures and optimization tools since then, we attempt to learn highly compact embedding directly from raw-data, using current best-practices, while circumventing the chicken-and-egg problem of not knowing the target intrinsic dimensionality before learning the embedding function. 

\begin{figure}
\centering
\begin{subfigure}[t]{0.48\textwidth}
    \centering
    \includegraphics[width=\textwidth]{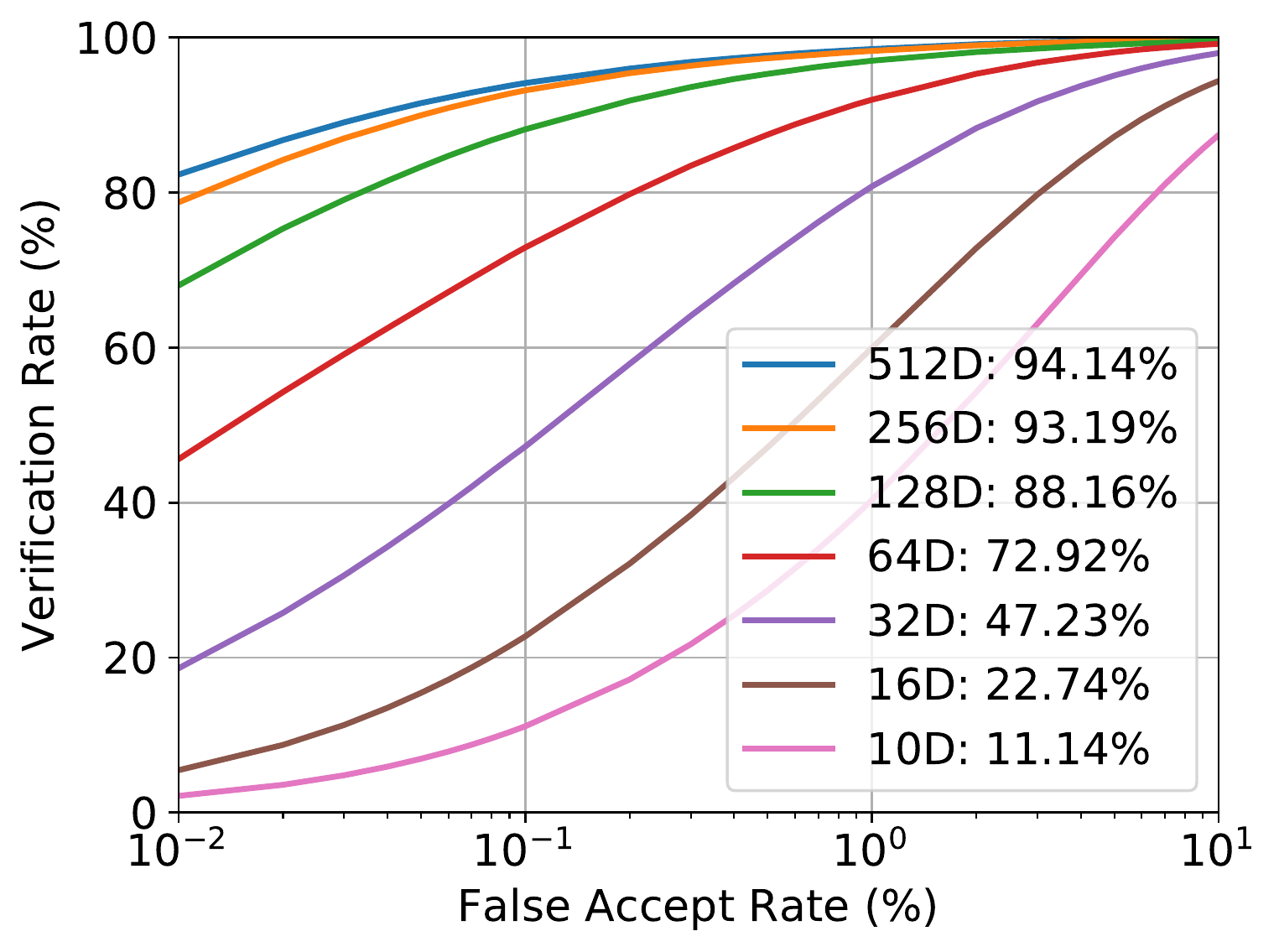}
    \caption{LFW}
\end{subfigure}
\begin{subfigure}[t]{0.48\textwidth}
    \centering
    \includegraphics[width=\textwidth]{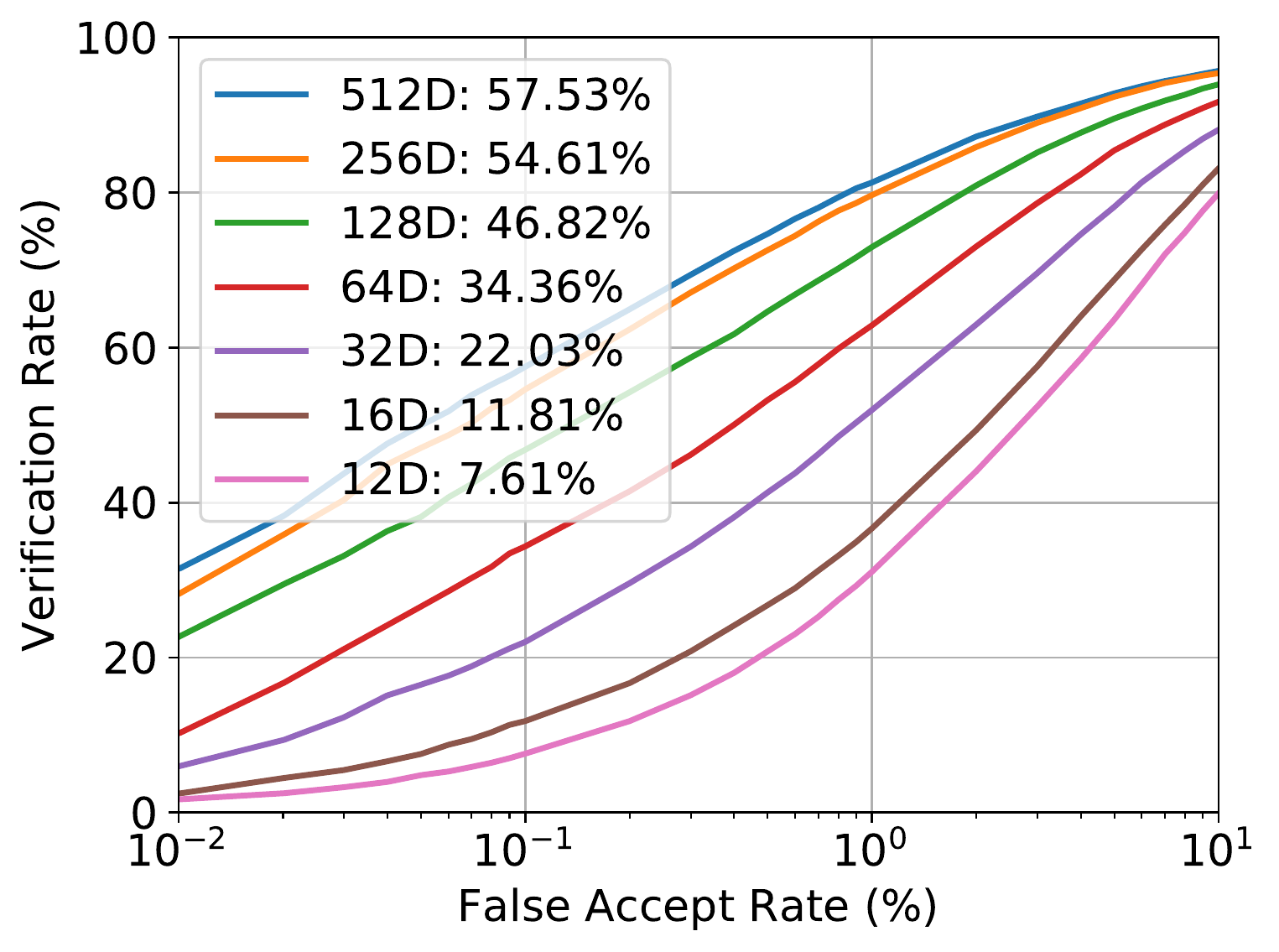}
    \caption{IJB-C}
\end{subfigure}
\caption{ROC curve on LFW and IJB-C datasets for the Inception ResNet V1 \cite{szegedy2017inception} model trained with different embedding dimensionality on the CASIA-WebFace \cite{yi2014learning} dataset.}
\label{fig:facenet-roc-train-scratch}
\end{figure}

We train\footnote{We build off of the publicly available implementation at \url{https://github.com/davidsandberg/facenet}} the Inception ResNet V1 \cite{szegedy2017inception} on the CASIA-WebFace \cite{yi2014learning} for embeddings of different sizes. Figure \ref{fig:facenet-roc-train-scratch} shows the ROC curves on the LFW and IJB-C datasets. The models suffer significant loss in performance as we decrease the dimensionality of the embeddings. In comparison the proposed DeepMDS based dimensionality reduction retains its discriminative ability even at high levels of compression. These results call for the development of algorithms that can directly learn compact and effective image representations.

\subsection{Intrinsic Dimensionality Estimation \label{sec:intrinsic}}
Table \ref{table:id-knn} and Table \ref{table:id-idea} reports the ID estimates from the k-nearest neighbor approach \cite{pettis1979intrinsic} and IDEA \cite{rozza2012novel}, respectively, for different representation models across different datasets that we consider. These approaches are known to underestimate the intrinsic dimensionality \cite{verveer1995evaluation}. We observe the same as our ID estimates for the baselines are lower than the estimates of the graph distance based approach that we use.
\begin{table}[!h]
	\centering
	\caption{Intrinsic Dimensionality: KNN \cite{pettis1979intrinsic}}
    \label{table:id-knn}
		\centering
        \scalebox{0.8}{
		\begin{tabular}{c c c c c c c c c c c}
		\toprule
		\multirow{2}{*}{\textbf{\textit{Representation}}} && \multirow{2}{*}{dataset} && \multicolumn{7}{c}{\textbf{k}} \\
		\cline{5-11}
		 && && 4 && 7 && 9 && 15 \\
		\midrule
		\multirow{2}{*}{FaceNet-128} && LFW && 10 && 10 && 11 && 11 \\
		&& IJB-C && 10 && 10 && 9 && 9\\
		\midrule
		\multirow{2}{*}{FaceNet-512} && LFW && 8 && 8 && 8 && 9 \\
		&& IJB-C && 10 && 10 && 9 && 9\\
		\midrule
		\multirow{2}{*}{Sphereface} && LFW && 6 && 7 && 7 && 8 \\
		&& IJB-C && 6 && 6 && 5 && 5\\
		\midrule
		ResNet-101 && ImageNet-100 && 25 && 20 && 19 && 16 \\
		\bottomrule
		\end{tabular}}
\end{table}
\begin{table}[!h]
	\centering
	\caption{Intrinsic Dimensionality: IDEA \cite{rozza2012novel}}
    \label{table:id-idea}
		\centering
        \scalebox{0.8}{
		\begin{tabular}{c c c c c c c c c c c}
		\toprule
		\multirow{2}{*}{\textbf{\textit{Representation}}} && \multirow{2}{*}{dataset} && \multicolumn{7}{c}{\textbf{k}} \\
		\cline{5-11}
		 && && 4 && 7 && 9 && 15 \\
		\midrule
		\multirow{2}{*}{FaceNet-128} && LFW && 14 && 13 && 13 && 12 \\
		&& IJB-C && 14 && 11 && 10 && 9\\
		\midrule
		\multirow{2}{*}{FaceNet-512} && LFW && 12 && 10 && 10 && 10 \\
		&& IJB-C && 14 && 11 && 10 && 9\\
		\midrule
		\multirow{2}{*}{Sphereface} && LFW && 10 && 9 && 9 && 9 \\
		&& IJB-C && 8 && 7 && 6 && 5\\
		\midrule
		ResNet-101 && ImageNet-100 && 21 && 21 && 20 && 20 \\
		\bottomrule
		\end{tabular}}
\end{table}

\subsection{Intrinsic Dimension Mapping \label{sec:baselines}}
\begin{figure*}[t]
    \centering
    \begin{subfigure}[t]{0.33\textwidth}
        \centering
        \includegraphics[width=\textwidth]{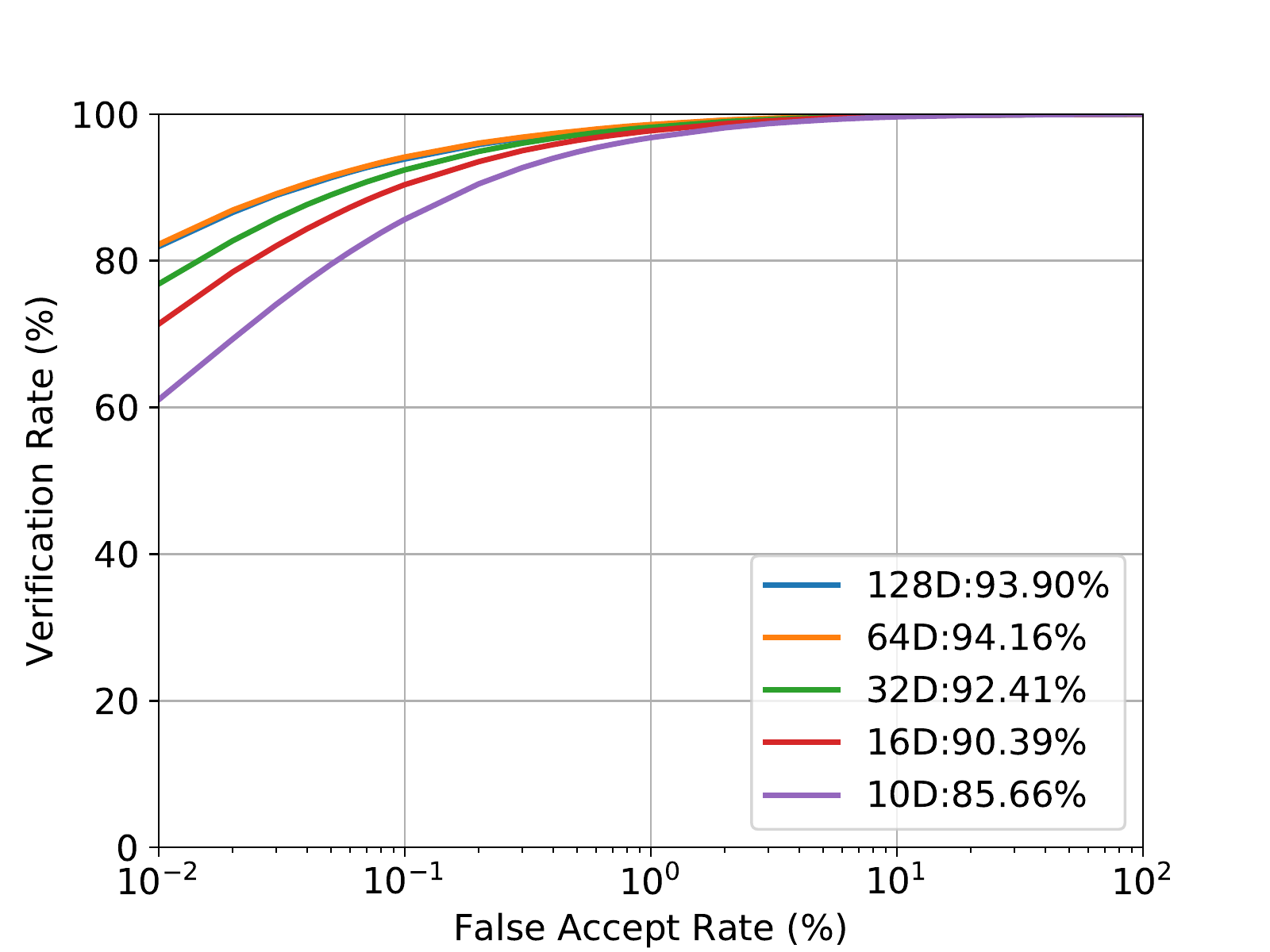}
        \caption{LFW: FaceNet-128}
    \end{subfigure}
    \begin{subfigure}[t]{0.33\textwidth}
        \centering
        \includegraphics[width=\textwidth]{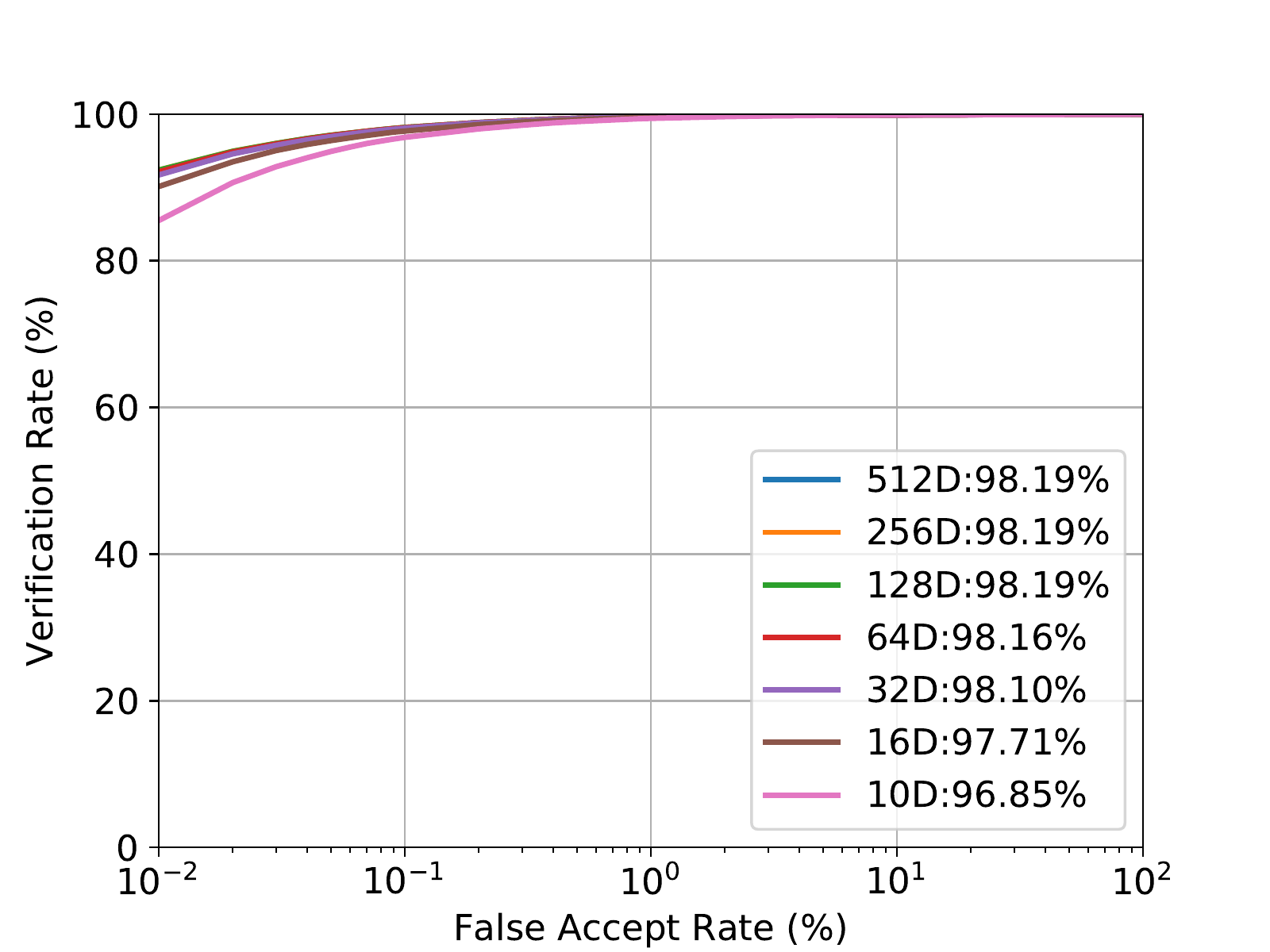}
        \caption{LFW: FaceNet-512}
    \end{subfigure}
    \begin{subfigure}[t]{0.33\textwidth}
        \centering
        \includegraphics[width=\textwidth]{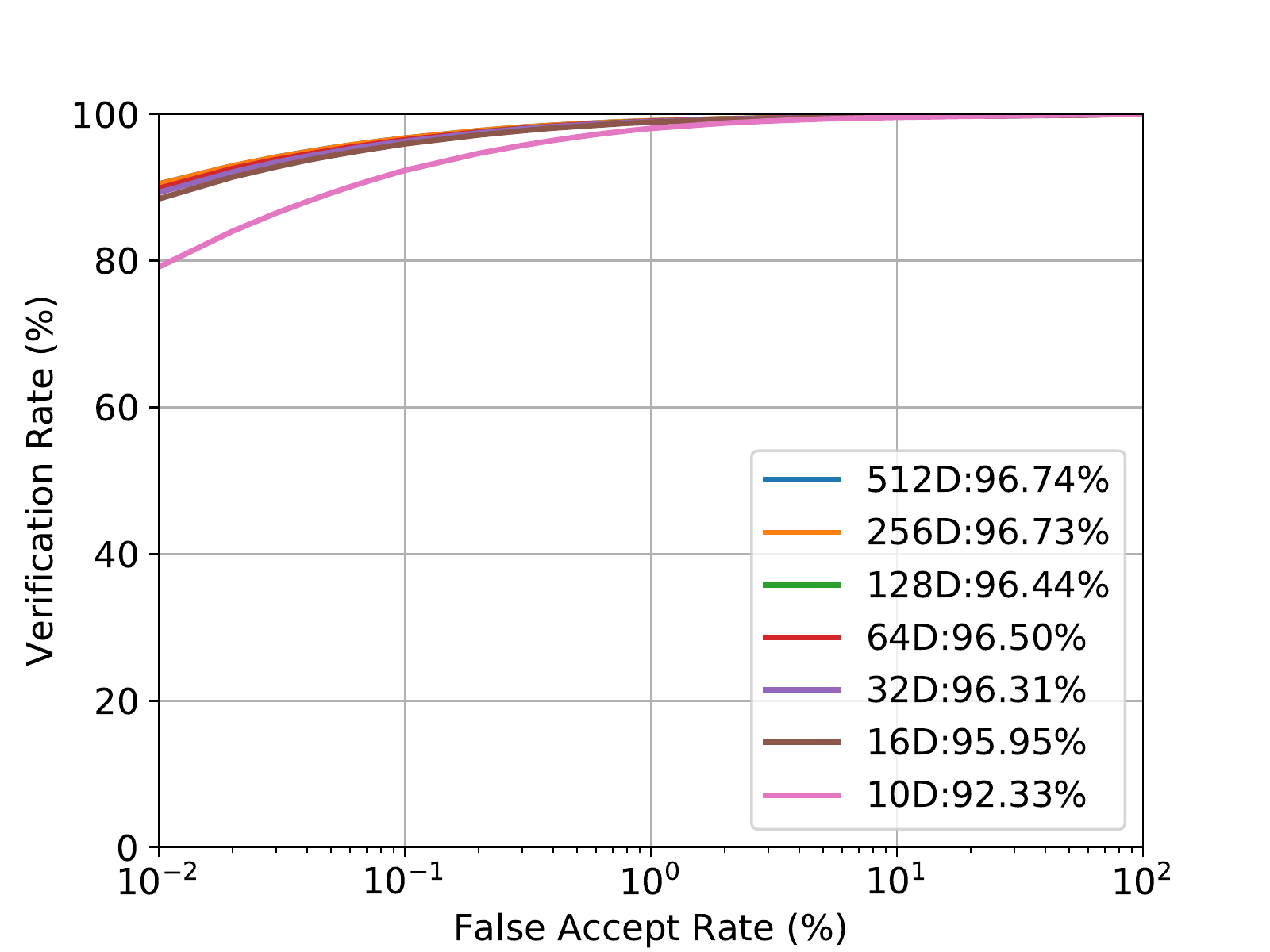}
        \caption{LFW: SphereFace}
    \end{subfigure}
    \caption{\textbf{DeepMDS:} Face Verification on LFW (BLUFR) dataset for the (a) FaceNet-128, (b) FaceNet-512 and (c) SphereFace embeddings.}
    \label{fig:deepmds}
\end{figure*}

\begin{figure*}[t]
    \centering
    \begin{subfigure}[t]{0.16\textwidth}
        \centering
        \includegraphics[width=\textwidth]{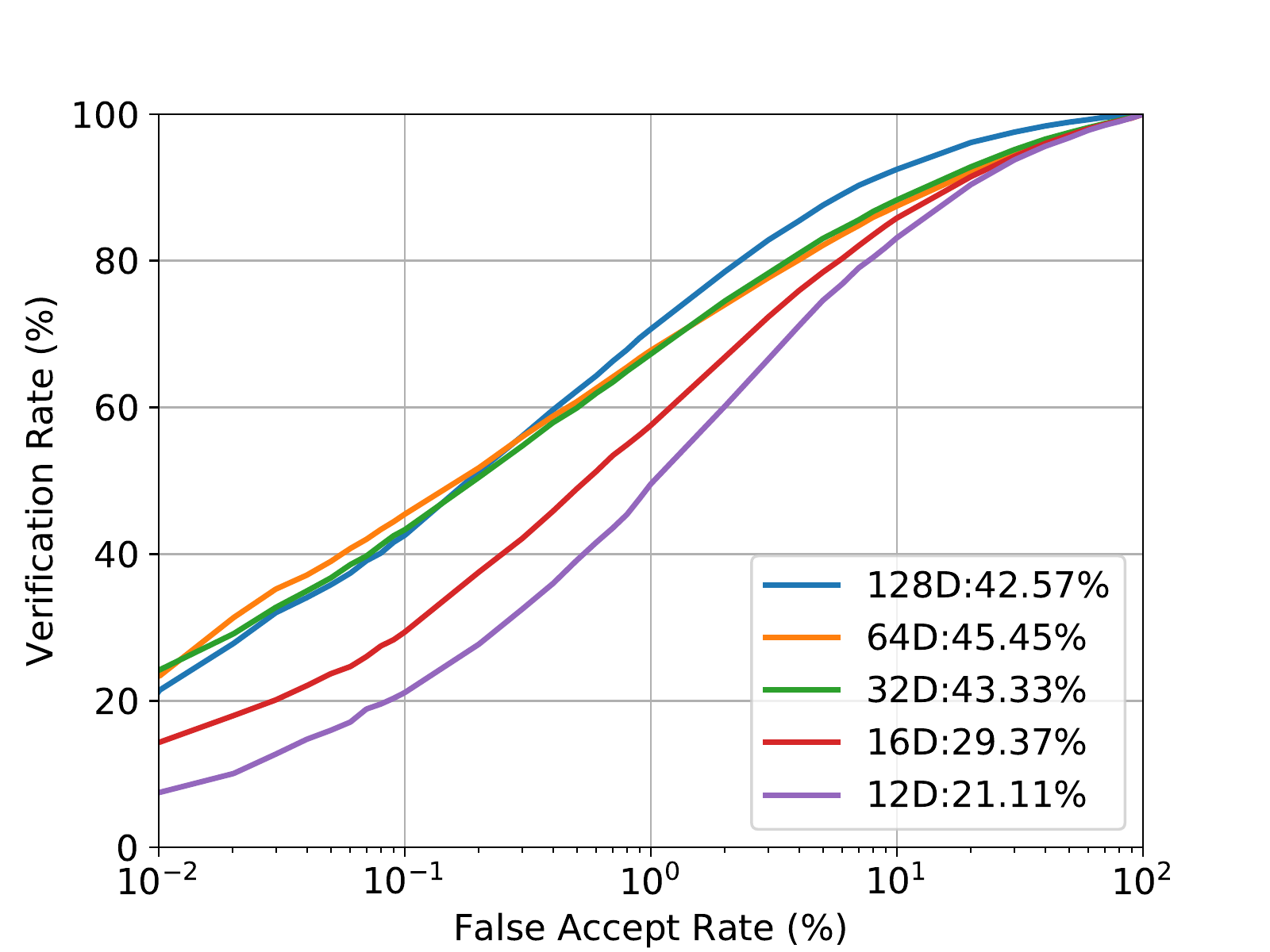}
        \caption{\tiny IJB-C: FaceNet-128}
    \end{subfigure}
    \begin{subfigure}[t]{0.16\textwidth}
        \centering
        \includegraphics[width=\textwidth]{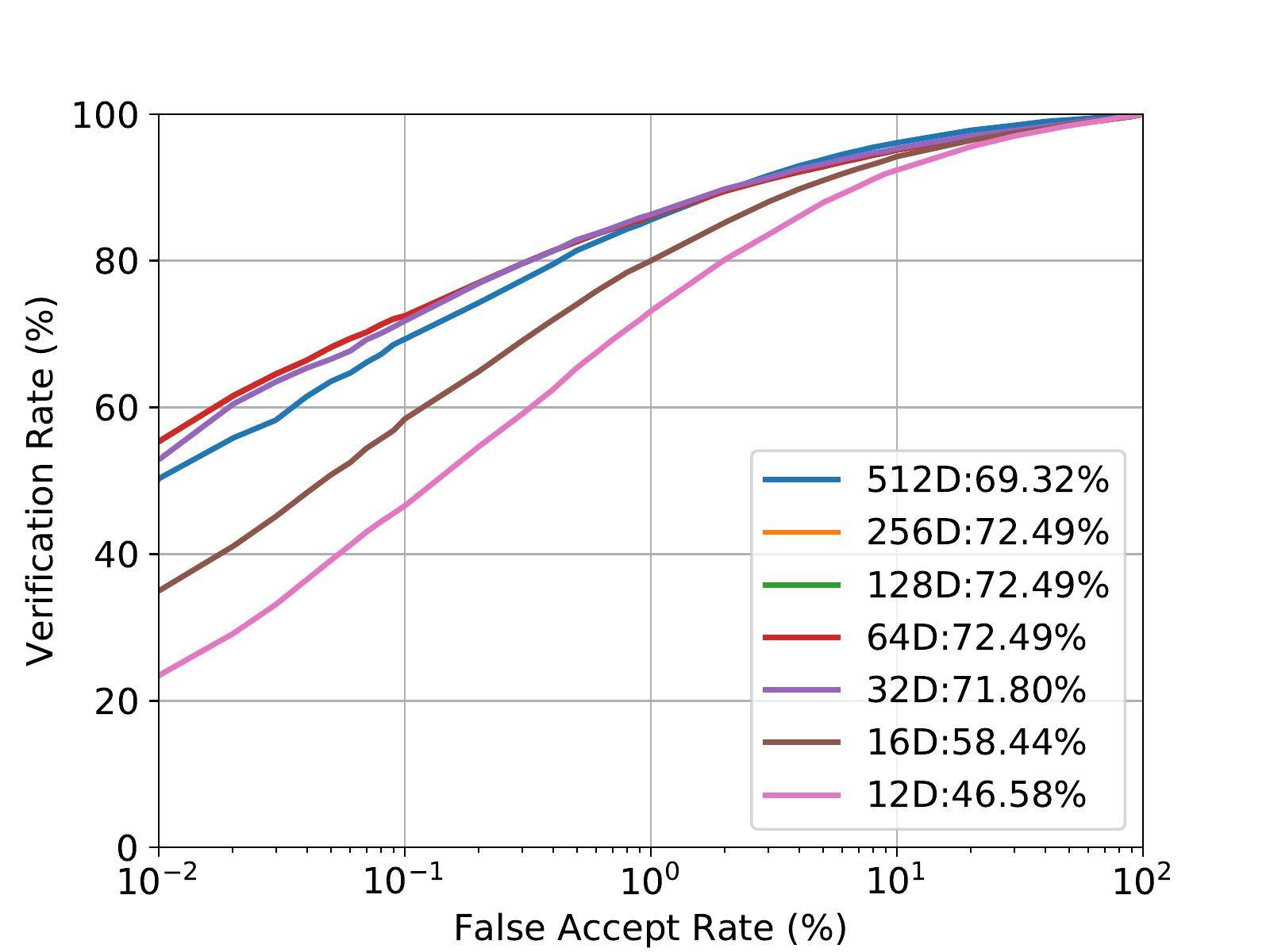}
        \caption{\tiny IJB-C: FaceNet-512}
    \end{subfigure}
    \begin{subfigure}[t]{0.16\textwidth}
        \centering
        \includegraphics[width=\textwidth]{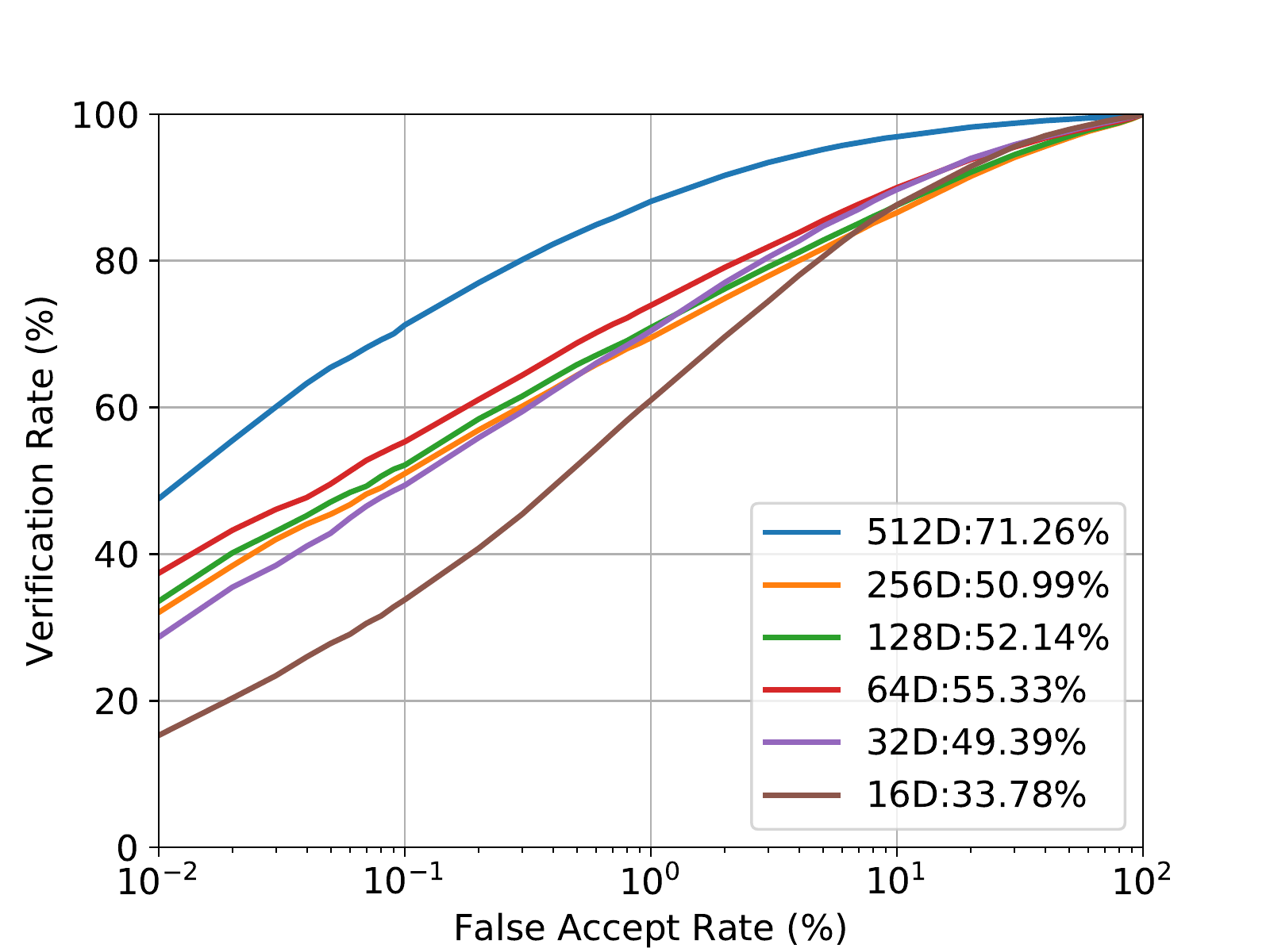}
        \caption{\tiny IJB-C: SphereFace}
    \end{subfigure}
    \begin{subfigure}[t]{0.16\textwidth}
        \centering
        \includegraphics[width=\textwidth]{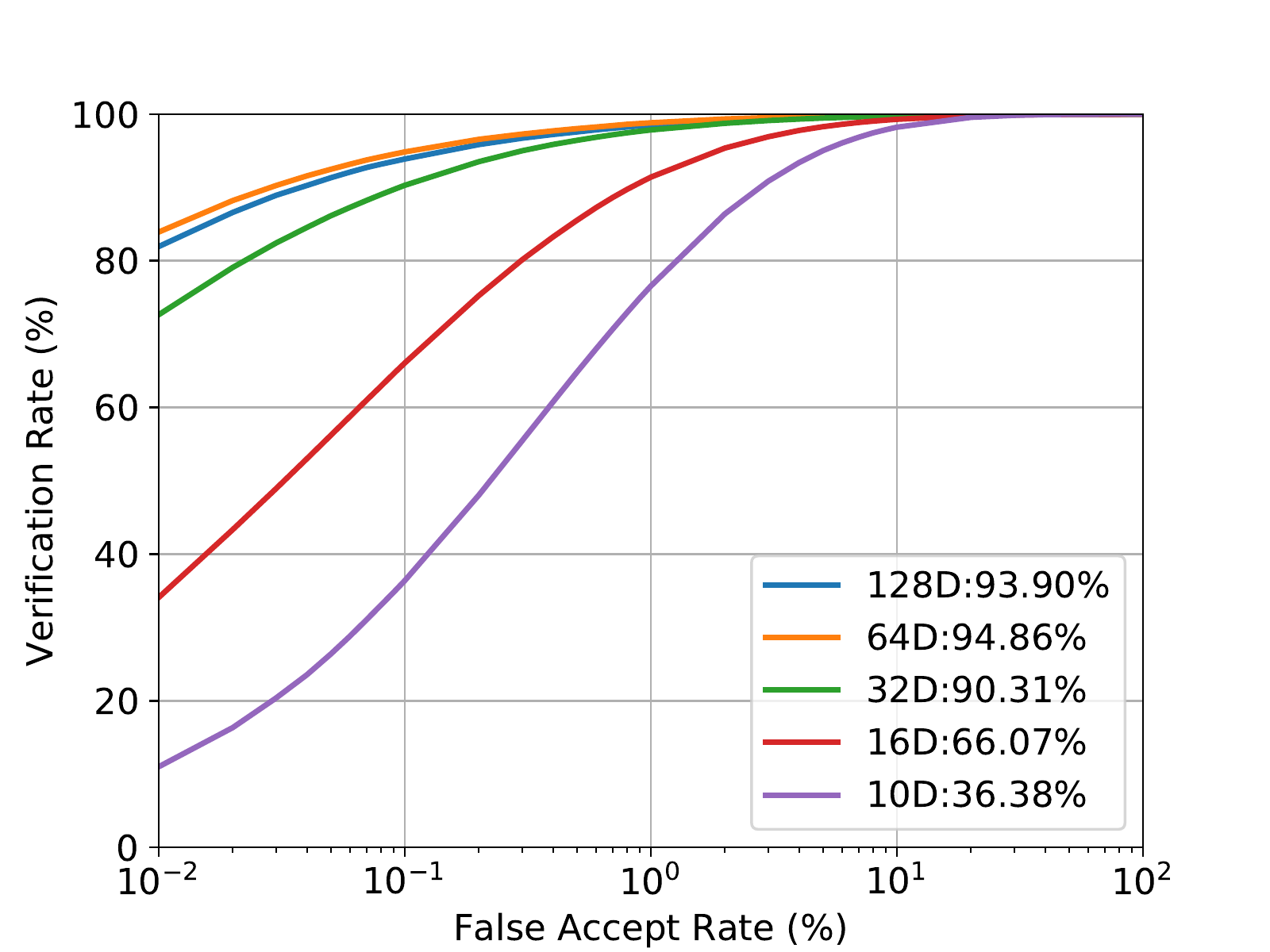}
        \caption{\tiny LFW: FaceNet-128}
    \end{subfigure}
    \begin{subfigure}[t]{0.16\textwidth}
        \centering
        \includegraphics[width=\textwidth]{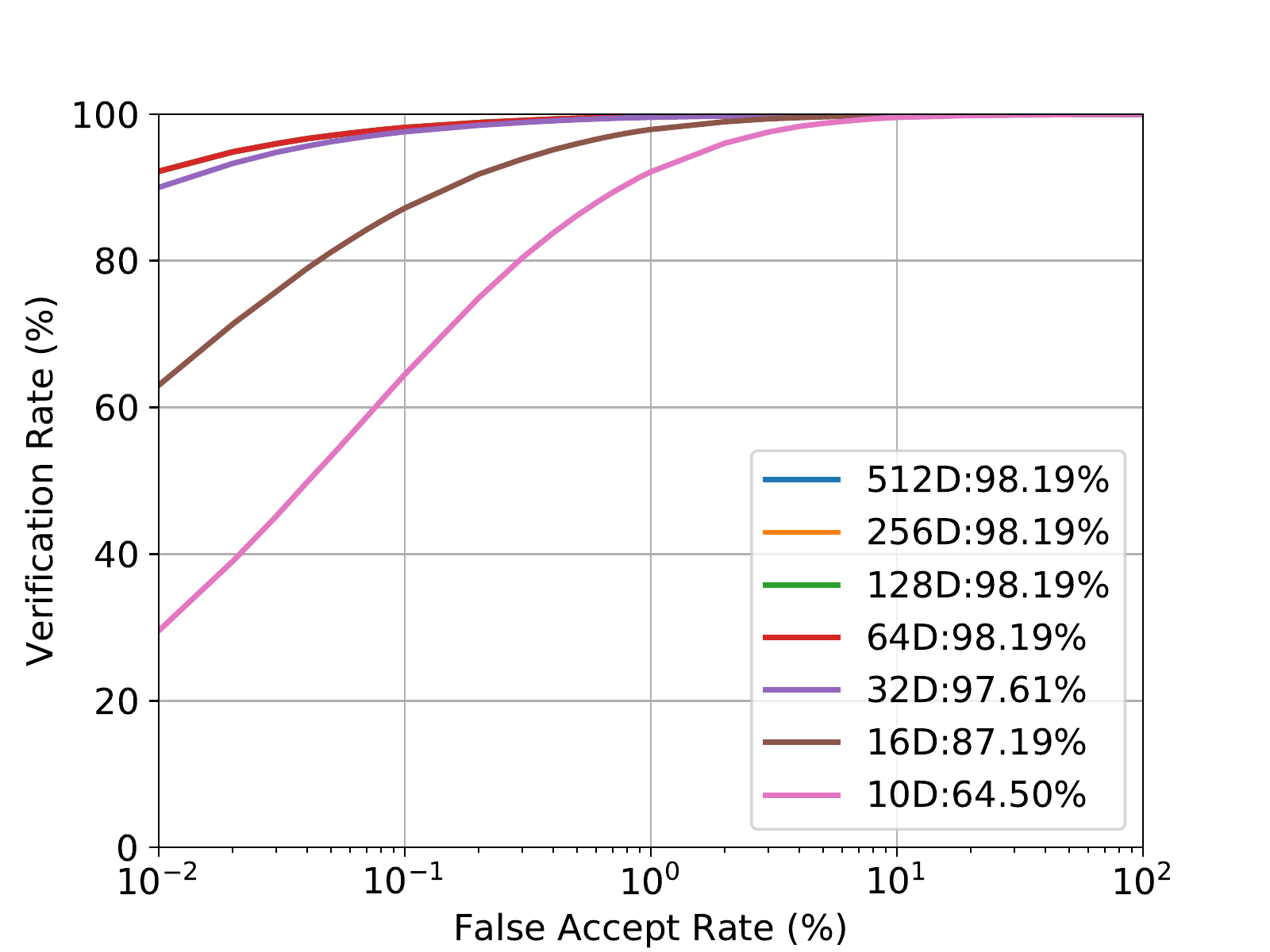}
        \caption{\tiny LFW: FaceNet-512}
    \end{subfigure}
    \begin{subfigure}[t]{0.16\textwidth}
        \centering
        \includegraphics[width=\textwidth]{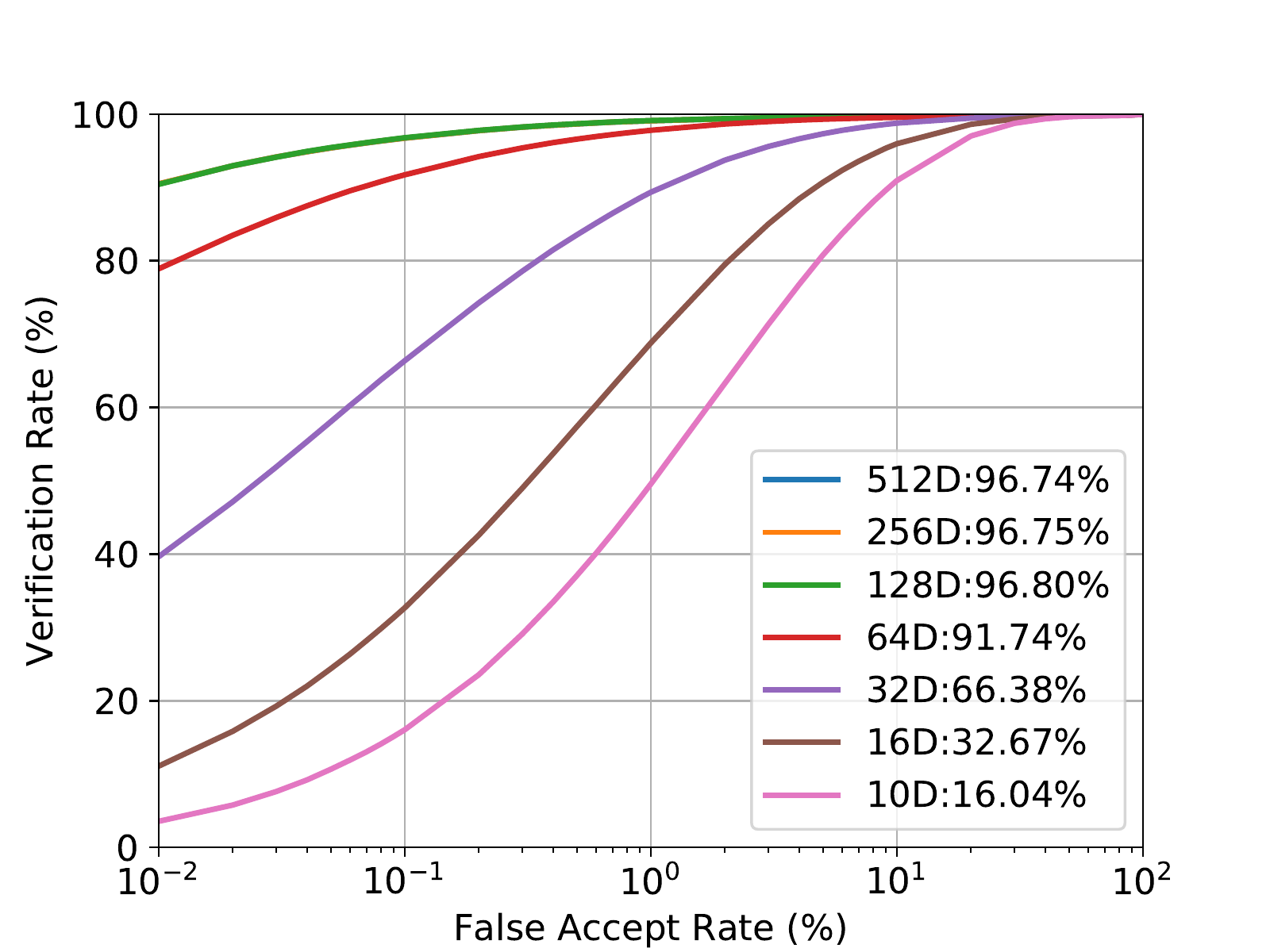}
        \caption{\tiny LFW: SphereFace}
    \end{subfigure}
    \caption{\textbf{PCA:} Face Verification on IJB-C and LFW (BLUFR) dataset for the (a) FaceNet-128, (b) FaceNet-512 and (c) SphereFace embeddings.}
    \label{fig:pca}
\end{figure*}

\begin{figure*}[t]
    \centering
    \begin{subfigure}[t]{0.16\textwidth}
        \centering
        \includegraphics[width=\textwidth]{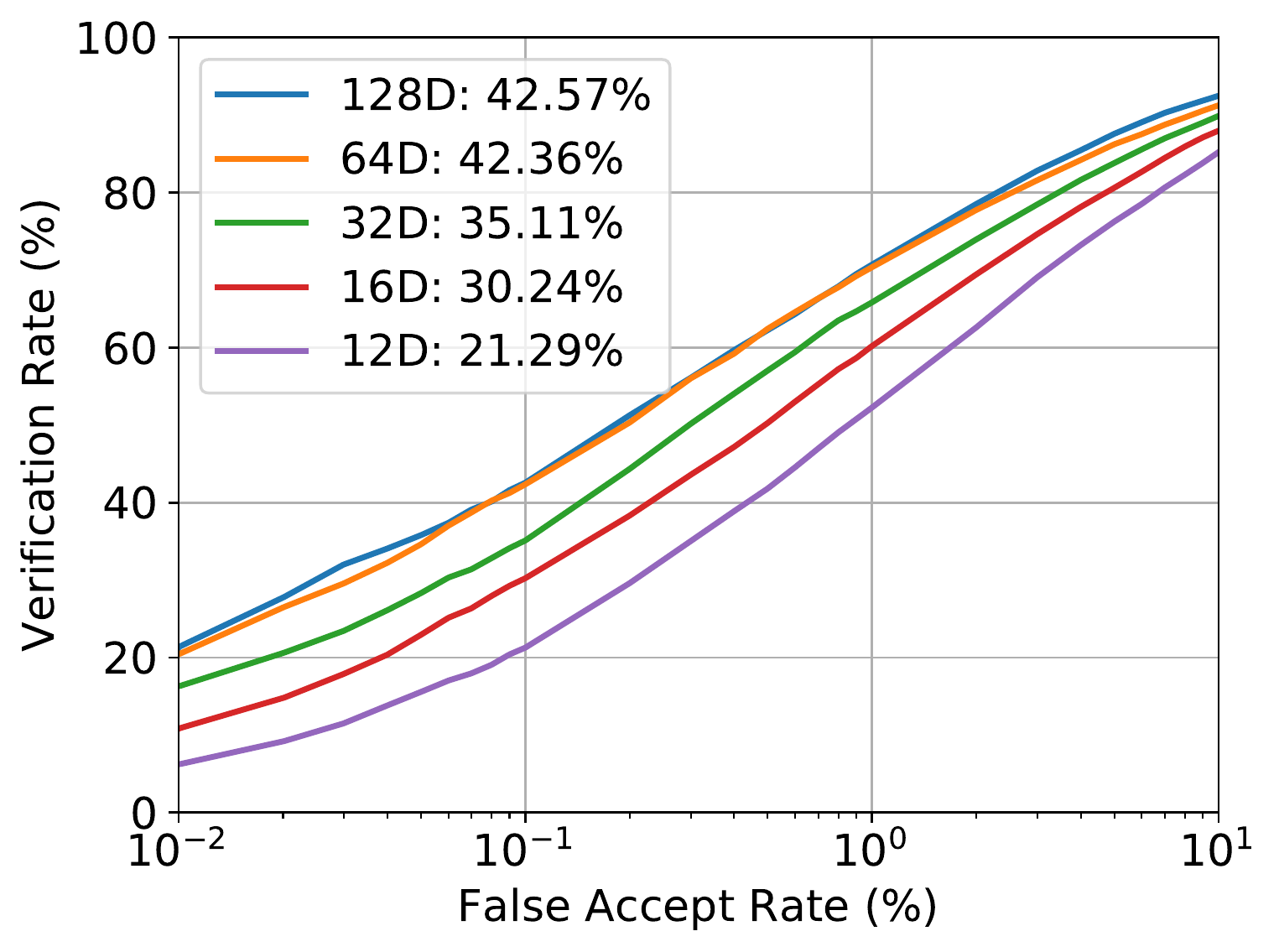}
        \caption{\tiny IJB-C: FaceNet-128}
    \end{subfigure}
    \begin{subfigure}[t]{0.16\textwidth}
        \centering
        \includegraphics[width=\textwidth]{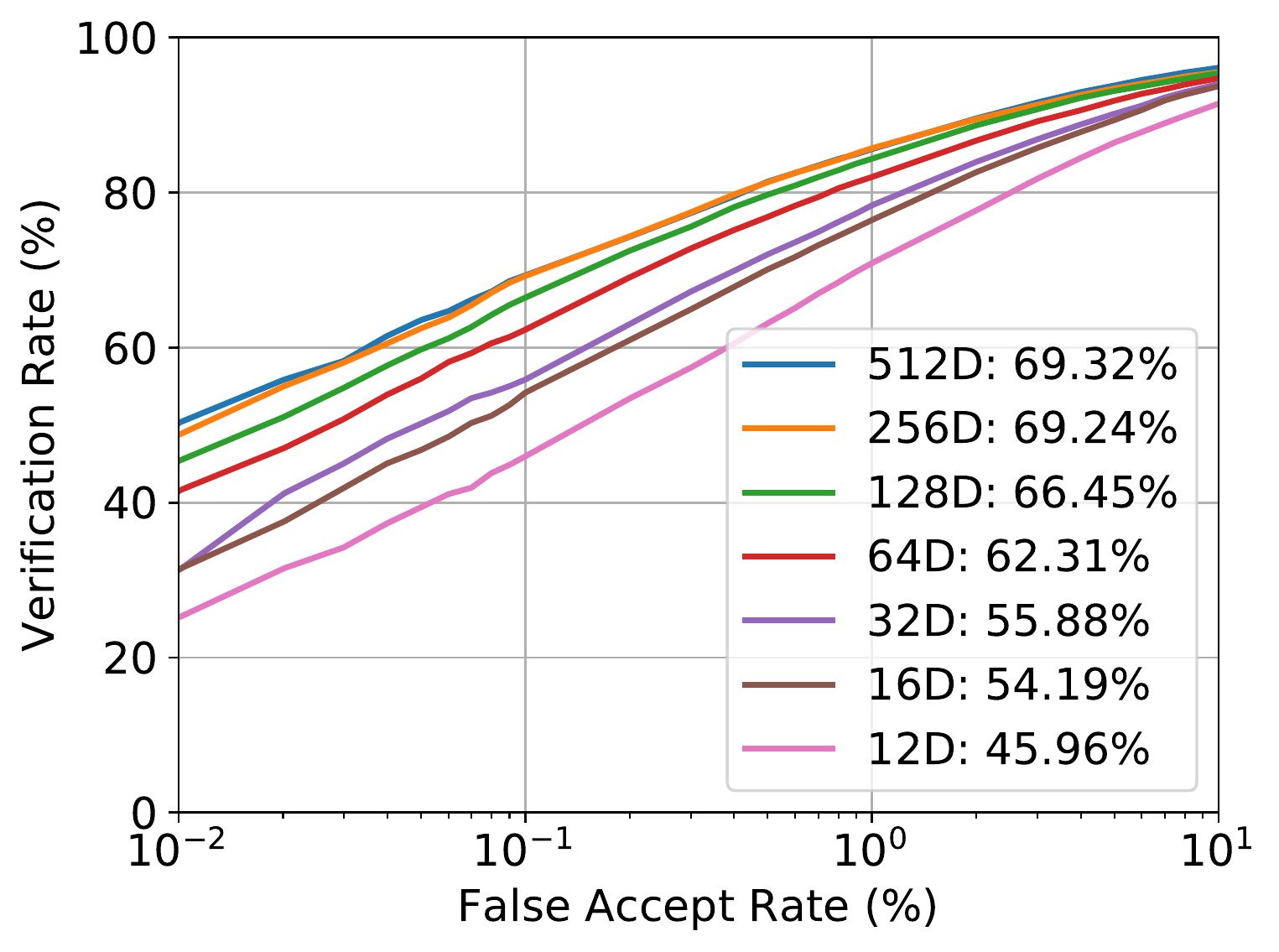}
        \caption{\tiny IJB-C: FaceNet-512}
    \end{subfigure}
    \begin{subfigure}[t]{0.16\textwidth}
        \centering
        \includegraphics[width=\textwidth]{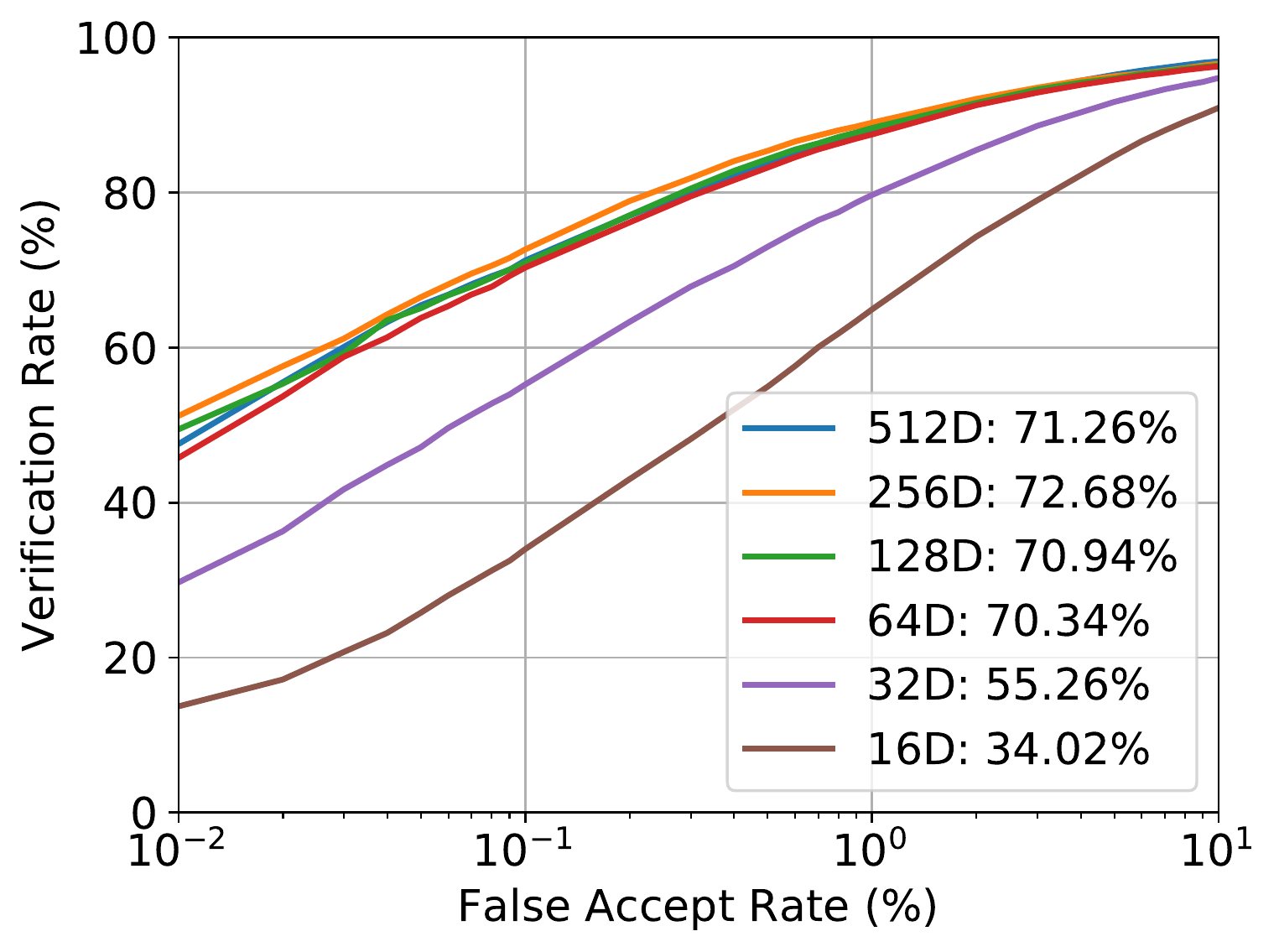}
        \caption{\tiny IJB-C: SphereFace}
    \end{subfigure}
    \begin{subfigure}[t]{0.16\textwidth}
        \centering
        \includegraphics[width=\textwidth]{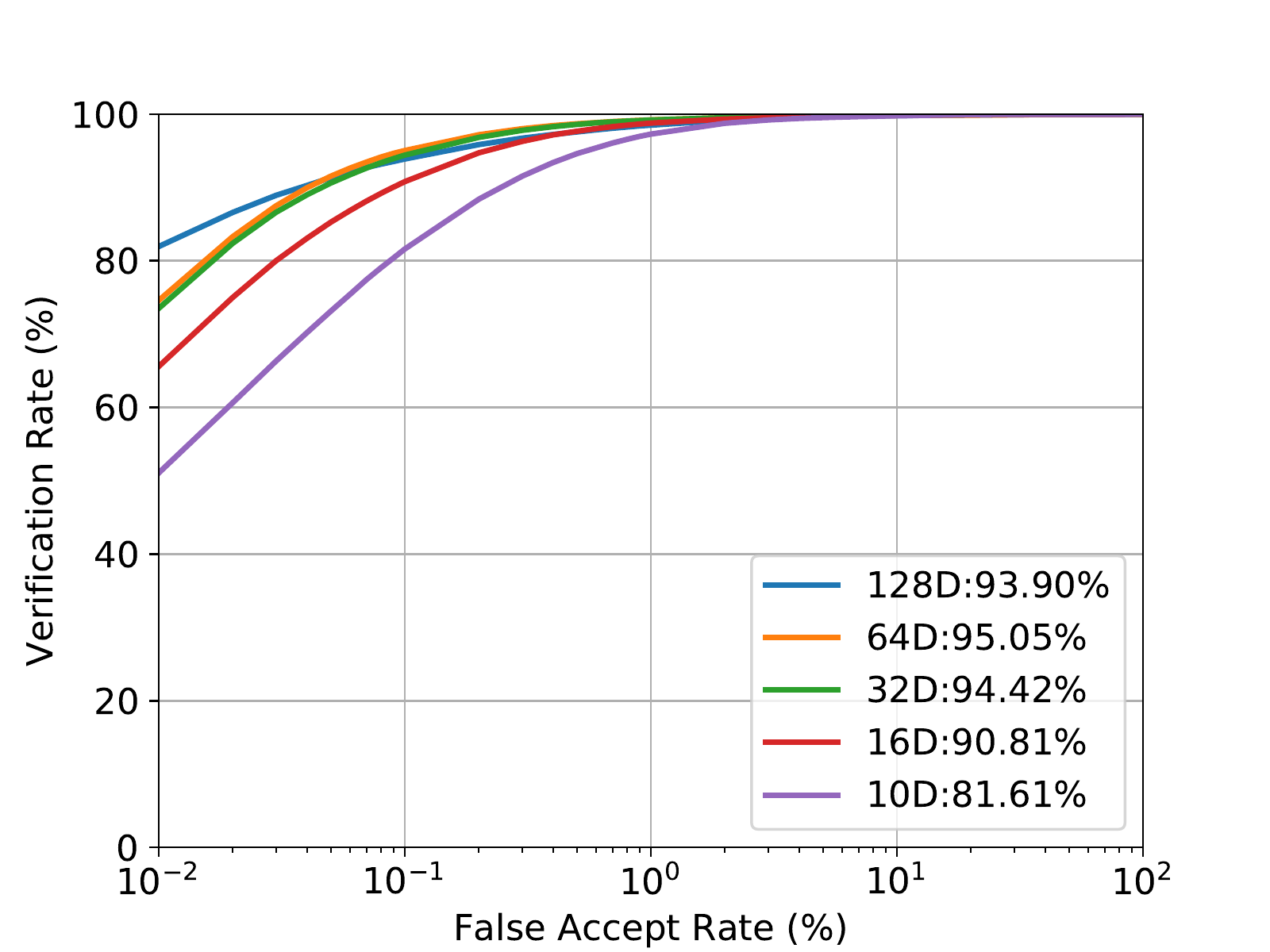}
        \caption{\tiny LFW: FaceNet-128}
    \end{subfigure}
    \begin{subfigure}[t]{0.16\textwidth}
        \centering
        \includegraphics[width=\textwidth]{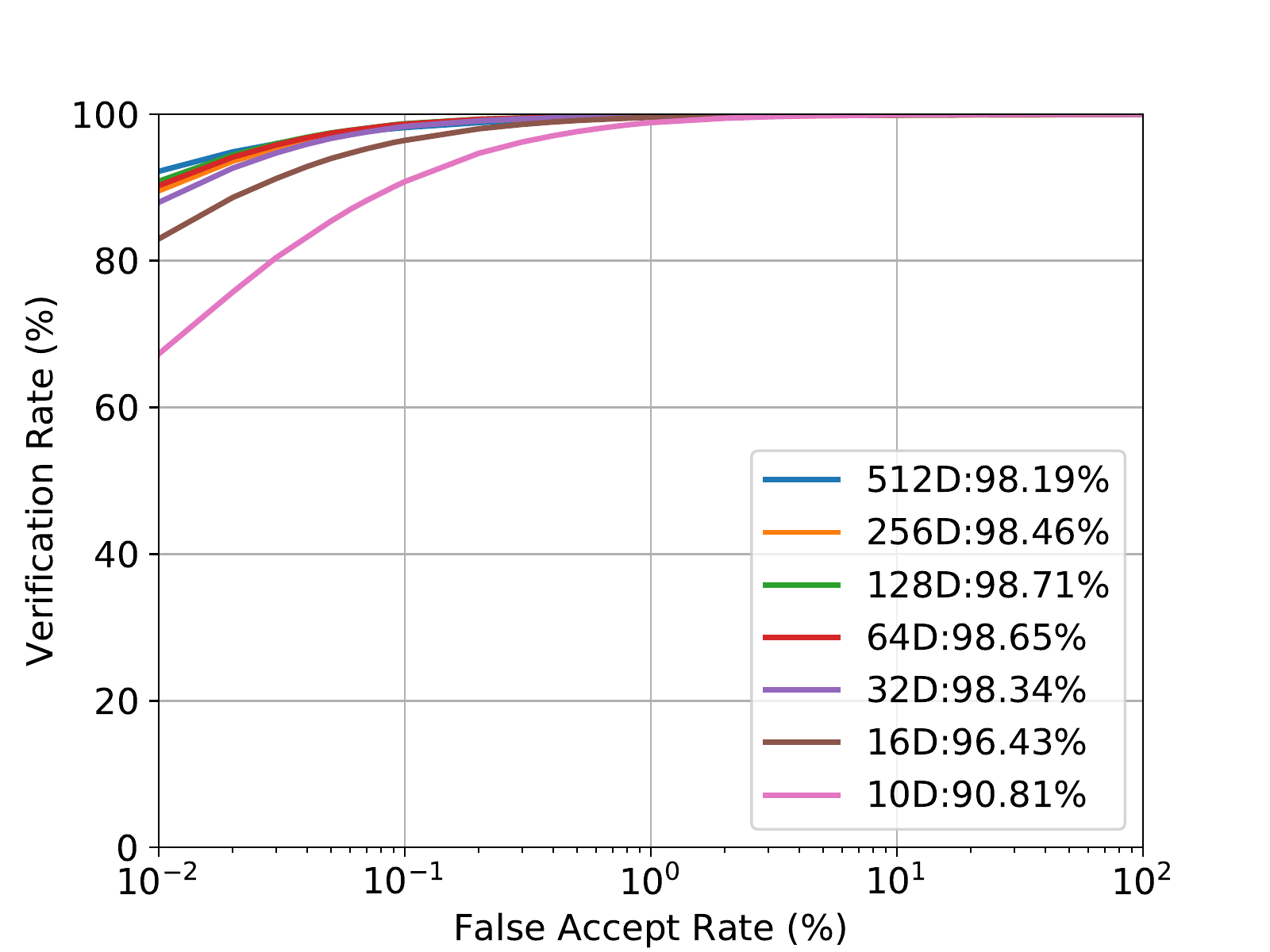}
        \caption{\tiny LFW: FaceNet-512}
    \end{subfigure}
    \begin{subfigure}[t]{0.16\textwidth}
        \centering
        \includegraphics[width=\textwidth]{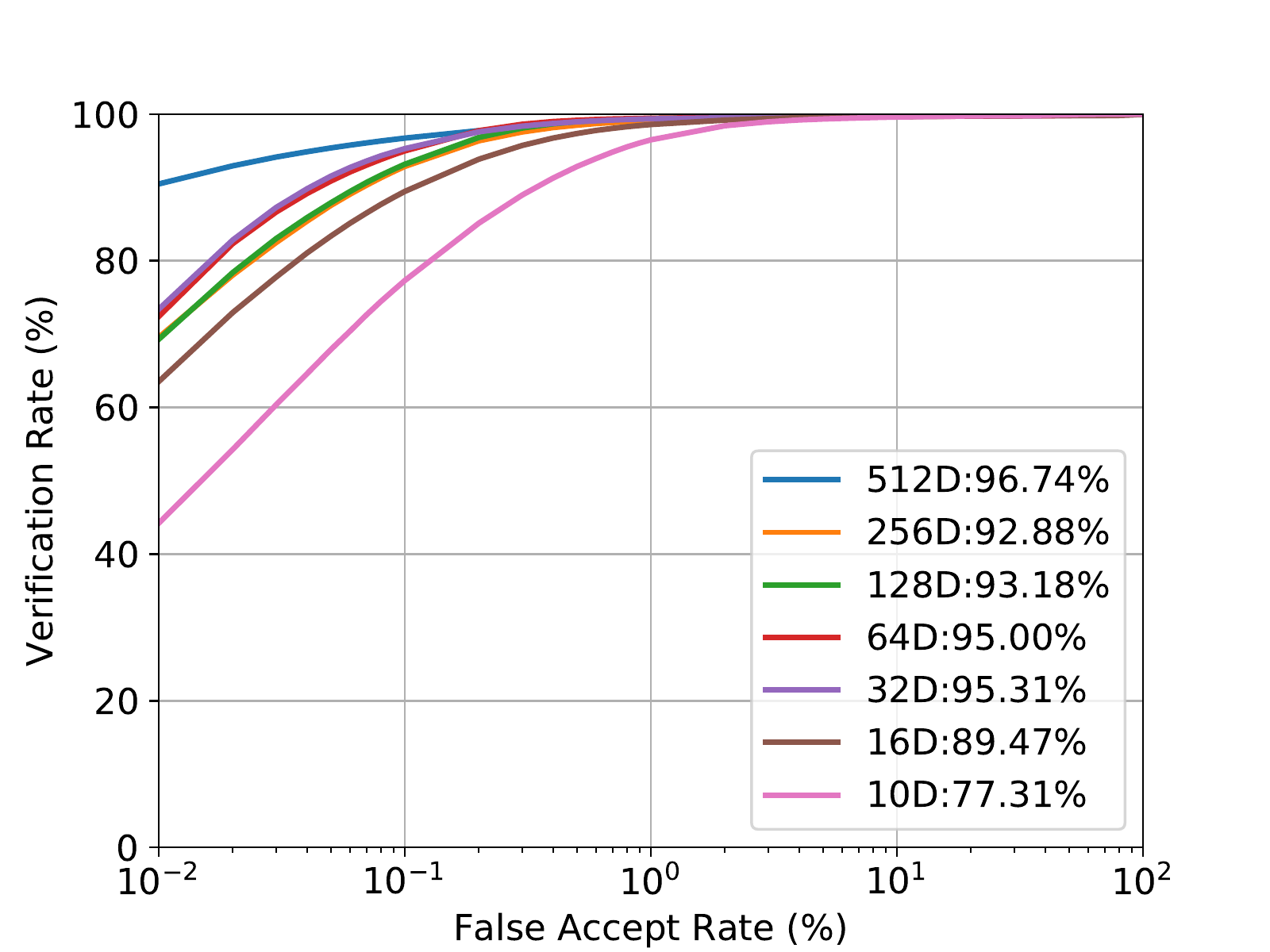}
        \caption{\tiny LFW: SphereFace}
    \end{subfigure}
    \caption{\textbf{Isomap:} Face Verification on IJB-C and LFW (BLUFR) dataset for the (a) FaceNet-128, (b) FaceNet-512 and (c) SphereFace embeddings.}
    \label{fig:isomap}
\end{figure*}

\begin{figure*}[t]
    \centering
    \begin{subfigure}[t]{0.16\textwidth}
        \centering
        \includegraphics[width=\textwidth]{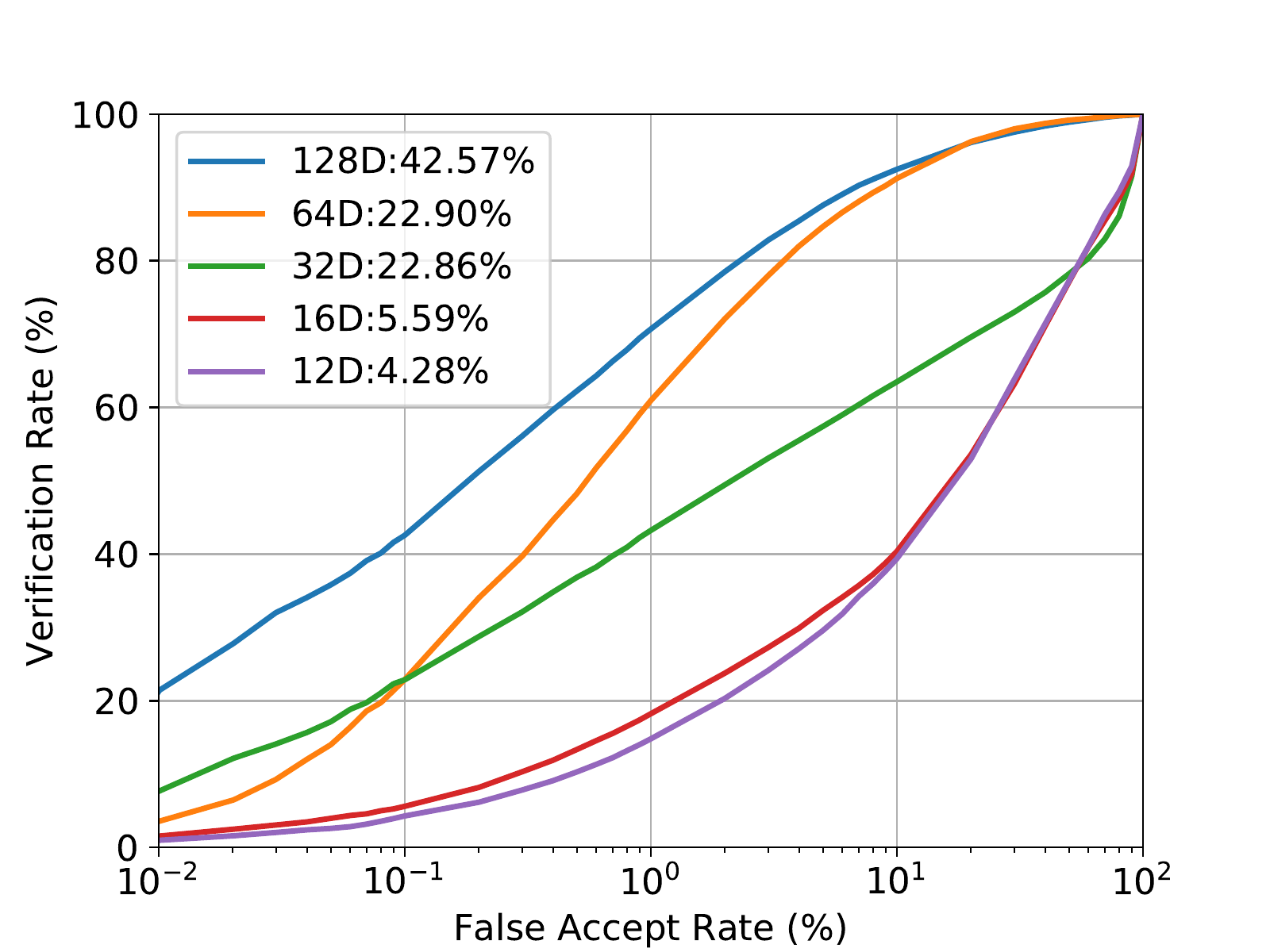}
        \caption{FaceNet-128}
    \end{subfigure}
    \begin{subfigure}[t]{0.16\textwidth}
        \centering
        \includegraphics[width=\textwidth]{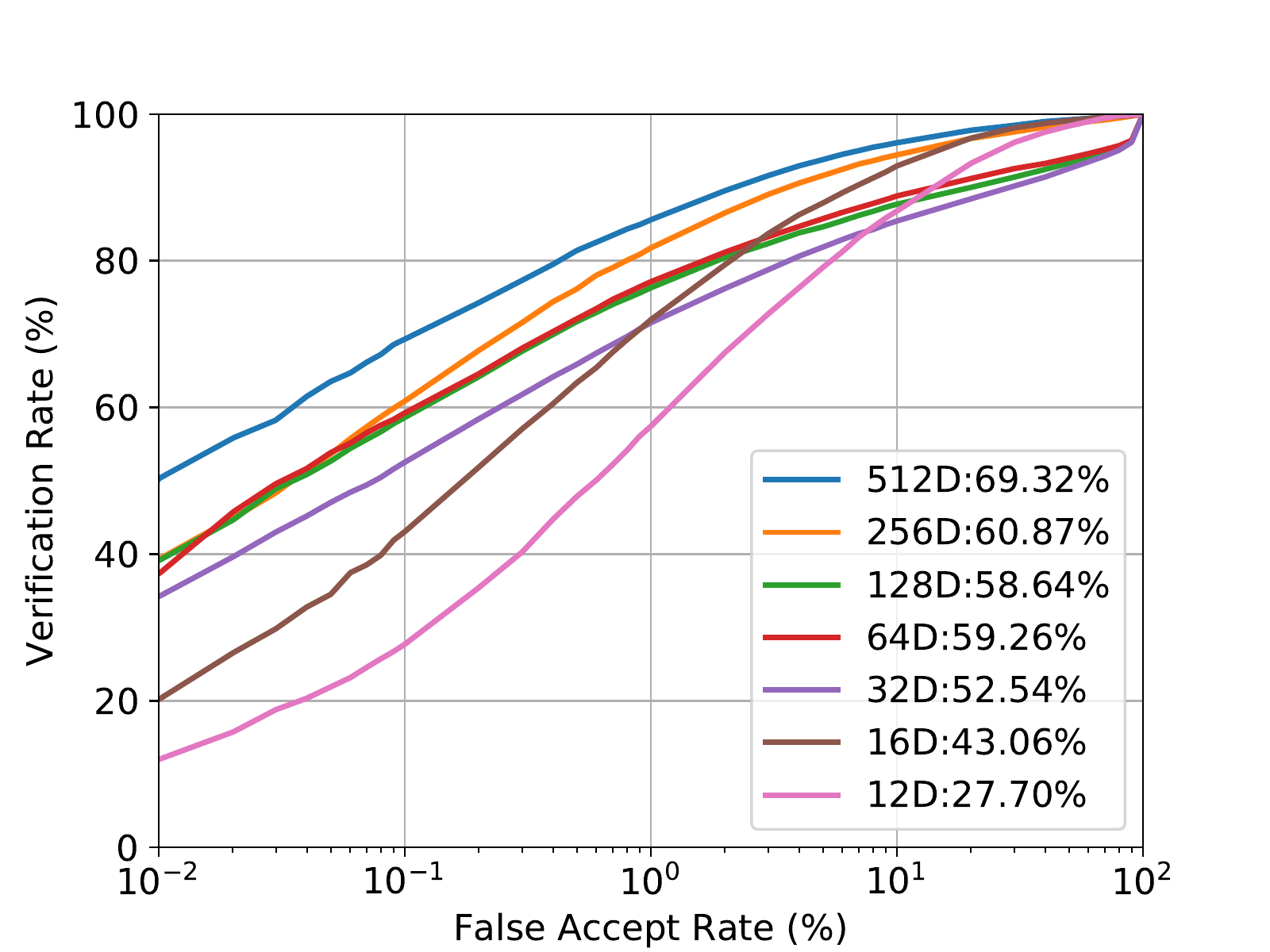}
        \caption{FaceNet-512}
    \end{subfigure}
    \begin{subfigure}[t]{0.16\textwidth}
        \centering
        \includegraphics[width=\textwidth]{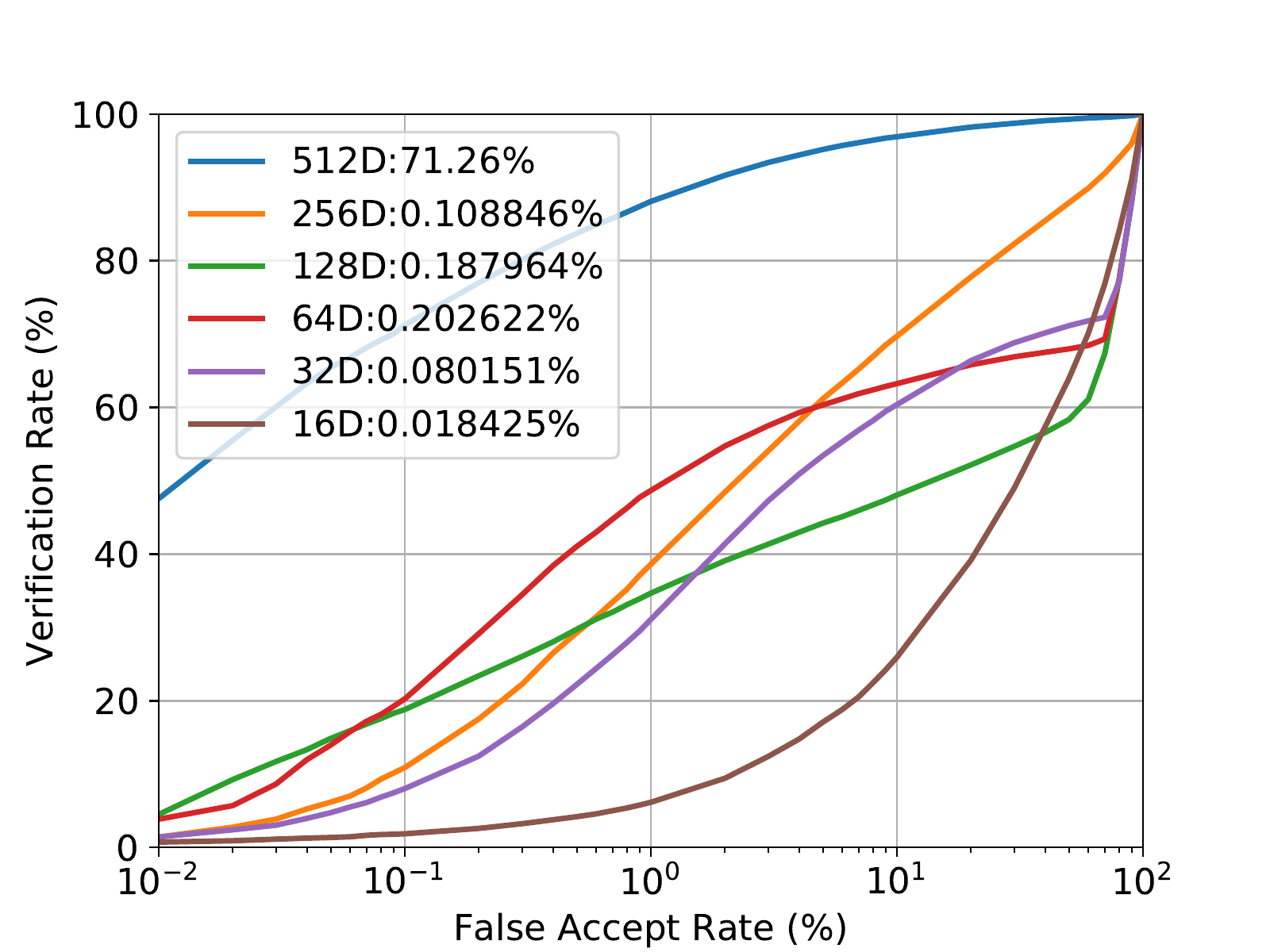}
        \caption{SphereFace}
    \end{subfigure}
    \begin{subfigure}[t]{0.16\textwidth}
        \centering
        \includegraphics[width=\textwidth]{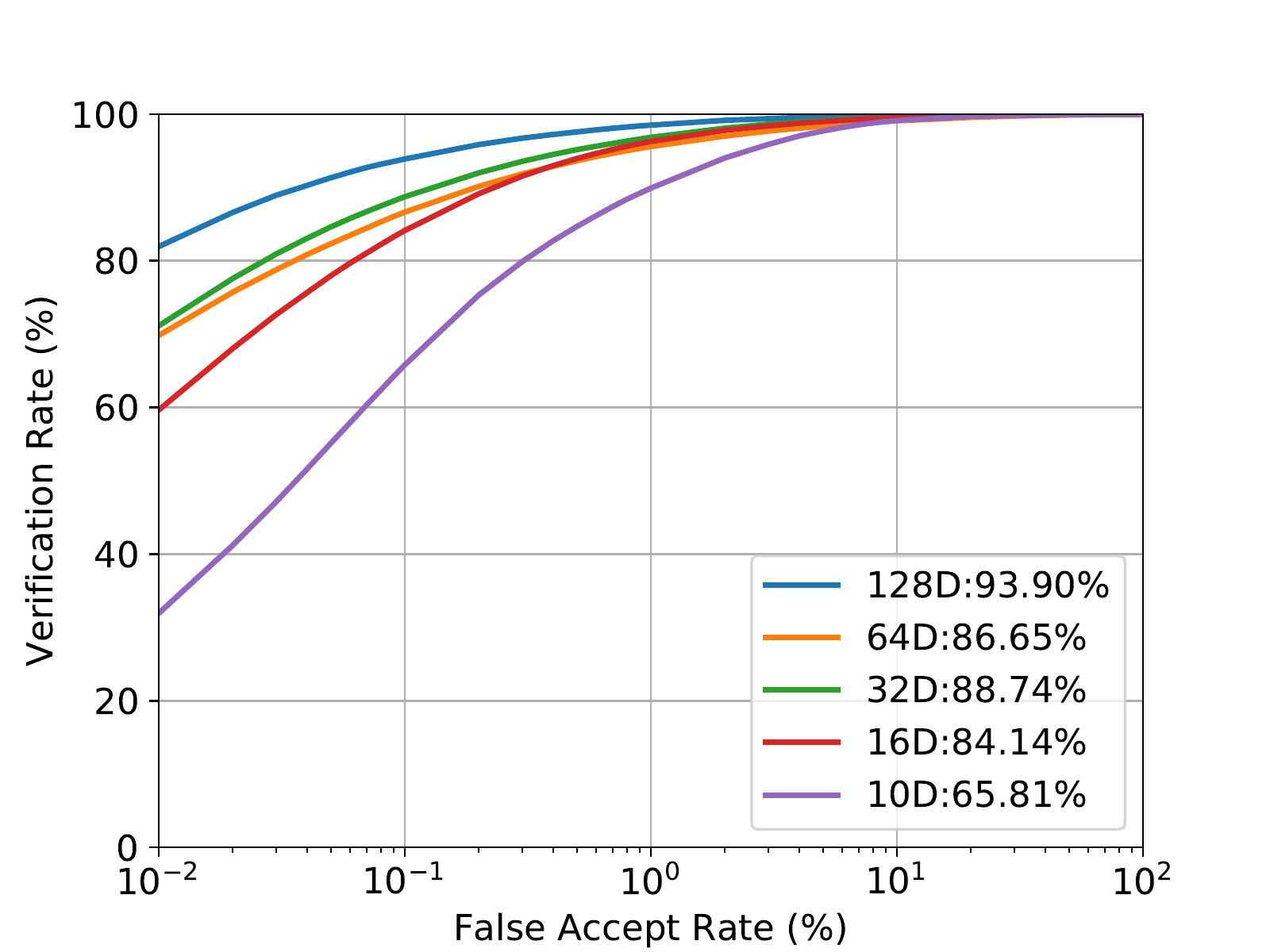}
        \caption{FaceNet-128}
    \end{subfigure}
    \begin{subfigure}[t]{0.16\textwidth}
        \centering
        \includegraphics[width=\textwidth]{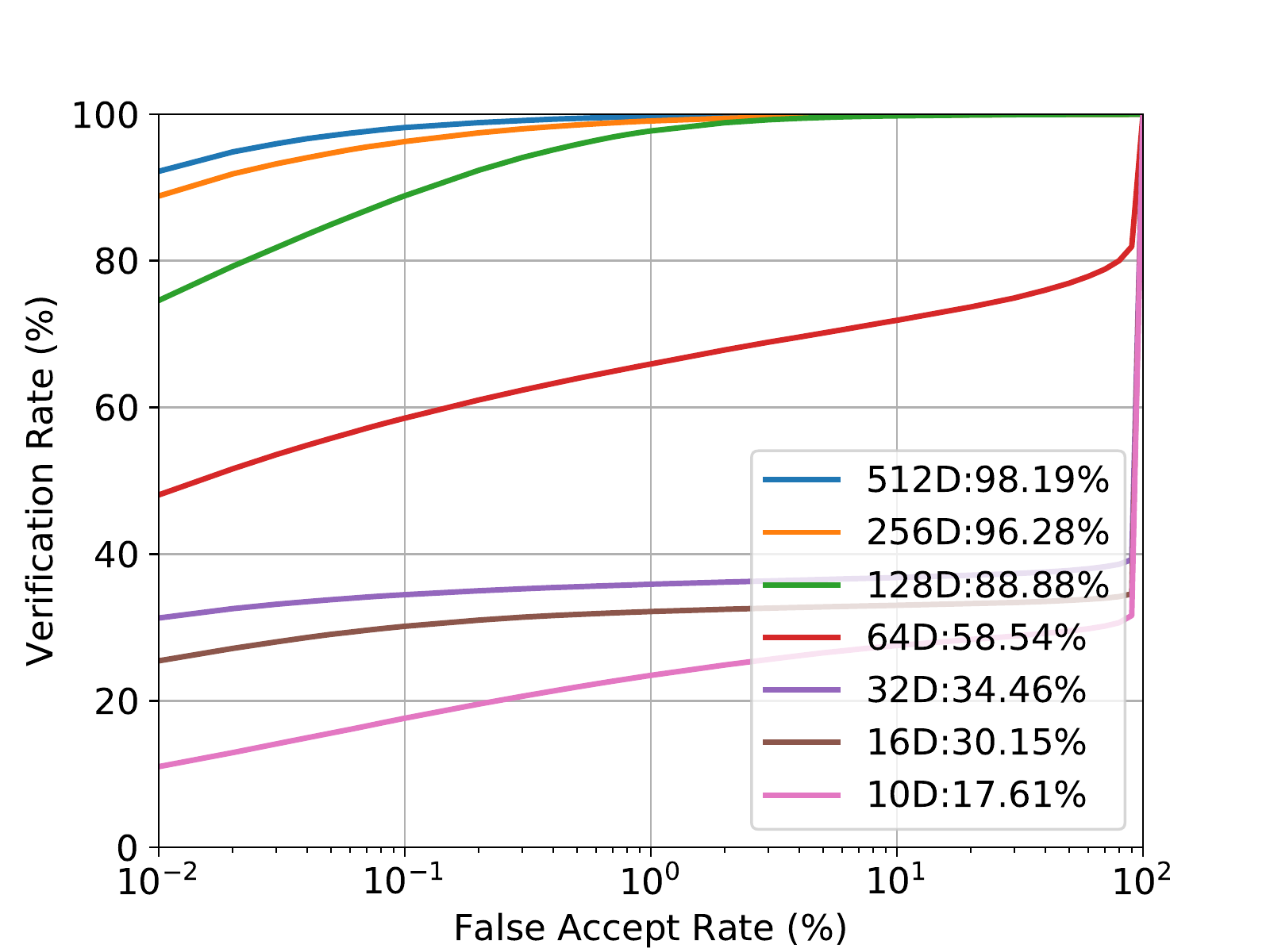}
        \caption{FaceNet-512}
    \end{subfigure}
    \begin{subfigure}[t]{0.16\textwidth}
        \centering
        \includegraphics[width=\textwidth]{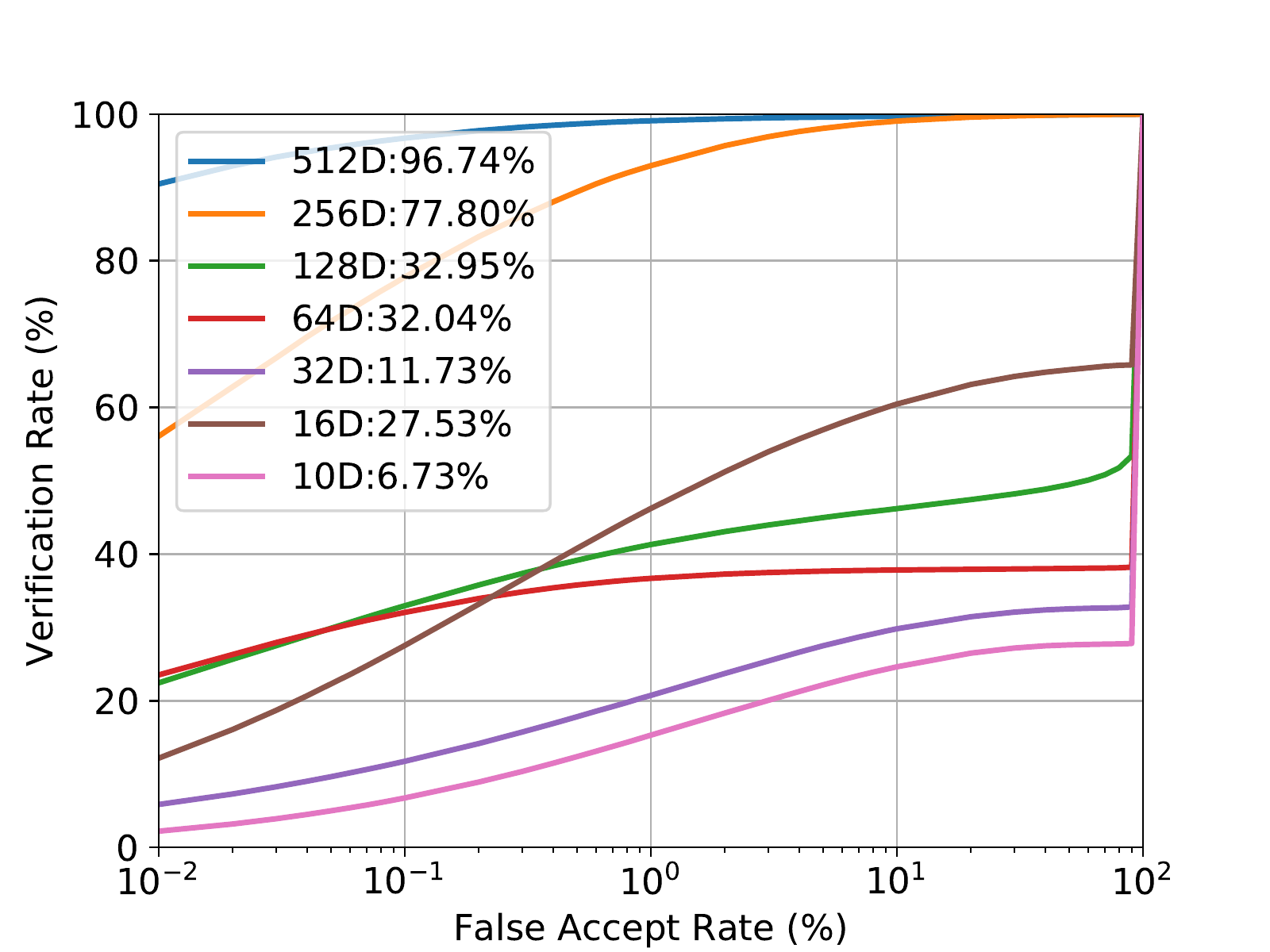}
        \caption{SphereFace}
    \end{subfigure}
    \caption{\textbf{Denoising Autoencoder:} Face Verification on IJB-C and LFW (BLUFR) dataset for the (a) FaceNet-128, (b) FaceNet-512 and (c) SphereFace embeddings.}
    \label{fig:auto}
\end{figure*}

In this section we present results of DeepMDS on LFW (BLUFR) dataset and the baseline dimensionality reduction methods for mapping from the ambient to the intrinsic space. Figure \ref{fig:deepmds} show the face verification ROC curves of DeepMDS on LFW dataset for FaceNet-128, FaceNet-512 and SphereFace representation models. Figure \ref{fig:pca} show the face verification ROC curves of Principal Component Analysis on the IJB-C and LFW (BLUFR) datasets for all the three representation models. Similarly, Fig. \ref{fig:isomap} and Fig. \ref{fig:auto} show the face verification ROC curves of the Isomap and Denoising Autoencoder baselines, respectively.


\subsection{Intrinsic Dimensionality Estimation (Derivations) \label{sec:derivations}}
We first show the derivation for estimating the intrinsic dimensionality $m$ that minimizes the RMSE with respect to a $m$-hypersphere,
\begin{equation}
    \footnotesize{
    \begin{aligned}
    \min_{m} \mbox{ } \int_{r_{max}-2\sigma}^{r_{max}}\left\|\log \frac{\hat{p}_{\mathcal{M}}(r)}{\hat{p}_{\mathcal{M}}(r_{max})} - (m-1)\log\left(\sin\left[\frac{\pi r}{2r_{max}}\right]\right)\right\|^2 \nonumber
    \end{aligned}}
    \label{eq:id}
\end{equation}

First we estimate $\sigma$ for the $m$-hypersphere by approximating the distribution $\hat{p}_{\mathcal{M}}(r)$ by a univariate Gaussian distribution around the mode of $p_{\mathcal{M}}(r)$. So, given samples $S=\{r_1,\dots,r_T\}$ from the distribution $p(r)$, the variance around the mode can be estimated as, $\sigma^2 = \frac{1}{T}\sum_{t=1}^T (r_t-r_{max})^2$, where $r_{max}$ is the radius at the mode of $\hat{p}_{\mathcal{M}}(r)$. Then, we estimate the distribution $\log \frac{\hat{p}_{\mathcal{M}}(r)}{\hat{p}_{\mathcal{M}}(r_{max})}$ vs $\log\left(sin\left[\frac{\pi r}{2r_{max}}\right]\right)$ and solve the following least-squares fit problem:
\begin{equation}
    \footnotesize{
    \begin{aligned}
    \min_{m} \mbox{ } \sum_{S \cap  r_{max} -2\sigma \leq r_i \leq r_{max}}\left(y_i - (m-1)x_i\right)^2 \nonumber
    \end{aligned}}
    \label{eq:id}
\end{equation}
\noindent where $y_i=\log\frac{\hat{p}\_{\mathcal{M}}(r_i)}{\hat{p}\_{\mathcal{M}}(r_{max})}$ and $x_i=\log\left(sin\left[\frac{\pi r}{2r_{max}}\right]\right)$.

In the case of comparison to a Gaussian distribution, the intrinsic dimensionality can also be estimated by comparing to the geodesic distance distribution for points sampled from a Gaussian distribution as,
\begin{equation}
    \min_{d} \int_{r_{max}-2\sigma}^{r_{max}}\left\|\log \frac{p(r)}{p(r_{max})} + (d-1)\frac{r^2}{4\sigma^2}\right\|_2^2
\end{equation}
\noindent The solution of this optimization problem can be found following the same procedure described above for a $m$-hypersphere.

\subsection{Intrinsic Dimensionality Estimation Fitting \label{sec:rmse}}
\begin{figure*}
    \begin{subfigure}[t]{0.23\textwidth}
        \centering
        \includegraphics[width=\textwidth]{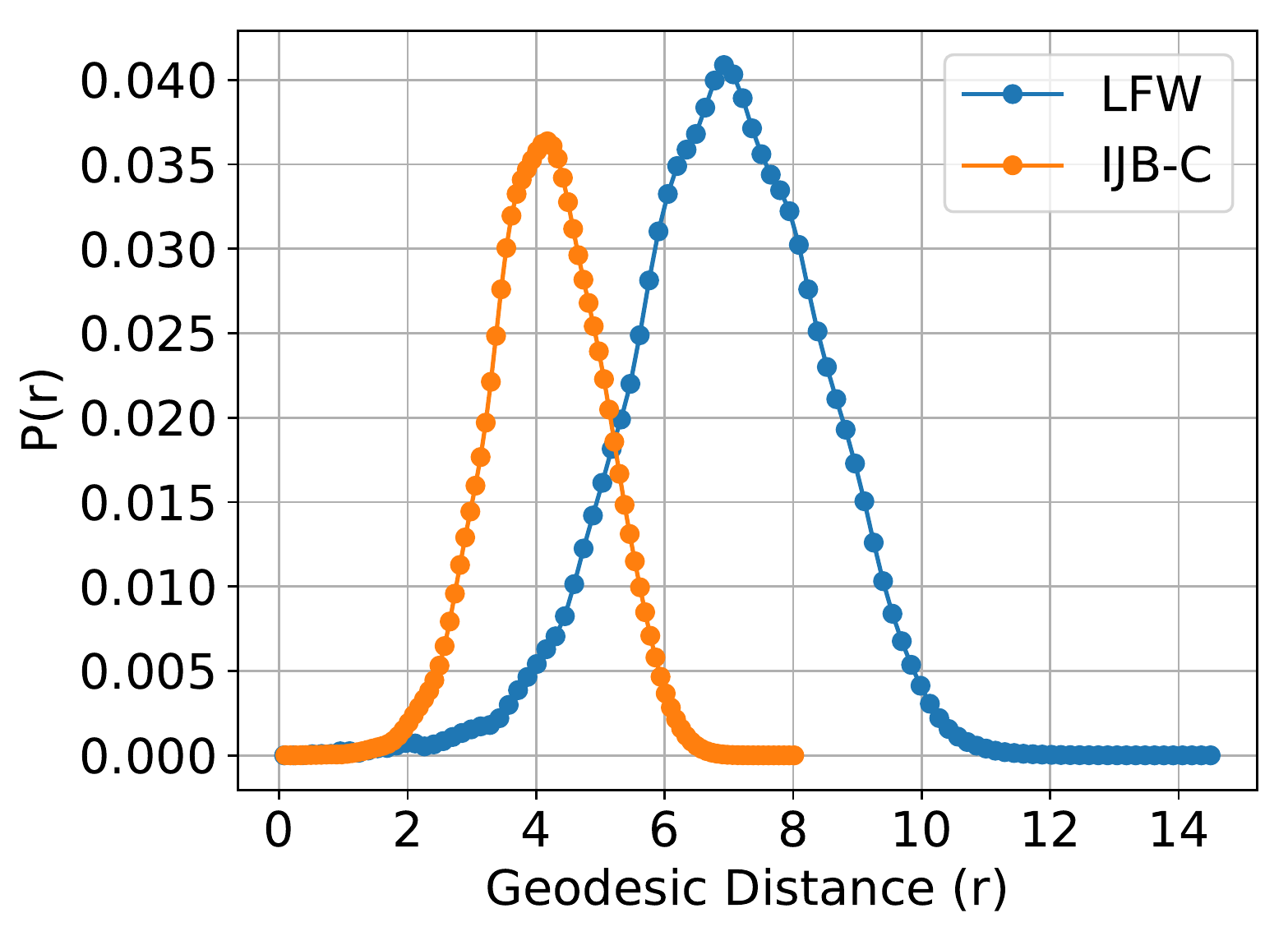}
        \caption{FaceNet-128}
    \end{subfigure}
    \begin{subfigure}[t]{0.23\textwidth}
        \centering
        \includegraphics[width=\textwidth]{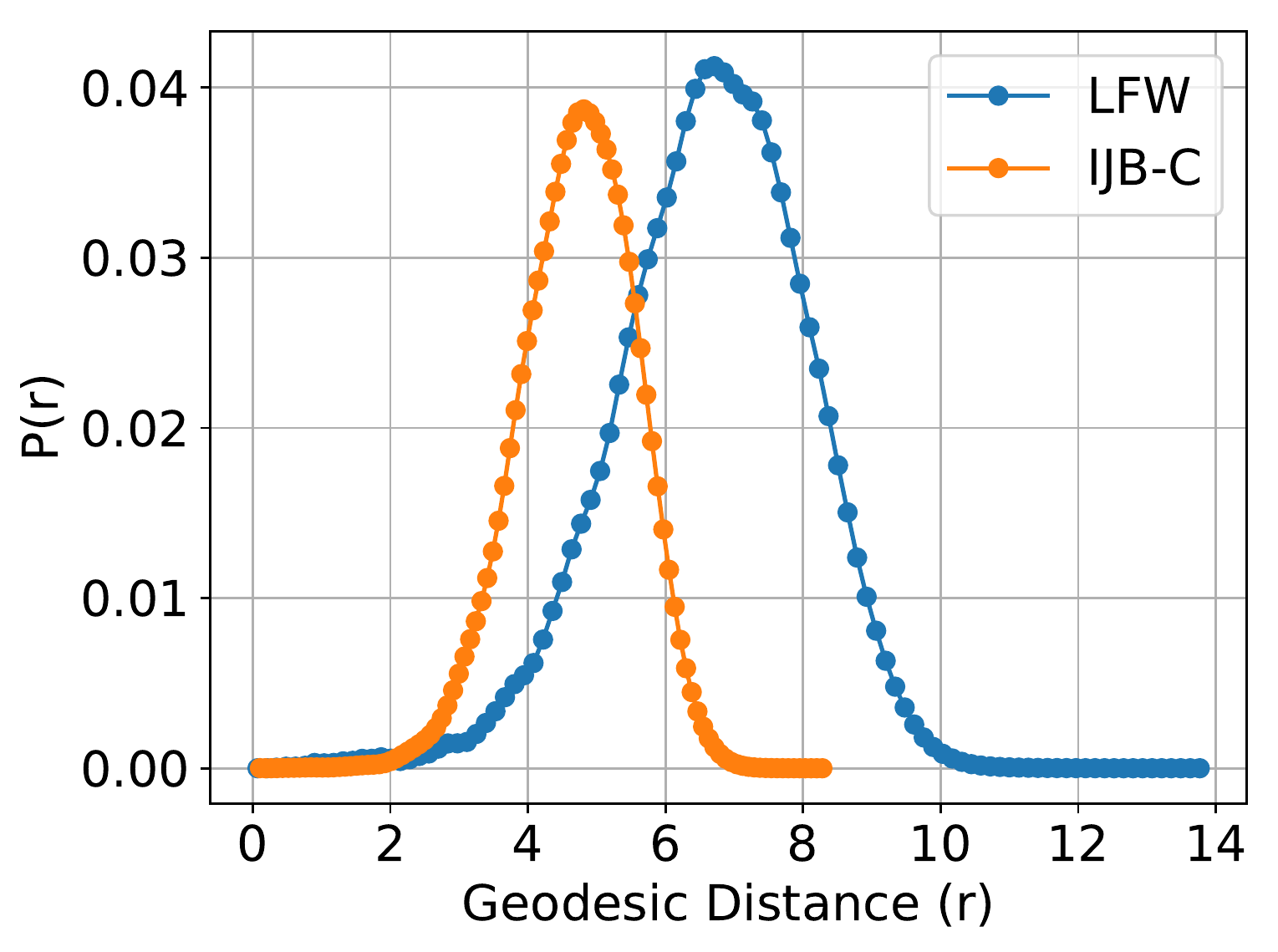}
        \caption{FaceNet-512}
    \end{subfigure}
    \begin{subfigure}[t]{0.23\textwidth}
        \centering
        \includegraphics[width=\textwidth]{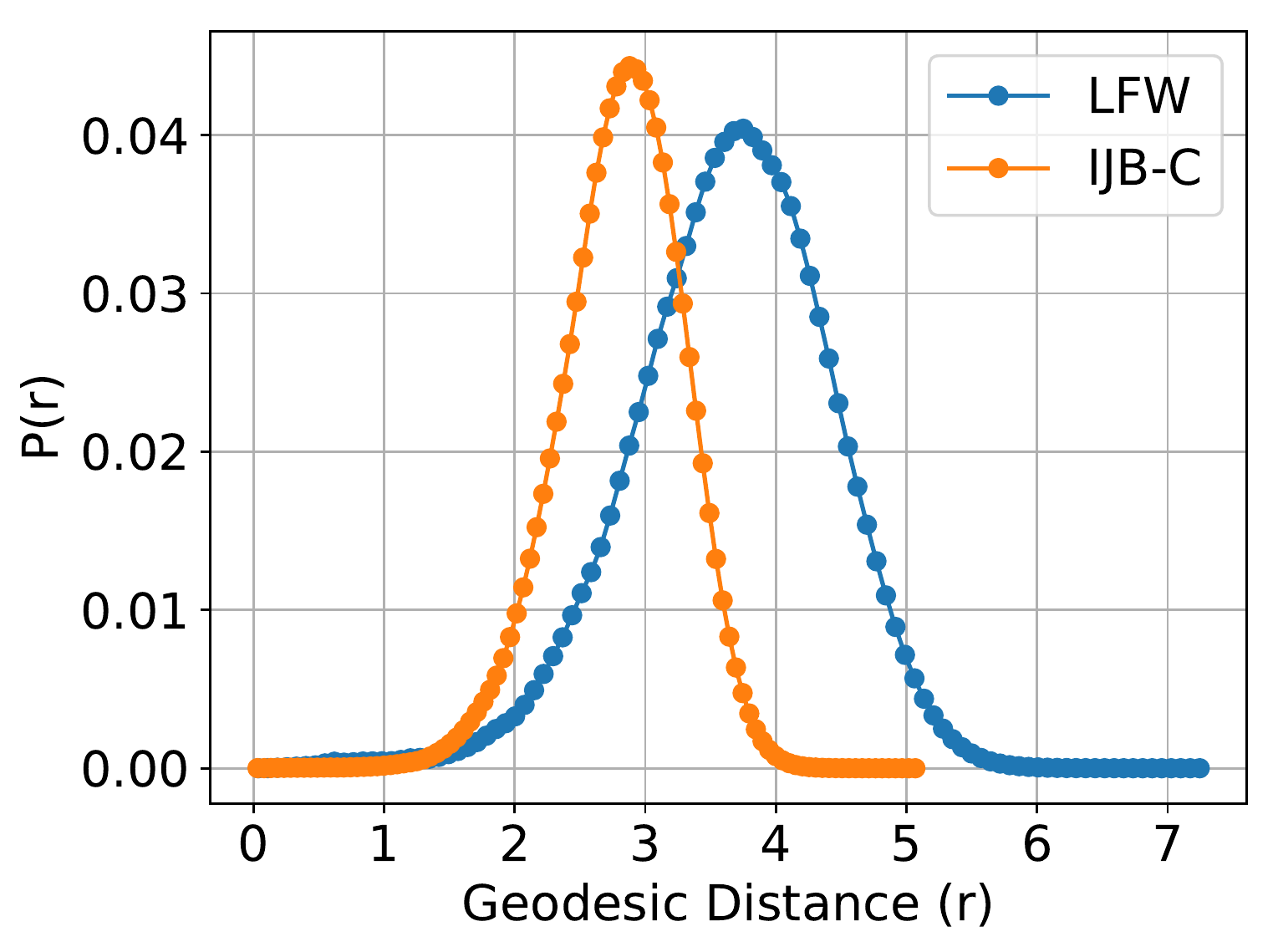}
        \caption{SphereFace}
    \end{subfigure}
    \begin{subfigure}[t]{0.23\textwidth}
        \centering
        \includegraphics[width=\textwidth]{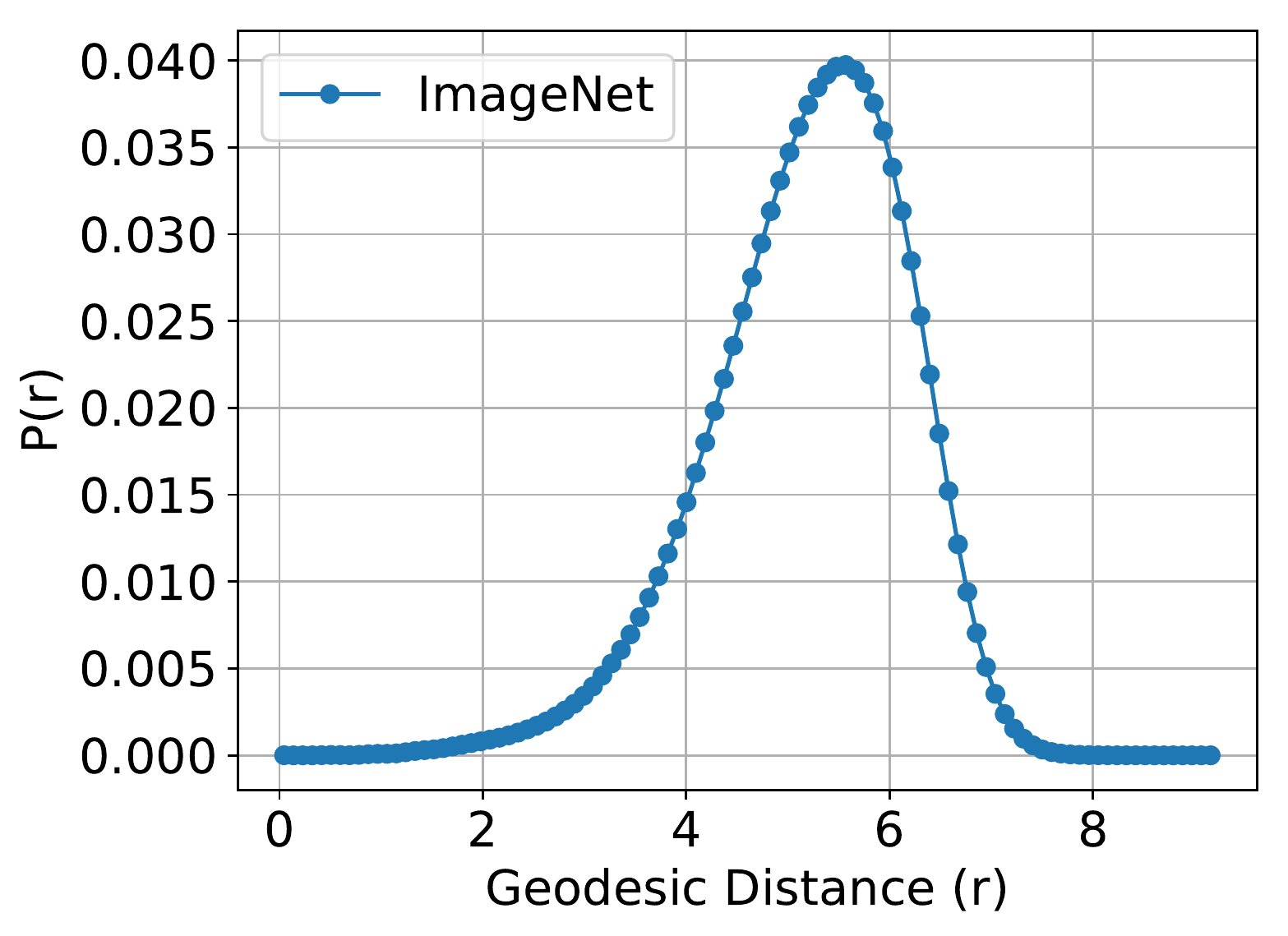}
        \caption{ResNet-34}
    \end{subfigure}
    \caption{Distribution of geodesic distances for different representation models and datasets. \label{fig:distribution}}
\end{figure*}

\begin{figure*}
    \centering
    \begin{subfigure}[t]{\textwidth}
        \centering
        \includegraphics[width=0.23\textwidth]{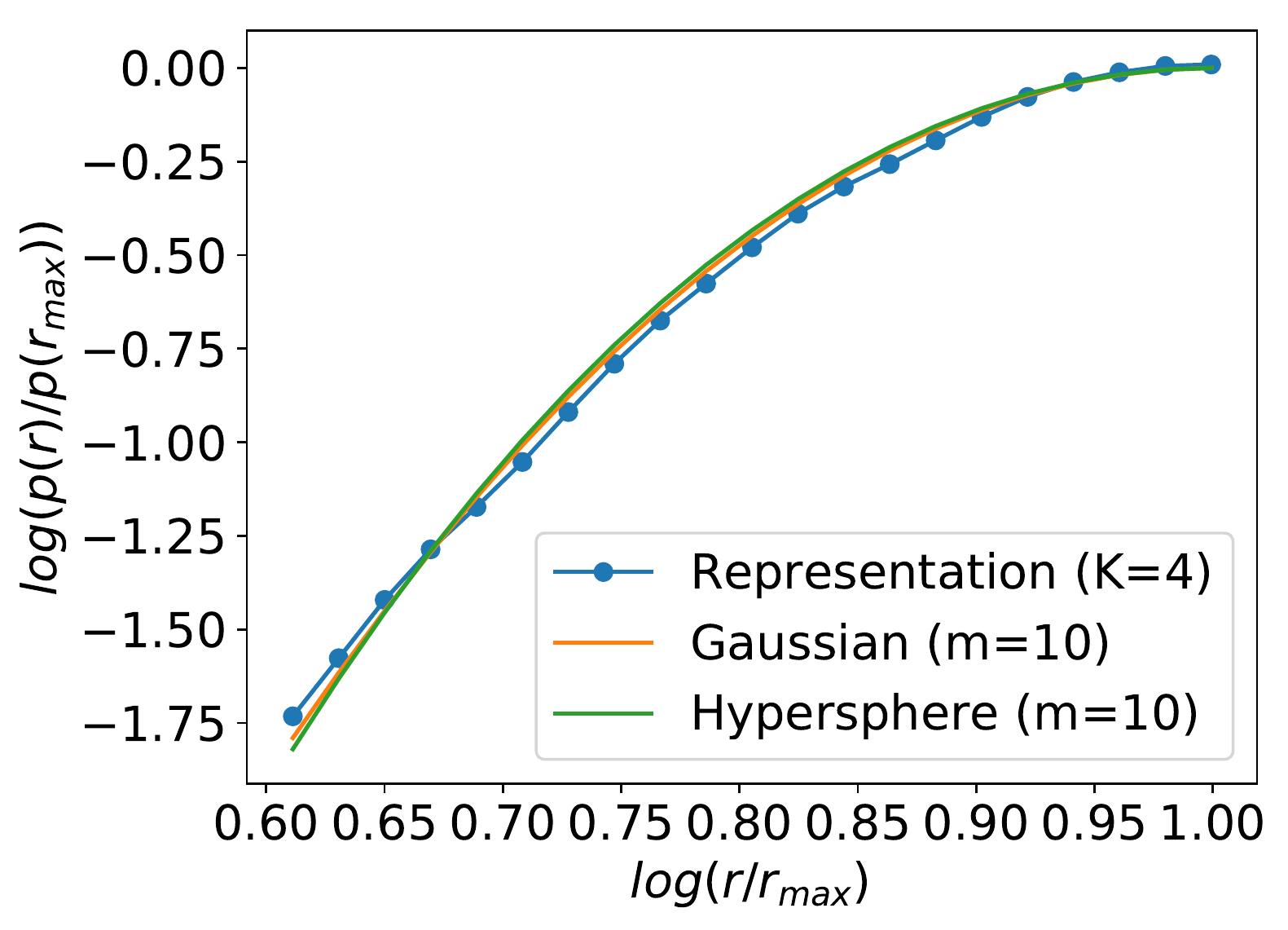}
		\includegraphics[width=0.23\textwidth]{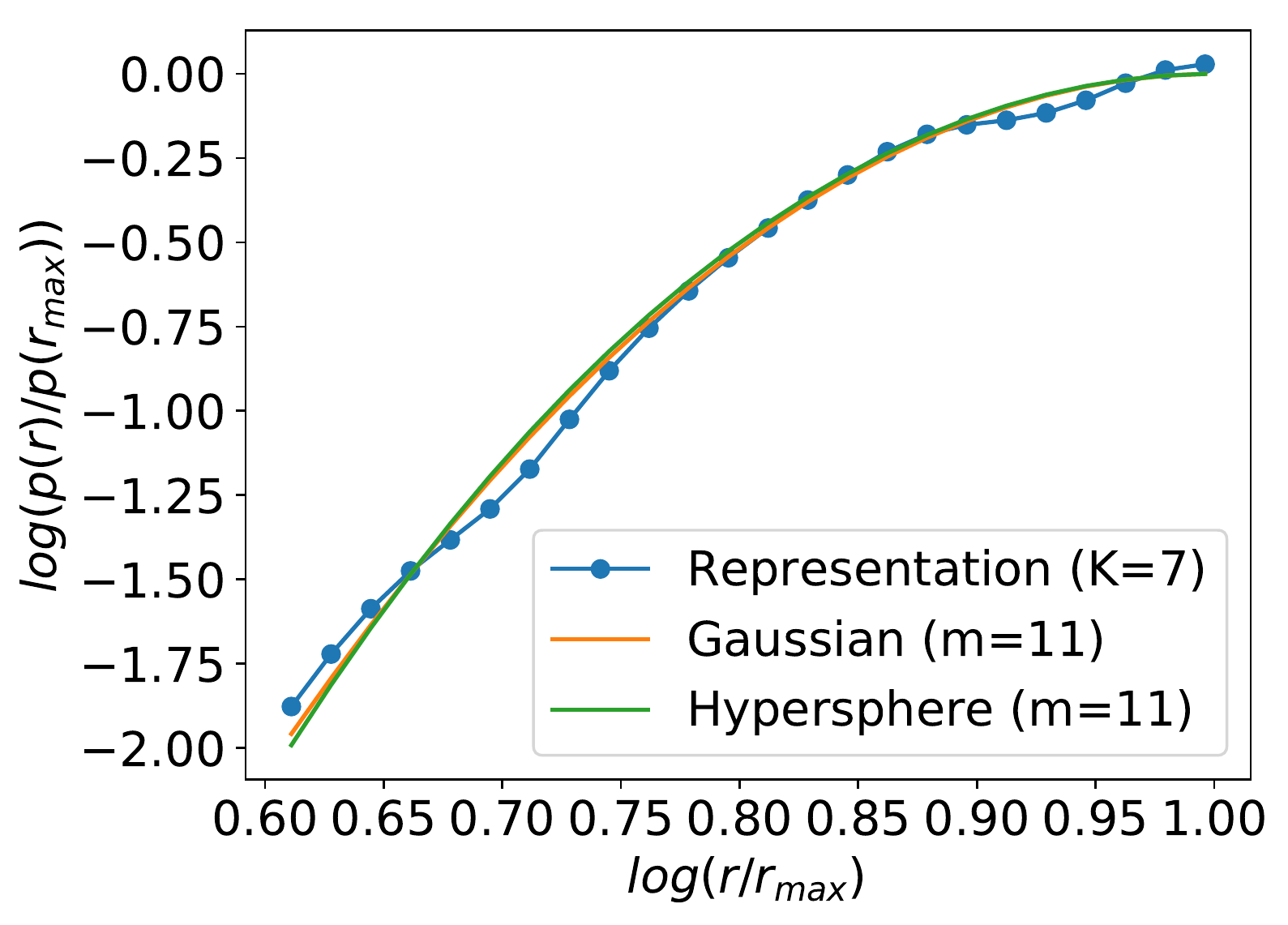}
		\includegraphics[width=0.23\textwidth]{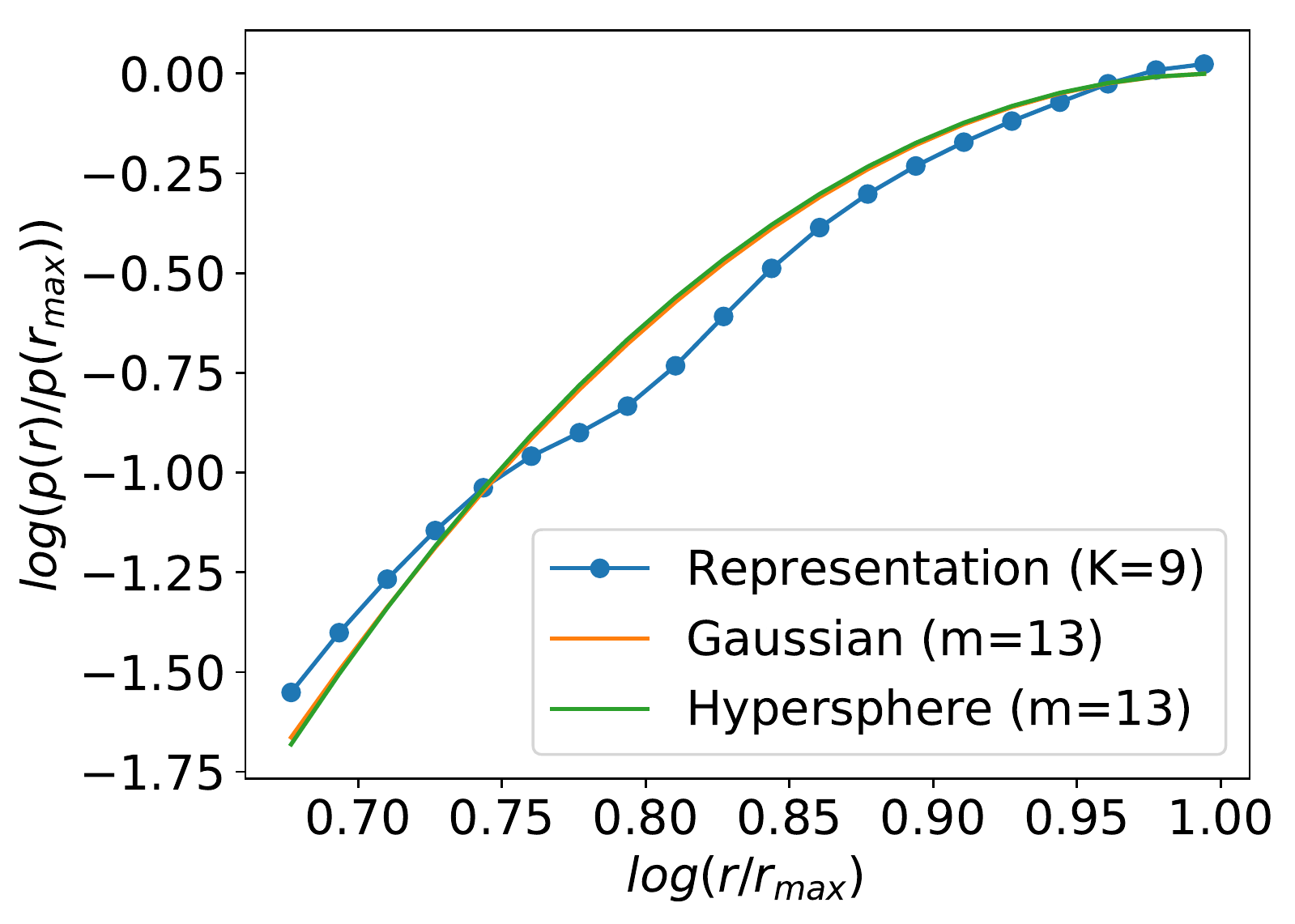}
		\includegraphics[width=0.23\textwidth]{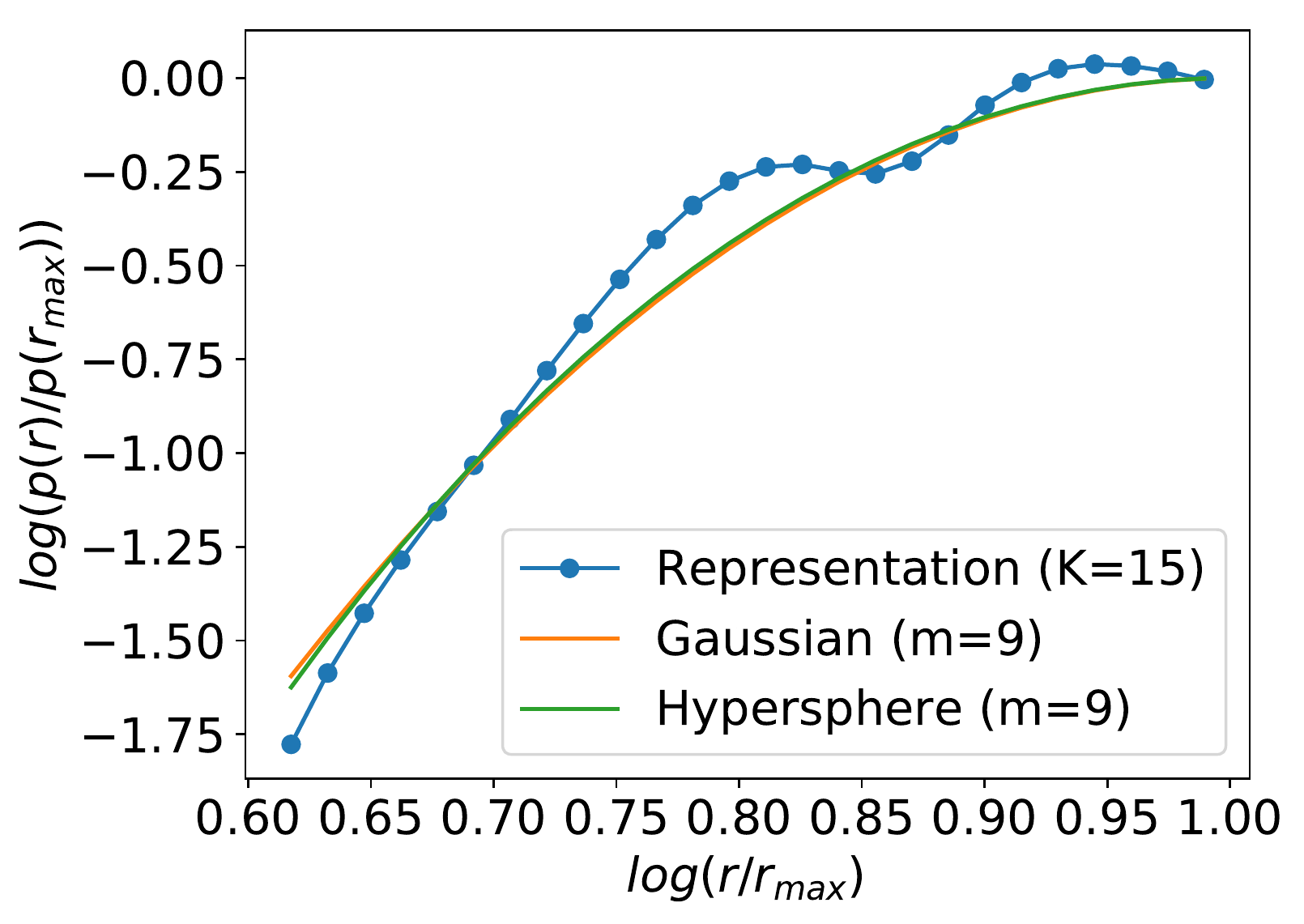}
        \caption{SphereFace on LFW Dataset} 
    \end{subfigure}
    \begin{subfigure}[t]{\textwidth}
        \centering
        \includegraphics[width=0.23\textwidth]{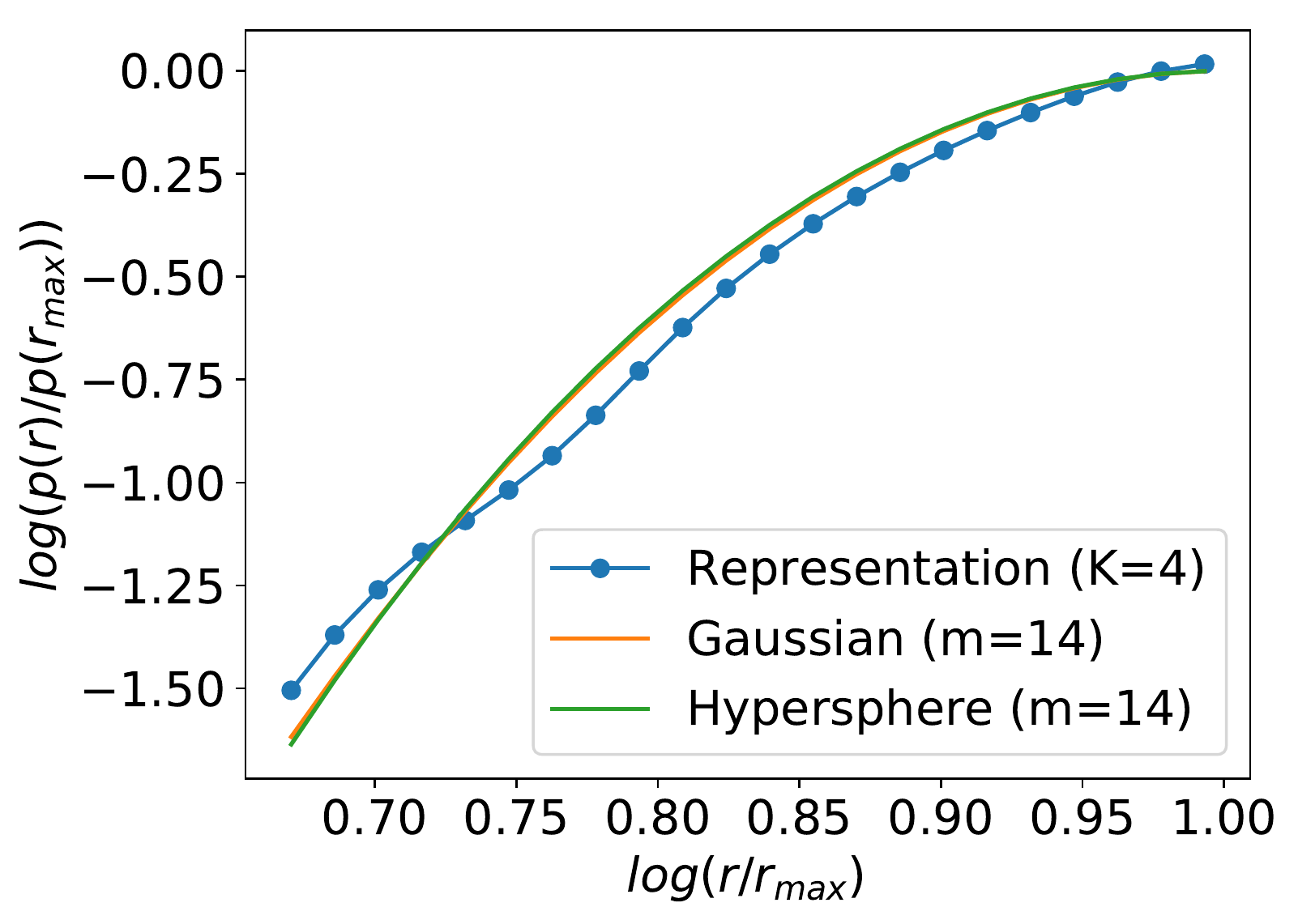}
		\includegraphics[width=0.23\textwidth]{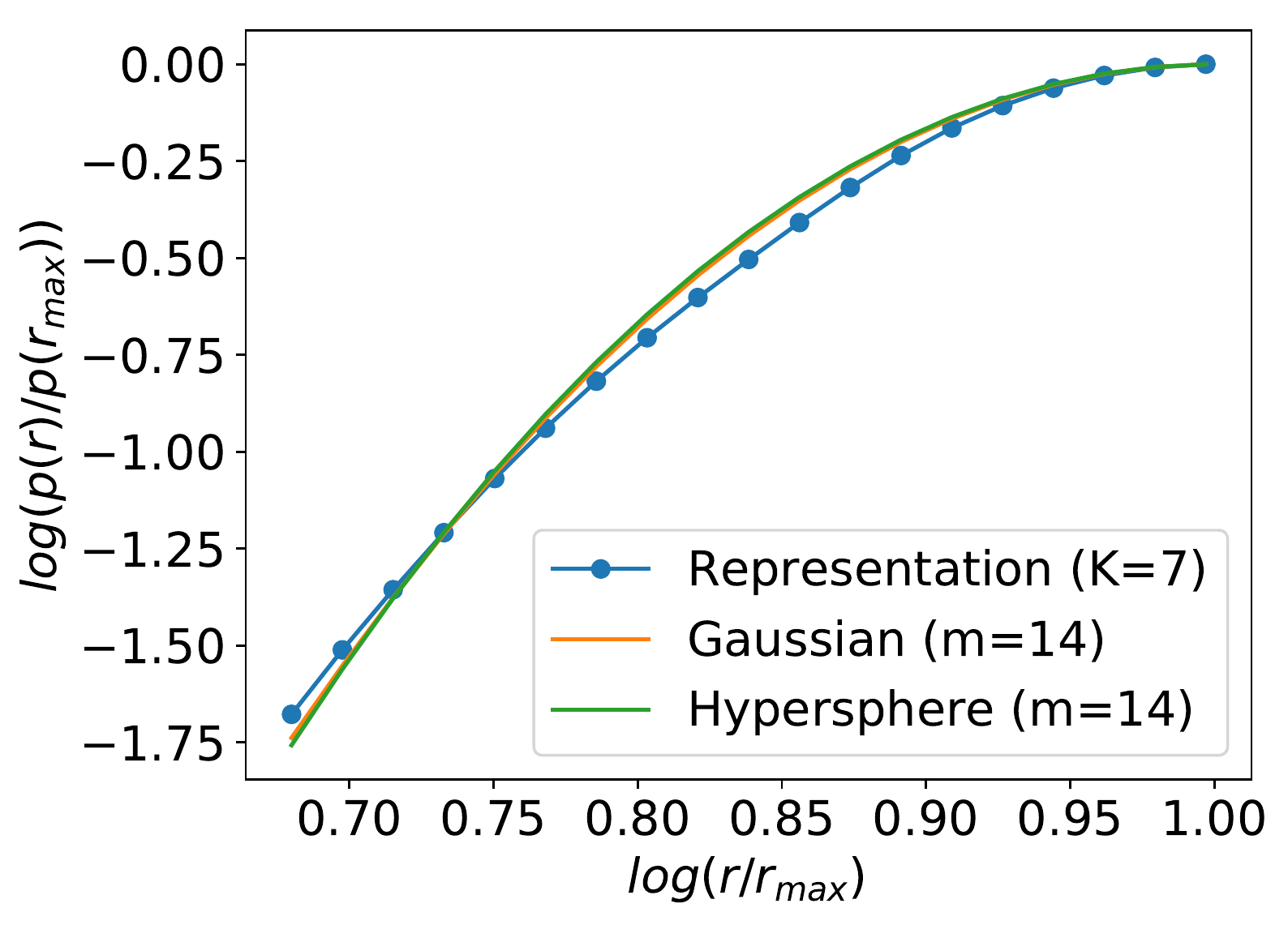}
		\includegraphics[width=0.23\textwidth]{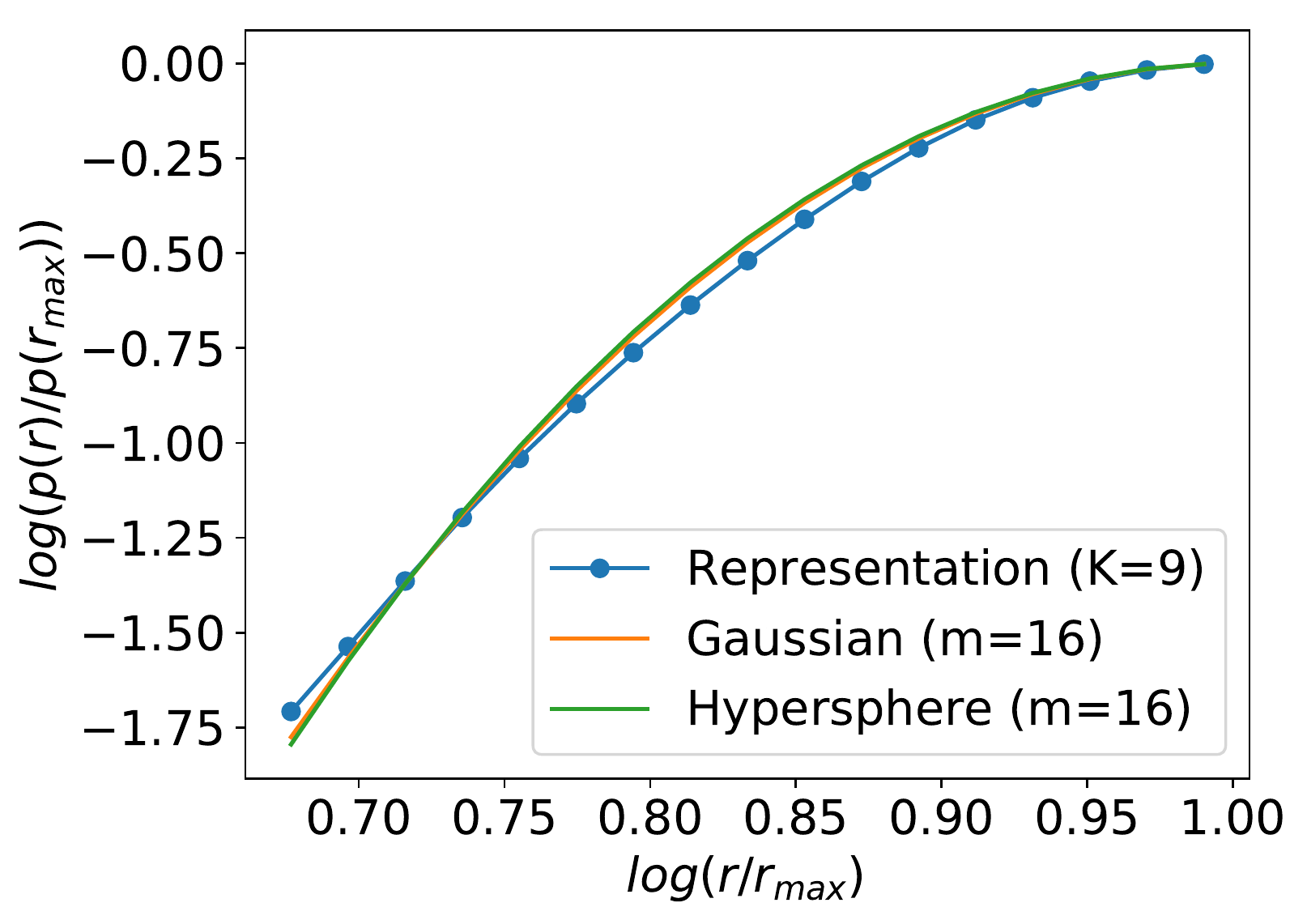}
		\includegraphics[width=0.23\textwidth]{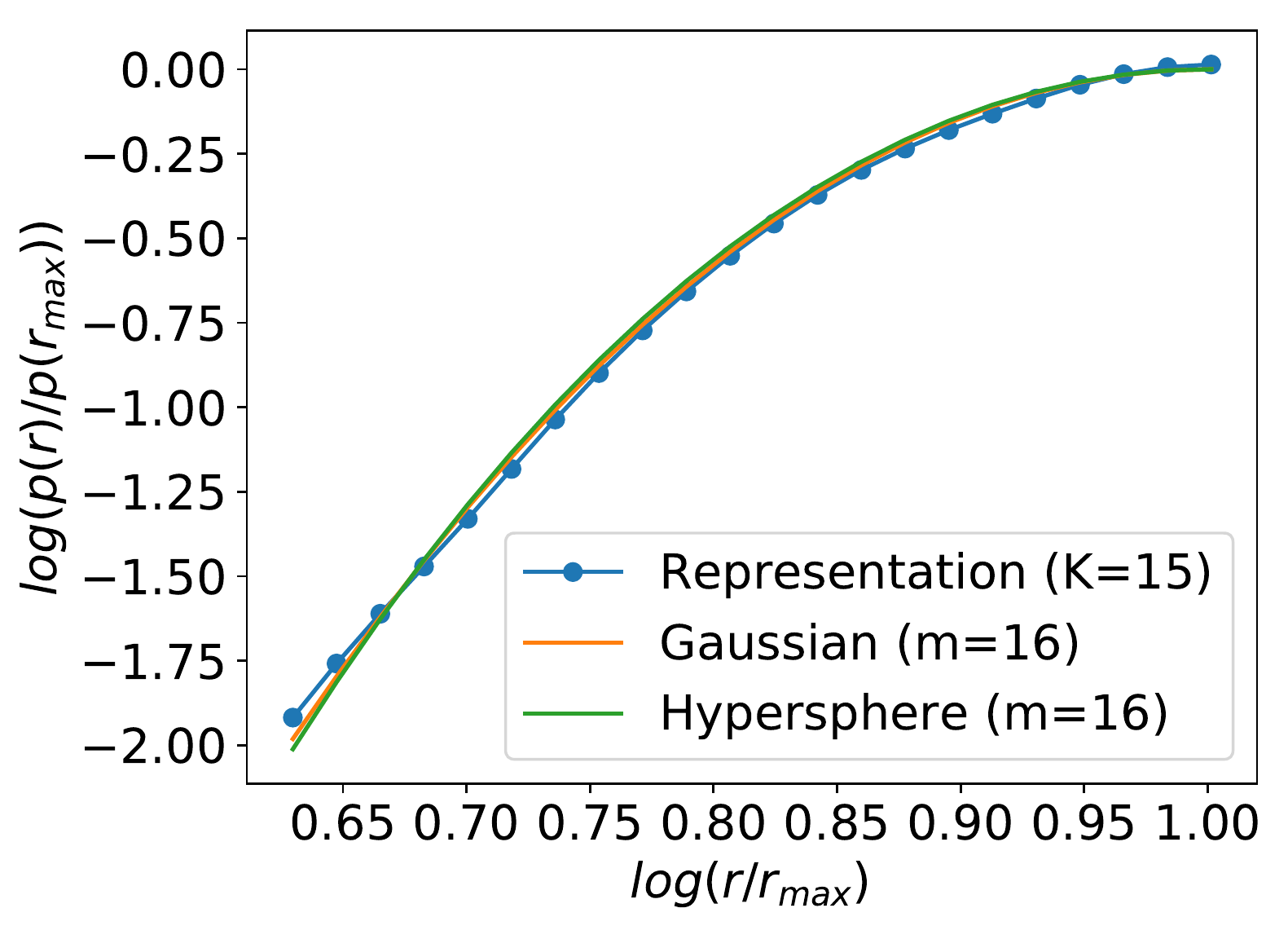}
        \caption{SphereFace on IJB-C Dataset}
    \end{subfigure}
    \begin{subfigure}[t]{\textwidth}
        \centering
        \includegraphics[width=0.23\textwidth]{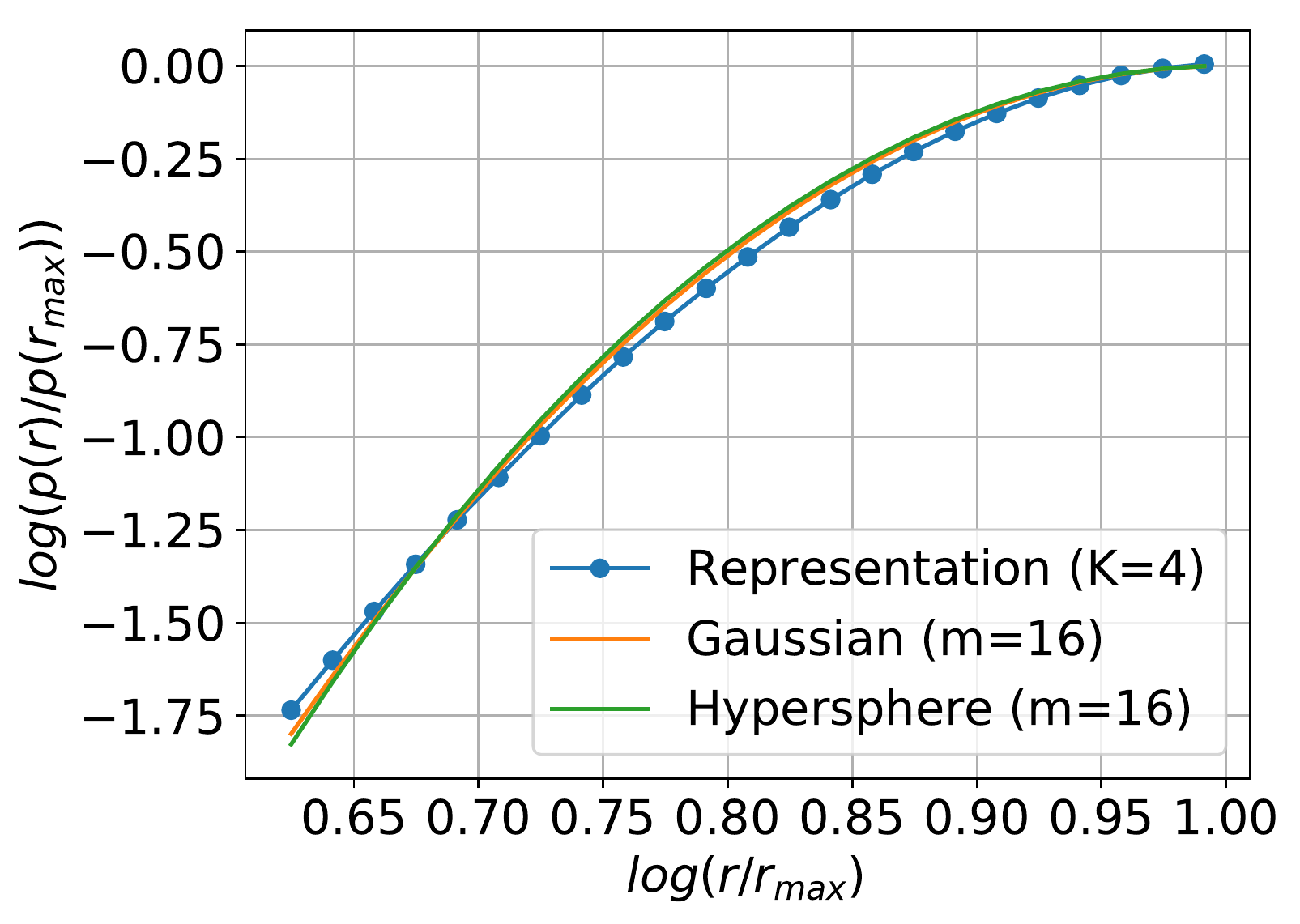}
		\includegraphics[width=0.23\textwidth]{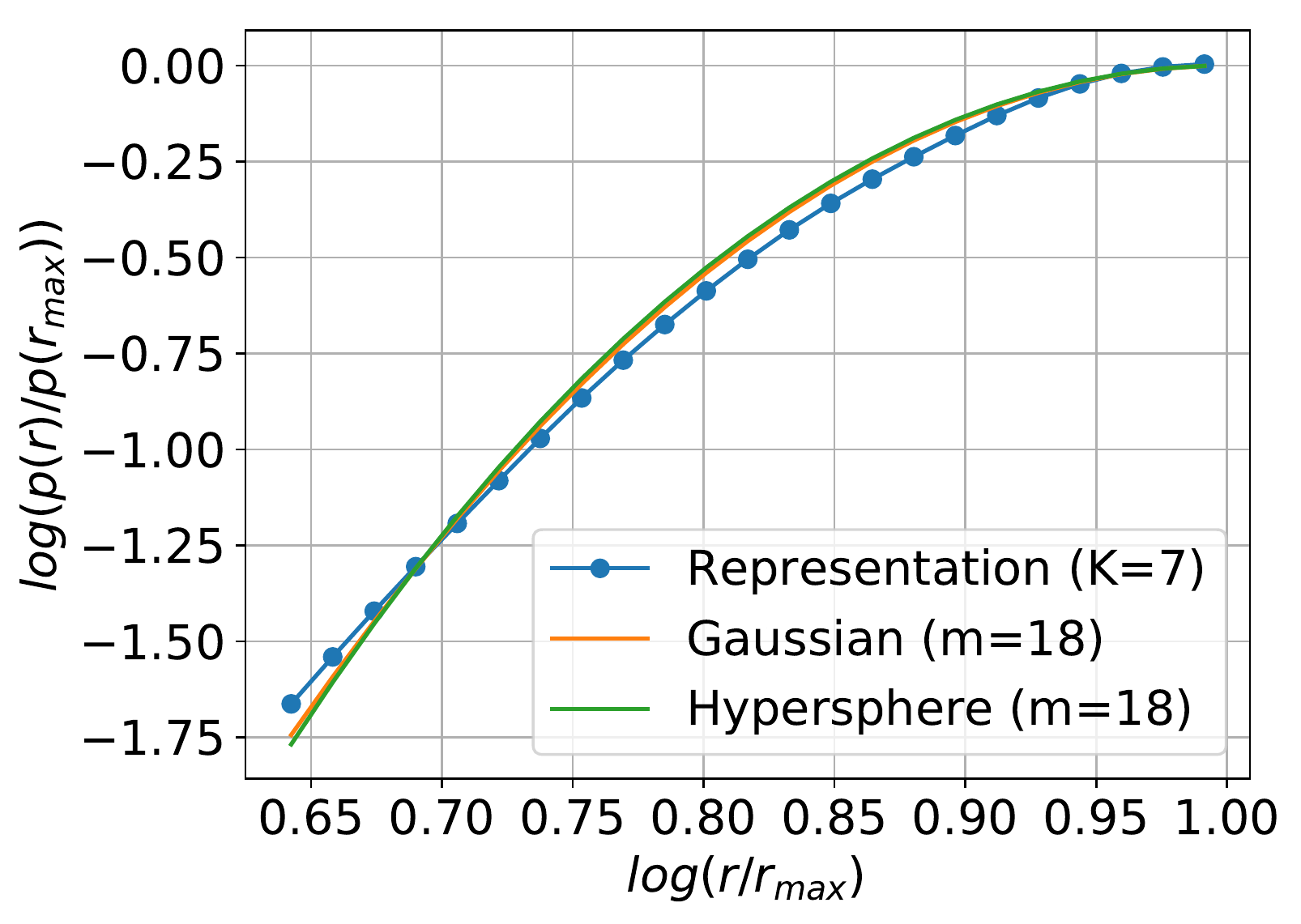}
		\includegraphics[width=0.23\textwidth]{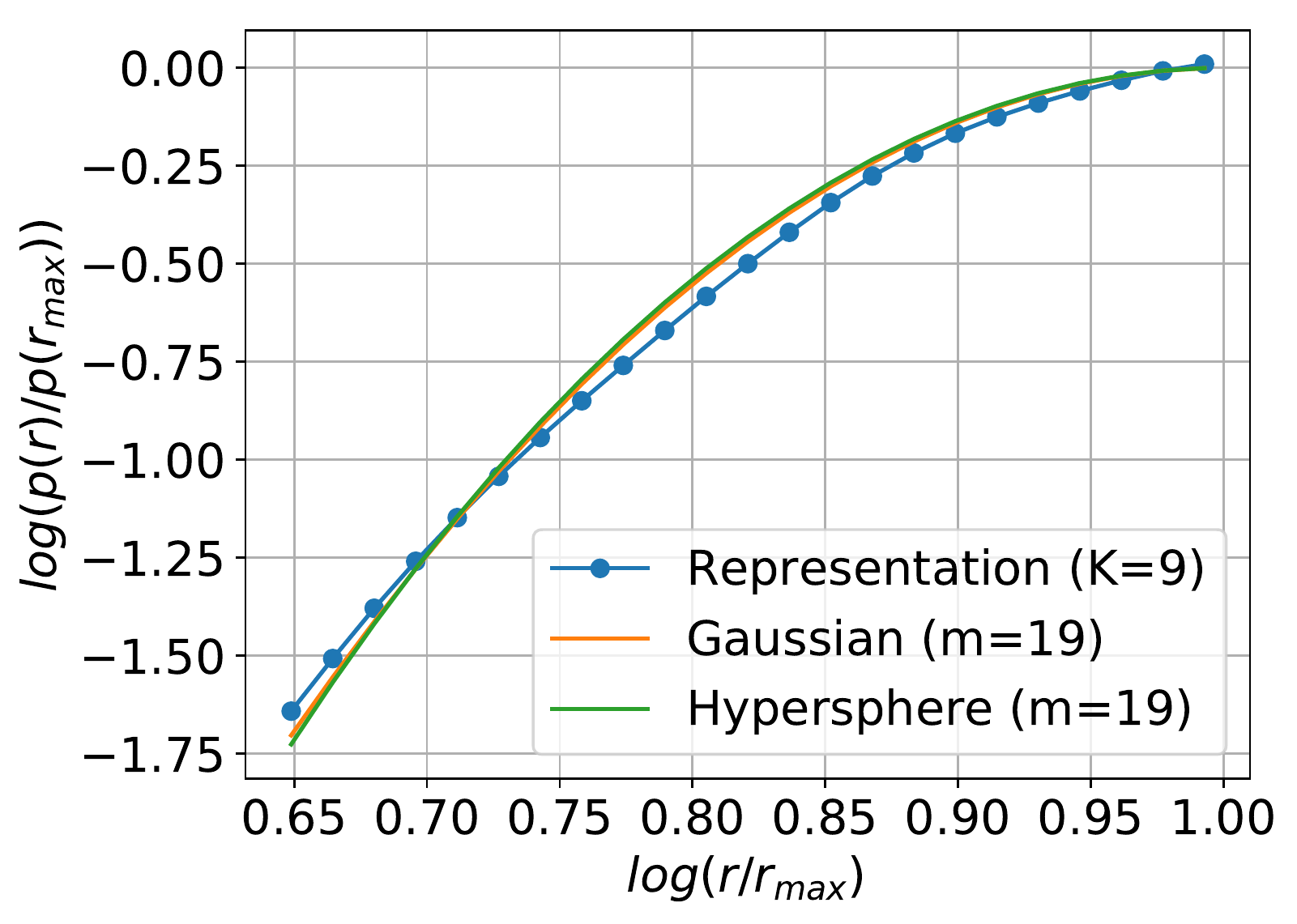}
		\includegraphics[width=0.23\textwidth]{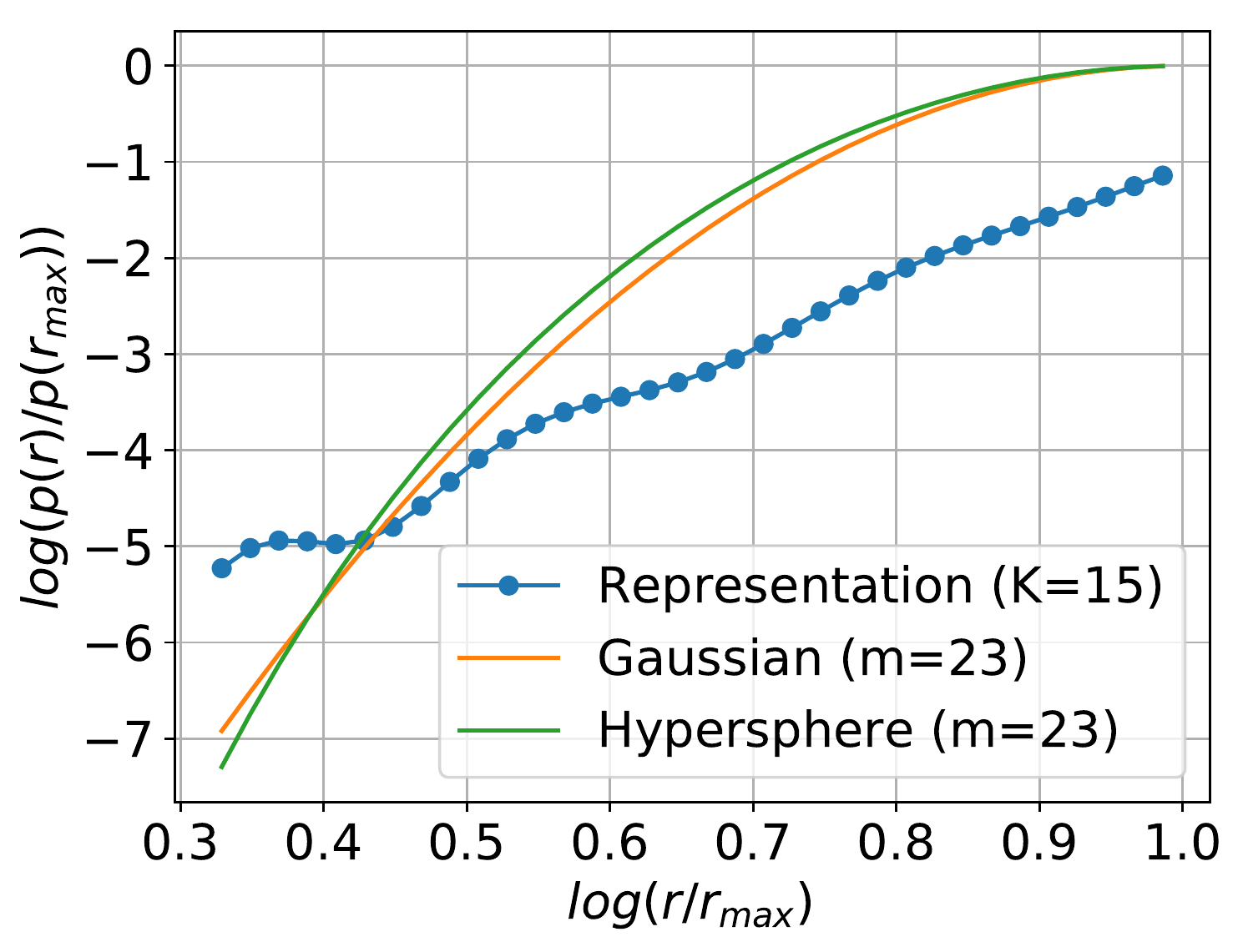}
        \caption{ResNet-34 on ImageNet Dataset}
    \end{subfigure}
    \caption{$\log \frac{\hat{p}_{\mathcal{M}}(r)}{\hat{p}_{\mathcal{M}}(r_{max})}$ vs $\log \frac{r}{r_{max}}$ plots as we vary number of neighbors $k$ for different representation models and datasets. \label{fig:fitting}}
\end{figure*}

Figure \ref{fig:distribution} shows the distribution of geodesic distances $p(r)$ for each of the datasets and representation models. Figure \ref{fig:fitting} shows the plot of $\log \frac{\hat{p}_{\mathcal{M}}(r)}{\hat{p}_{\mathcal{M}}(r_{max})}$ vs $\log \frac{r}{r_{max}}$, as we vary the number of neighbors $k$, for the SphereFace representation model on the LFW and IJB-C datasets and ResNet-34 on the ImageNet dataset.

\subsection{Swiss Roll \label{sec:swiss-roll}}
\begin{figure*}
    \begin{subfigure}[t]{0.33\textwidth}
        \centering
        \includegraphics[width=\textwidth]{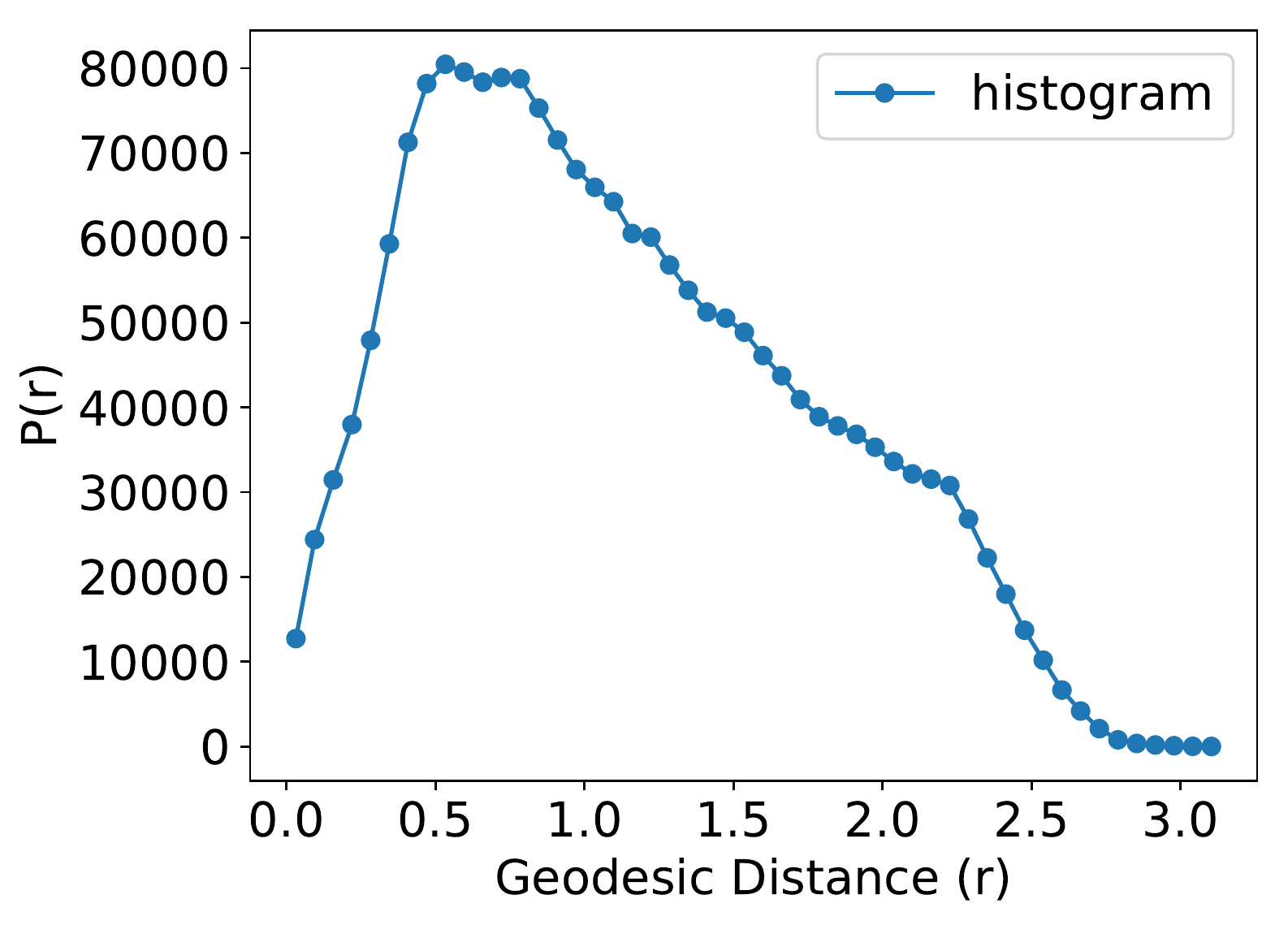}
        \caption{Histogram of Swiss Roll}
    \end{subfigure}
    \begin{subfigure}[t]{0.33\textwidth}
        \centering
        \includegraphics[width=\textwidth]{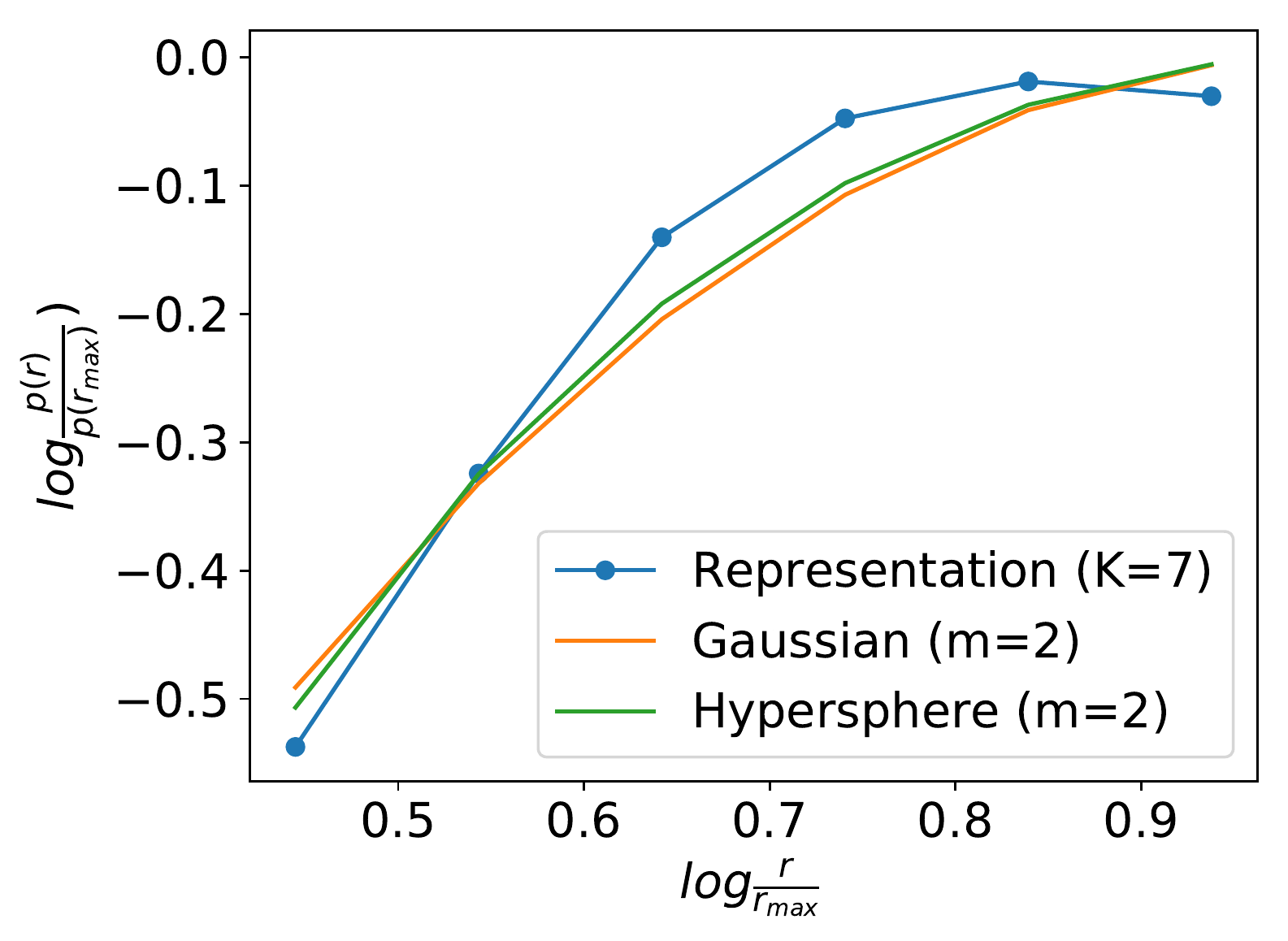}
        \caption{$\log \frac{\hat{p}_{\mathcal{M}}(r)}{\hat{p}_{\mathcal{M}}(r_{max})}$ vs $\log \frac{r}{r_{max}}$ of Swiss Roll}
    \end{subfigure}
    \begin{subfigure}[t]{0.33\textwidth}
        \centering
        \includegraphics[width=\textwidth]{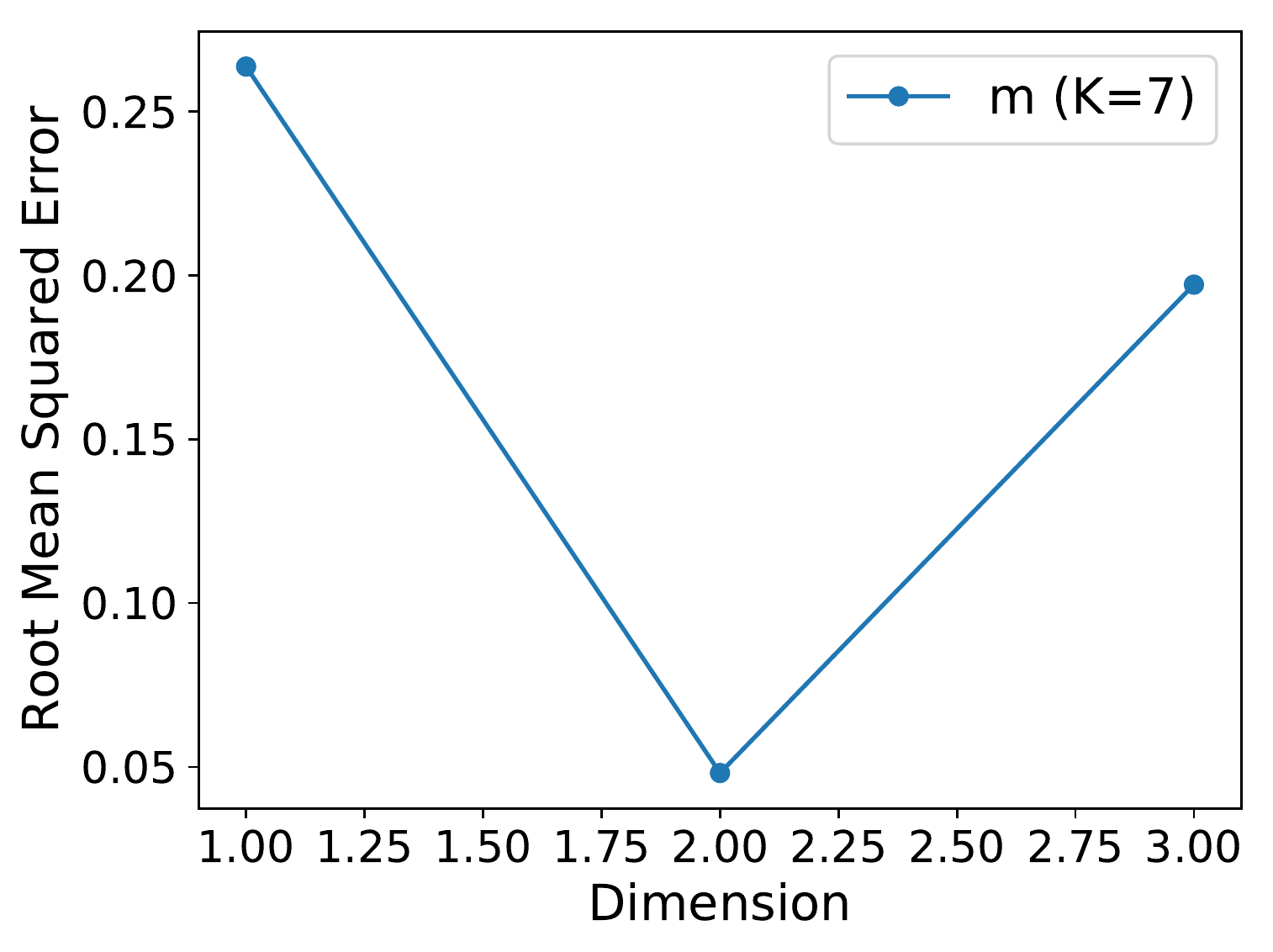}
        \caption{Dimensionality of Swiss Roll \label{fig:mapping-swiss-roll}}
    \end{subfigure}
    \caption{Intrinsic Dimensionality of Swiss Roll \label{fig:swiss-roll-id}}
\end{figure*}

In this section we consider the swiss roll dataset, as a means of providing visual validation of the estimated intrinsic space on a known dataset. First we estimate the intrinsic dimensionality of the swiss roll dataset and then we learn a low-dimensional mapping from the ambient 3-$dim$ space to the intrinsic space. We sample 2000 points from the swiss roll dataset and use these points for the experiments. For this dataset, the intrinsic dimensionality estimate is 2 dimensions (see Figure \ref{fig:swiss-roll-id}, which is indeed the ground truth intrinsic dimensionality of swiss-roll.

\begin{figure*}
    \begin{subfigure}[t]{0.33\textwidth}
        \centering
        \includegraphics[width=\textwidth]{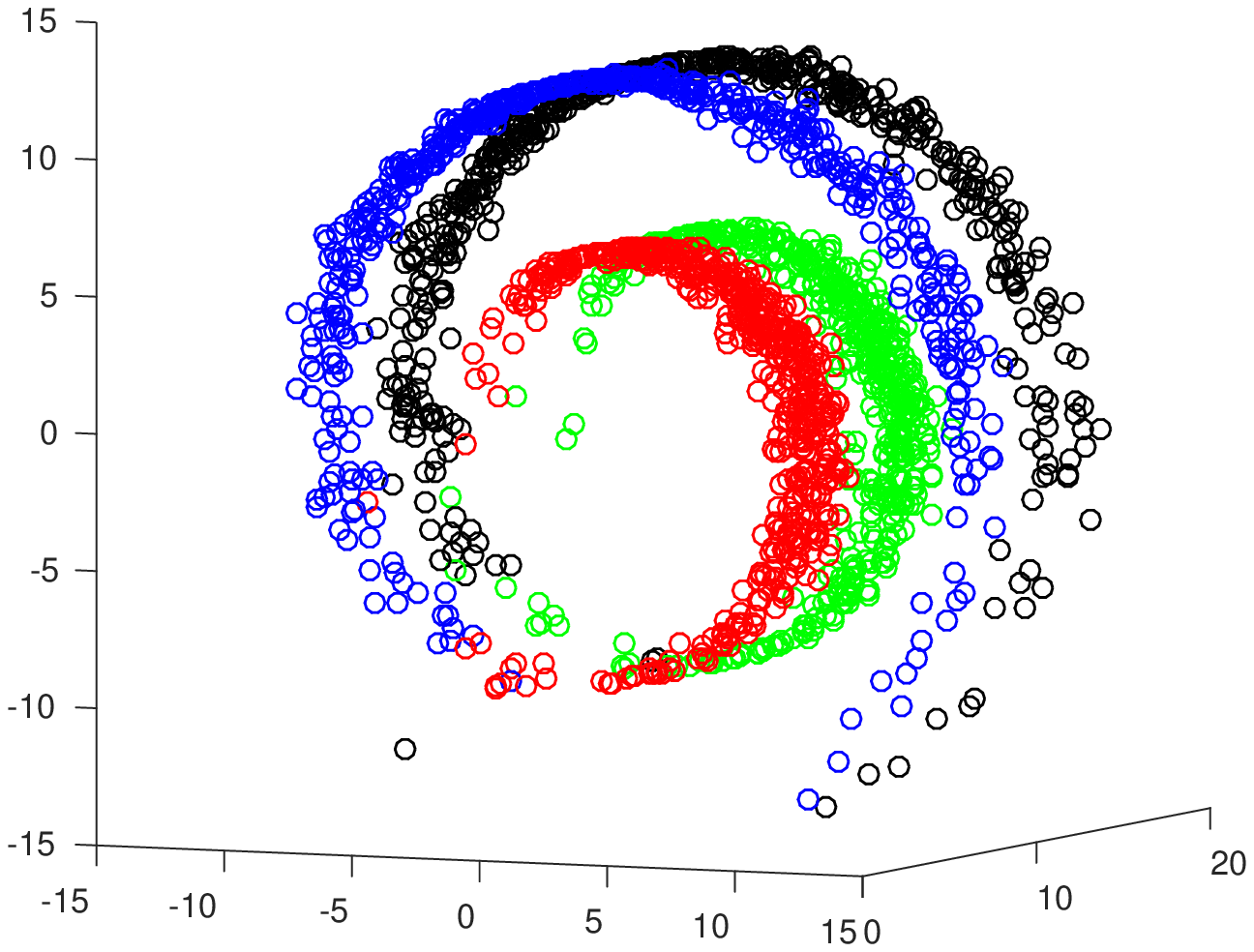}
        \caption{Swiss Roll}
    \end{subfigure}
    \begin{subfigure}[t]{0.33\textwidth}
        \centering
        \includegraphics[width=\textwidth]{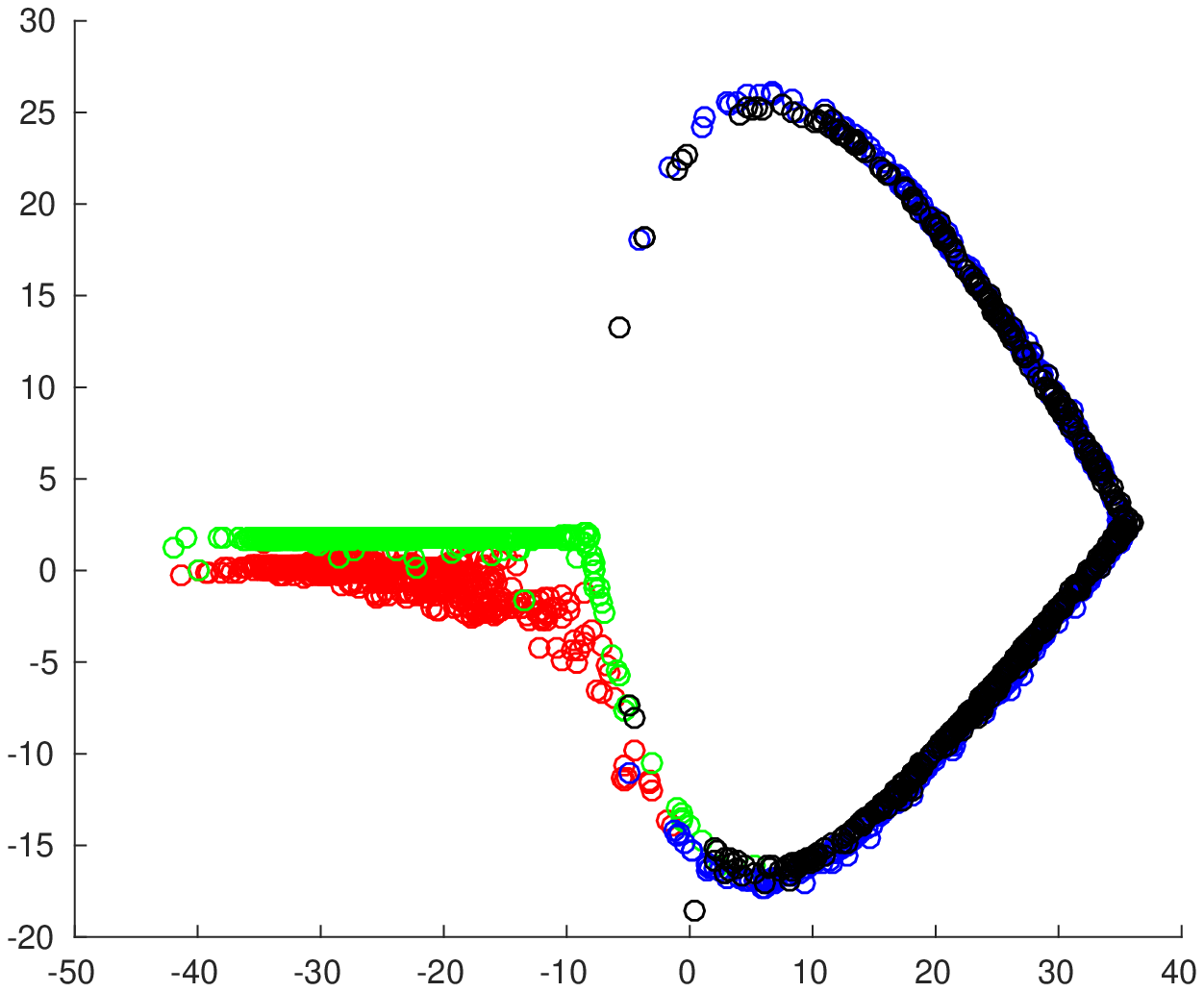}
        \caption{Projection of Isomap}
    \end{subfigure}
    \begin{subfigure}[t]{0.33\textwidth}
        \centering
        \includegraphics[width=\textwidth]{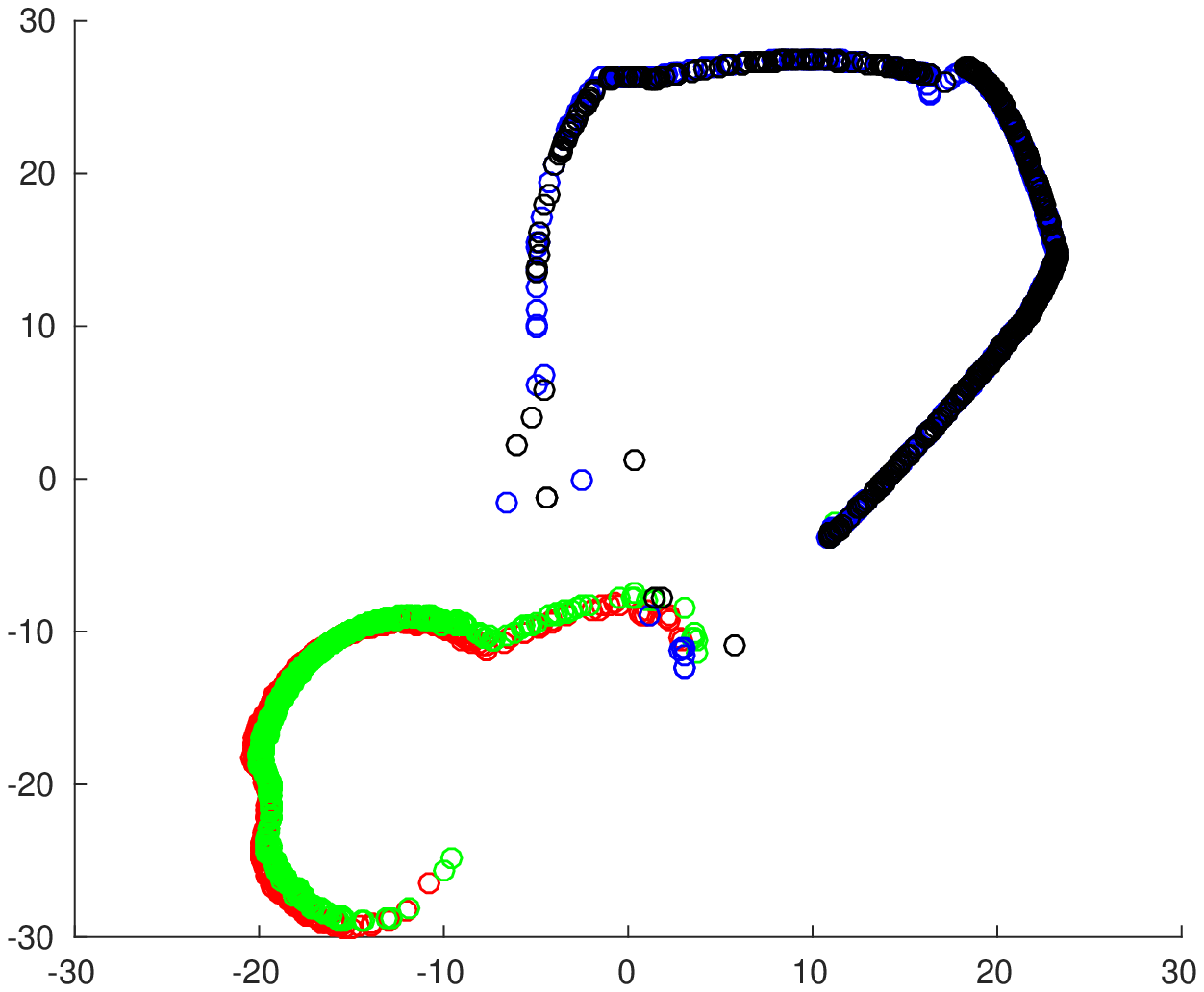}
        \caption{Projection of DeepMDS}
    \end{subfigure}
    \caption{\textbf{Swiss Roll:} (a) the original 2000 points from the swiss roll manifold, (b) the 2-$dim$ intrinsic space estimated by Isomap, and (3) the 2-$dim$ intrinsic space estimated by our proposed method DeepMDS. In both cases, the blue and black points, and correspondingly green and red points, are close together in both the intrinsic and ambient space. \label{fig:id-swiss-roll}}
\end{figure*}

{\small
\bibliographystyle{ieee}
\bibliography{egbib}
}

{\small
\bibliographystyle{ieee}
\bibliography{egbib}
}

\end{document}